%% file: main.tex
\documentclass[11pt,a4paper]{article}
\usepackage[margin=1.05in]{geometry}
\usepackage{amsmath,amssymb,amsthm,mathtools,bm}
\usepackage[hidelinks]{hyperref}
\usepackage{booktabs,array,tabularx,graphicx,microtype,rotating}
\AtBeginDocument{%
  \setlength{\abovedisplayskip}{8pt plus 2pt minus 2pt}%
  \setlength{\belowdisplayskip}{8pt plus 2pt minus 2pt}%
  \setlength{\abovedisplayshortskip}{0pt plus 2pt}%
  \setlength{\belowdisplayshortskip}{5pt plus 2pt minus 2pt}}
\usepackage{tikz}
\usetikzlibrary{arrows.meta}
\usepackage{authblk}

\hypersetup{pdftitle={The Prior-Force Family of Implicit Denoiser Priors and Its Decisions in Linear Inverse Problems},
  pdfauthor={Weijie Liang, Chuling Wen, Jian Lu}}
\usepackage[capitalise,noabbrev]{cleveref}

\usepackage{aliascnt}
\newcommand{\sharedthm}[2]{%
  \newaliascnt{#1}{theorem}\newtheorem{#1}[#1]{#2}\aliascntresetthe{#1}%
  \crefname{#1}{#2}{#2s}\Crefname{#1}{#2}{#2s}}
\theoremstyle{plain}
\newtheorem{theorem}{Theorem}[section]
\crefname{theorem}{Theorem}{Theorems}\Crefname{theorem}{Theorem}{Theorems}
\sharedthm{proposition}{Proposition}
\sharedthm{lemma}{Lemma}
\sharedthm{corollary}{Corollary}
\theoremstyle{definition}
\sharedthm{definition}{Definition}
\sharedthm{assumption}{Assumption}
\sharedthm{example}{Example}
\theoremstyle{remark}
\sharedthm{remark}{Remark}
\newtheorem*{remark*}{Remark}
\crefname{appendix}{Appendix}{Appendices}\Crefname{appendix}{Appendix}{Appendices}

\newcommand{\gph}{\operatorname{gph}}
\newcommand{\dom}{\operatorname{dom}}
\newcommand{\ran}{\operatorname{ran}}
\newcommand{\zer}{\operatorname{zer}}
\newcommand{\Fix}{\operatorname{Fix}}
\newcommand{\prox}{\operatorname{prox}}
\newcommand{\dist}{\operatorname{dist}}
\newcommand{\Cov}{\operatorname{Cov}}
\newcommand{\Tr}{\operatorname{Tr}}
\newcommand{\supp}{\operatorname{supp}}
\newcommand{\Lip}{\operatorname{Lip}}
\newcommand{\inn}[2]{\langle #1,\, #2\rangle}
\newcommand{\norm}[1]{\lVert #1 \rVert}
\newcommand{\abs}[1]{\lvert #1 \rvert}
\DeclareMathOperator*{\argmin}{arg\,min}
\DeclareMathOperator*{\argmax}{arg\,max}

\newcommand{\PP}{\mathcal{P}}

\title{\bf The Prior-Force Family of Implicit Denoiser Priors\\ and Its Decisions in Linear Inverse Problems}
\author[a]{Weijie Liang}
\author[a]{Chuling Wen}
\author[a,b,$\ast$]{Jian Lu}
\affil[a]{Shenzhen Key Laboratory of Advanced Machine Learning and Applications, School of Mathematical Sciences,
Shenzhen University, Shenzhen, 518060, China}
\affil[b]{National Center for Applied Mathematics Shenzhen (NCAMS), Shenzhen, 518055, China}
\date{}

\begin{document}
\maketitle
{\renewcommand\thefootnote{}\footnotetext{\raggedright$^\ast$Corresponding author.\\
Email addresses: liangweijie2022@email.szu.edu.cn (Weijie Liang), 2400191020@mails.szu.edu.cn (Chuling Wen),
jianlu@szu.edu.cn (Jian Lu).}}

\begin{abstract}
\input{sections/abstract}

\end{abstract}



\input{sections/s1_intro}
\input{sections/s2_family}
\input{sections/s3_preference}

\input{sections/s4_decisions}
\input{sections/s5_recovery}
\input{sections/s6_experiments}

\input{sections/s7_conclusion}

\appendix
\crefalias{section}{appendix}\crefalias{subsection}{appendix}
\input{sections/app_proofs}

{\footnotesize\raggedright
\bibliographystyle{unsrt}
\bibliography{refs}
}

\end{document}

%% file: sections/abstract.tex
Implicit priors given by Gaussian denoisers successfully regularize a wide range of inverse problems in approaches such as plug-and-play (PnP), regularization by denoising (RED) and score-based methods. Such a prior depends on the noise level, and how it selects among signals consistent with the data remains largely uncharacterized.
To put implicit and explicit priors on the same footing, we define the prior-force family $W_t=t^{-1}(D_t^{-1}-I)$ of any Gaussian denoiser $D_t$ at noise variance $t>0$ (inverse in the graph sense), the normalized denoising residual as an operator on clean signals. At a common effective prior weight, the equilibrium equations differ only in their prior terms: PnP uses $W_t$, RED/score-based methods the score force (the normalized residual as an operator on noisy inputs), and maximum a posteriori (MAP) estimation the MAP force (the negative log-prior gradient). For the minimum mean-square error (MMSE) denoisers of a prior $\pi$, we compare the corresponding energies at three levels. Without observation, raising the noise level favors PnP minima that retain more posterior variance. On signals consistent with the data, the leading-order PnP energy for separated modes is the proximal hull of the MAP energy, which lets PnP select signals between modes; modes of equal data fit are ranked by peak height at small noise levels and by mass at large ones. For recovery, $W_t$ at the truth bounds the local bias, and an oracle noise level balances this bias against denoiser error. These results show that the noise level is a parameter of the prior rather than a step size.
Experiments with exact priors support these predictions, including the mixture components that PnP selects on unseen data.

%% file: sections/s1_intro.tex
\section{Introduction}\label{sec:intro}

Plug-and-play (PnP), regularization by denoising (RED) and score-based methods bring a prior into the variational solution
of imaging inverse problems through a denoiser for Gaussian noise \cite{Kamilov2023,MilanfarDelbracio2024}. PnP methods replace the proximal map of a regularizer in a
splitting algorithm by a denoiser \cite{Venkatakrishnan2013,Ryu2019}. RED builds the regularizing term
from the denoising residual \cite{RomanoEladMilanfar2017}. Score-based and diffusion methods use the same residual, through Tweedie's
formula, as an estimate of the score of a smoothed prior \cite{Song2021,ChungDPS2023,DarasSurvey2024}. In each case a denoiser for Gaussian
noise of variance $t$ takes the place of an explicit prior, and the data term and the physics of the measurement remain those of
the classical variational model. Any Gaussian denoiser can be used; trained deep networks are the most common choice.

In these methods the prior that enters the reconstruction is implicit. It is defined by the denoiser and by the noise
level at which the denoiser is called, not by a formula that can be inspected. Even an exact minimum mean-square error (MMSE) denoiser
does not penalize by the negative log-density of its prior \cite{Gribonval2011}, and in practice the noise level is a tuning
parameter. The difference matters most when the measurements leave part of the signal undetermined, as in inpainting,
super-resolution, compressed sensing or limited-angle tomography. There the data carry no information about the unobserved directions, and
the choice among the signals consistent with the data is made by the implicit prior alone. Existing theory describes the implicit prior through an energy at a fixed noise level, the
implicit regularizer, or through the score of the smoothed prior, and mostly concerns the convergence of the iterations. It does
not say which reconstruction the implicit prior selects in the unobserved directions, how that choice compares with the maximum a posteriori (MAP) estimate of the prior, or how it changes with the noise level.

This paper answers these questions for Gaussian denoisers. A denoiser $D_t$ for additive Gaussian noise of variance $t$ enters
an inverse problem only through its evaluations, usually at many noise levels. Each evaluation records a noisy input $z$, a
denoised output $D_t(z)$ and the normalized denoising residual; we call the set of such triples the \emph{denoising graph},
\begin{equation}\label{eq:graph-intro}
 \mathcal G_t=\Bigl\{\bigl(z,\ x,\ w\bigr):\ x=D_t(z),\ \ w=\tfrac{z-x}{t}\Bigr\}.
\end{equation}
Expressing $w$ as a function of the noisy input gives
\begin{equation}\label{eq:score-intro}
 \nabla E_t(z)=\tfrac{z-D_t(z)}{t},
\end{equation}
which equals the negative score $-\nabla\log p_t(z)$ for the MMSE denoiser, by Tweedie's formula.
Here $p_t$ is the density after Gaussian smoothing of the prior.
Expressing the same $w$ as a function of the denoised output gives a map that we call the \emph{prior force},
\begin{equation}\label{eq:W-intro}
 W_t(x)=\tfrac{D_t^{-1}(x)-x}{t},\qquad \nabla E_t(z)\in W_t\bigl(D_t(z)\bigr),
\end{equation}
so every forward evaluation supplies a point of the prior-force graph.
For an injective denoiser the inclusion is an equality: the two forces take the same values and differ only in where they
are attached.

We argue that the object to study is the \emph{prior-force family} $(W_t)_{t>0}$. The normalization by $t^{-1}$ makes $W_t$ act on
the same variable as the MAP force $\nabla g$ of a prior $\pi=e^{-g}$ on clean signals, and gives it the same size: for smooth priors
$W_t\to\nabla g$ as $t\to0$, a nontrivial limit. The prior force, the score force and the MAP force can therefore be compared
directly. Three facts support this choice.
\begin{itemize}
\item It is obtained from forward evaluations by \eqref{eq:W-intro}, and it is defined for every denoiser, including trained networks
 that have no induced energy.
\item The equilibrium equations of PnP, score-based and MAP reconstruction, \eqref{eq:pnp-intro}--\eqref{eq:map-intro}
 below, differ only in the prior term: $W_t$, the score force $\nabla E_t$ and the MAP force $\nabla g$. Among signals with the
 same measurements, the reconstruction is determined by the component of the prior term in $\ker A$ alone.
\item The prior force depends on $t$: the prior-force family is constant in $t$ only for Gaussian priors (\Cref{thm:rigidity}, after
 \cite{Liu2026TwoLevels}). The noise level is therefore a parameter of the prior, not a step size.
\end{itemize}
For a Bayes denoiser $D_t=\prox_{tR_t}$ and $W_t=\nabla R_t$ for an energy $R_t$ \cite{Gribonval2011}: at each scale the prior force
is the gradient of the implicit regularizer of PnP, hence its name. The score force is the force of the smoothed prior $p_t$, the
prior implicit in a denoiser in the sense of \cite{KadkhodaieSimoncelli2021}, and acts on the noisy input. We nevertheless work with
the prior force rather than with $R_t$. It is defined where no energy exists, it is the term that enters the equilibrium
equations, and its graph is available from forward evaluations, whereas differences of $R_t$ must be obtained by integrating the
prior force. Existing work uses the dependence on $t$ for specific purposes, such as convergence, annealing or the existence of a common
regularizer. Here the family itself is the object of study.

The reconstruction method determines which prior term enters its equilibrium equation.
Let $x^*$ be reconstructed from $y^\delta=Ax^*+v$, $\norm v\le\delta$, with $\ker A\neq\{0\}$.
Fixed points of PnP forward--backward splitting, the alternating direction method of multipliers (ADMM), Douglas--Rachford splitting and consensus equilibrium solve
\begin{equation}\label{eq:pnp-intro}
 A^*(Ax-y^\delta)+\alpha\,W_t(x)\ni0,
\end{equation}
where the prior force is evaluated at the denoised output.
Fixed points of RED and stationary points of score-based objectives satisfy
\begin{equation}\label{eq:red-intro}
 A^*(Ax-y^\delta)+\alpha\,\nabla E_t(x)=0,
\end{equation}
where the reconstruction is the denoiser input.
The MAP estimate of $\pi$ satisfies
\begin{equation}\label{eq:map-intro}
 A^*(Ax-y^\delta)+\alpha\,\nabla g(x)=0,
\end{equation}
with effective prior weight $\alpha$.
For denoiser-based methods, the step size, penalty or weight fixes it (\Cref{prop:noisy-coordinates}).
A forward--backward step $\gamma$ gives $\alpha=t/\gamma$.
An ADMM penalty $\rho$ gives $\alpha=\rho t$, and a RED weight $\lambda$ gives $\alpha=\lambda t$.
Changing the noise level while holding these algorithmic parameters fixed therefore also changes the effective prior weight.

The three equations share the data term.
Along $\ker A$ the data term vanishes, so among signals with the same measurements the prior term alone determines the
reconstruction.
The \emph{measurement fiber} $S_y=\{Au=y\}$ is the set of signals consistent with exact data.
On this set, when do the PnP, score and MAP decisions coincide, when do they differ, and how do they change with $t$?
A penalized solution is stationary on its own measurement fiber (\Cref{prop:own-fibre}).
A global minimizer also minimizes its energy on that fiber.
Comparing decisions on a common fiber isolates the choice in the unobserved directions.

The prior-force formulation applies more broadly than its energy interpretation.
The inverse in \eqref{eq:W-intro} is taken in the sense of graphs.
For a general denoiser $W_t$ may therefore be set-valued, and \eqref{eq:pnp-intro} is an inclusion.
This inclusion holds for any denoiser, whereas the energies and curvatures below concern Bayes denoisers.
For a general network the denoising residual in \eqref{eq:red-intro} need not be a gradient \cite{ReehorstSchniter2019}.
Diffusion-based posterior samplers use the residual at both points and do not solve an equilibrium equation.
They are outside \eqref{eq:red-intro}.

For a Bayes denoiser, the energy whose gradient is the prior force is described by the posterior covariance (\Cref{sec:object}).
We call the entropy--variance energy $R_t$ the \emph{induced energy}.
It is the minimum, over distributions with a prescribed mean, of their relative entropy to the prior plus their variance
divided by $2t$.
Its gradient is the prior force, and its curvature at matched input and output points is the posterior precision minus $t^{-1}I$.
The score side is its Moreau envelope.
The curvature criteria for the decisions studied below are therefore conditions on the posterior covariance $\Sigma_t$ of the
denoising channel; under observation, on its blocks along $\ker A$ and their Schur complements.
\Cref{tab:frame} in \Cref{sec:known} collects this known structure with its sources.

\begin{table}[t]
\centering\small
\caption{Three forces for one prior. For a Bayes denoiser of a smooth prior, all three agree as $t\to0$.}
\label{tab:three-forces}
\begin{tabular}{@{}lllll@{}}
\toprule
method & prior term & evaluated at & energy (Bayes denoiser) & $t\to0$ \\
\midrule
MAP & $\nabla g$ & --- & $g$ & --- \\
PnP family & $W_t$ & denoised output & $R_t$, curvature $\Sigma_t^{-1}-t^{-1}I$ & $\nabla g$ \\
RED / score & $\nabla E_t$ & noisy input & $E_t$, Moreau envelope of $R_t$ & $\nabla g$ \\
\bottomrule
\end{tabular}
\end{table}

\Cref{tab:three-forces} summarizes how the forces compare across scales.
For smooth priors the three forces agree as $t\to0$ (\Cref{prop:small-sigma}).
For $t>0$ they differ in general, and the difference depends on $t$.
For a Gaussian prior $W_t=\nabla g$ holds exactly at every $t$, whereas $\nabla E_t\neq\nabla g$.
The prior-force family is constant in $t$ only for Gaussian priors, a restatement of \cite[Thm.~F.1]{Liu2026TwoLevels}
(\Cref{thm:rigidity}).
Thus changing the scale of a non-Gaussian Bayes denoiser changes its prior force even at a fixed effective weight.
One exact member identifies the prior, but not stably: nearly identical prior forces can belong to substantially different priors
(\Cref{prop:prior-inverse-unstable}).

Existing theory explains much of this structure and its role in convergence.
MMSE denoisers are proximal maps of an energy \cite{Gribonval2011,GribonvalNikolova2021Bayes}.
That energy is weakly convex and admits an upper Moreau-envelope representation \cite{PritchardParhi2025}.
Denoisers are studied as resolvents of monotone operators \cite{PesquetRepettiTerrisWiaux2021}, and PnP and RED as
monotone inclusions \cite{SuzukiOno2026}.
The denoising residual is a gradient field only if the denoiser Jacobian is symmetric, and it is the gradient of the RED
penalty only under local homogeneity as well \cite{RomanoEladMilanfar2017,ReehorstSchniter2019}.
PnP iterations converge under nonexpansiveness or gradient-step structure
\cite{Ryu2019,HuraultLeclairePapadakis2022,HuraultChambolleLeclairePapadakis2024}.
For linear denoisers, the reconstructions converge to the solution minimizing a quadratic functional built from the denoiser
\cite{Hauptmann2024}.
In a non-variational framework, they converge to the solution singled out by the operator \cite{ObmannHaltmeier2026}.
Annealing the noise level of MMSE denoisers approaches the MAP \cite{PesmeMeantiArbelMairal2025}.
The scale derivative of the implicit regularizer also enters the convergence of annealed PnP \cite{Hurault2026annealed}.

These results mostly concern convergence at one noise level and characterize the limit through the denoiser itself.
They do not relate that limit to the decisions of $\pi$ or describe how it changes with the noise level.
How annealing affects which stationary point is selected is raised as an open question in \cite{Hurault2026annealed}.
The penalty of an MMSE denoiser is known to differ from $-\log\pi$ \cite{Gribonval2011}.
We study what that difference does to a reconstruction.

\begin{figure}[t]
\centering
\includegraphics[width=\textwidth]{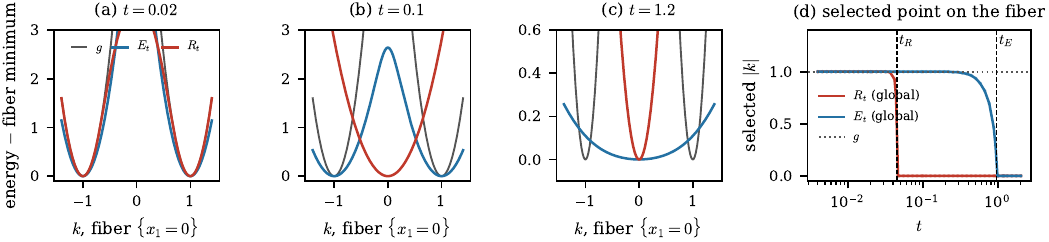}
\caption{A Bayes denoiser whose prior force selects an improbable reconstruction (\Cref{ex:two-mode}). Prior
$\pi=\tfrac12\mathcal N(-(1,1),0.05\,I)+\tfrac12\mathcal N((1,1),0.05\,I)$, measurement of the first coordinate, datum $0$. (a)--(c) Energies along the
fiber $\{x_1=0\}$: the prior energy $g=-\log\pi$, the smoothed energy $E_t=-\log p_t$ and the induced energy
$R_t$ with $\nabla R_t=W_t$, each minus its fiber minimum. (d) Global fiber minimizers against the noise variance $t$
(logarithmic axis). Dots mark all local minimizers. $R_t$ selects the midpoint, where the prior density is $9.1\cdot10^{-5}$ times
its fiber maximum, for every $t\ge t_R=0.0452$. The prior energy $g$ selects points near the modes.
The smoothed energy $E_t$ selects points away from the midpoint up to $t_E=0.95$.}
\label{fig:transition}
\end{figure}

\subsection*{Main results}

The normalization and the organization of the known structure around the family form the framework of the paper. Its main
contributions are results on the decisions of the family, from preference without observation to fiber decisions and recovery.
Energy preference does not guarantee that an iteration reaches a global minimizer.

\emph{Preference along the family.} Without observation, the induced energy and the smoothed energy $E_t=-\log p_t$
have the same critical points, with corresponding types, and the global minimizers of $R_t$ are the global maximizers of $p_t$,
a known fact (\Cref{thm:stationary-structure}). Its minimizers, and the order of their energy values, still change with $t$.
Any two nondegenerate branches of minimizers obey a variance-ordering law, also on a fiber:
the derivative in $t$ of their energy gap is minus the difference of their retained posterior variances divided by $2t^2$
(\Cref{thm:switching}).
As the scale grows, the energy of the branch with greater retained variance decreases faster, so under a persistent variance
order any switch of the lower-energy branch is toward the broader minimum.
The pointwise scale derivative underlying this law is due to \cite{Liu2026TwoLevels} for compactly supported priors.

\emph{Local decisions on a fiber.} On $S_y$ the PnP and score decisions differ in general, because the prior force evaluates the score at a point
displaced off the fiber by $t$ times the Lagrange multiplier.
Here $K$ is the unobserved subspace and $N$ its orthogonal complement.
PnP is locally stable on the fiber when the Schur complement $\Sigma_{K\mid N}$ of the posterior covariance is below $t$.
The score decision uses the marginal block $\Sigma_{KK}$ at its own evaluation point.
These give distinct curvature diagnostics: PnP uses the uncertainty left after linear prediction from the observed
directions, while the score decision uses marginal uncertainty in the unobserved directions.
Explicit coefficients describe the first-order separation of the three decisions in $t$ (\Cref{thm:fibre-selection}).

\emph{Separation from MAP.} For Gaussian priors, the prior force selects exactly the MAP, while the score force does not in the
example of \Cref{ex:gaussian-selection}.
For a symmetric two-mode prior and a fiber through the middle, \Cref{fig:transition} shows that the prior force selects the midpoint
for every $t$ above an explicit threshold, through a supercritical pitchfork (\Cref{ex:two-mode}).
Stability on the fiber therefore does not imply agreement with MAP.

\emph{The induced energy is, at leading order, the proximal hull of the MAP energy.} When the modes of the prior are far apart compared with the
noise level and their width, the induced energy is, at leading order, the proximal hull of the MAP energy with parameter $t$, and the
smoothed energy its Moreau envelope. The proximal hull is the largest minorant whose sum with $\norm\cdot^2/(2t)$ is convex.
For Gaussian mixtures this holds at every separation up to an explicit constant (\Cref{cor:hull}). The regime is not one of small
noise: the noise level may be comparable with, or larger than, the width of the modes.
The hull coincides with the MAP energy at the proximal points of the latter, which include its global minimizers, and lies
strictly below it elsewhere, typically on the barriers between modes.
The core case occurs when the minimum energies on the fiber agree: the limit preserves every MAP choice.
Otherwise every limiting decision lies on a bridge, where the hull is strictly below the prior energy. In the examples, bridges lie between modes, where the prior density is exponentially small.
PnP can therefore select a blend of modes even when the MAP selects a mode.

\emph{Height against mass.} Explicit component scores resolve the choice when the fiber passes through several modes
(\Cref{thm:mixture-selection,cor:mixture-offsets}).
We prove this for separated Gaussian mixtures with exponentially small errors and for priors with isolated nondegenerate global
modes far apart. In the latter case only the global modes are candidates, and their heights in the MAP energy may differ by bounded
amounts, small compared with the barriers between them.
PnP uses the MAP data fit and adds a penalty on narrow modes that grows with $t$.
As $t$ increases, the offsets change from ordering the modes by peak height to ordering them by mass.
In practice, the lowest scores determine a unique choice when their gap exceeds the theorem's error margin.

\emph{Consequences for recovery.} The prior force at the truth along $\ker A$ controls the error caused by the prior alone, locally
and on average over a class of truths (\Cref{thm:bias,rem:random-truth}).
Only fiber-critical truths can be recovered as the parameters vanish (\Cref{prop:only-critical}).
In the separated-mixture setting selecting a wrong mode causes an error of the order of the separation, and the bridge example gives a
persistent error even with exact data (\Cref{prop:wrong-core}).
The data residual is unchanged by displacements along $\ker A$ and therefore does not reveal this bias.

\emph{Choosing the noise level.} Under positive curvature along $\ker A$, the local error has a complete leading-order expansion
with scale bias $-tb$ determined by prior curvature.
Denoiser error contributes a force-error term amplified by $t^{-1}$.
In the oracle regime, balancing these effects fixes a noise level set by denoiser accuracy
(\Cref{thm:sharp-local-reconstruction,cor:sharp-scale}).
Thus decreasing the noise level need not improve reconstruction with an approximate denoiser.
The oracle benchmark requires the unknown truth.

\emph{Evidence.} Exact priors confirm the identities, coefficients and thresholds.
For image inpainting with an exact Gaussian-mixture prior, the component scores predict the induced energy's decisions on unseen
instances, including departures from MAP, while the predicted switching scales are less reliable (\Cref{sec:experiments}).

\subsection*{Relation to existing work}
We collect the known identities needed to compare decision rules.
\Cref{sec:known} states them with their sources: Tweedie's identity
\cite{Robbins1956,Miyasawa1961,Efron2011}, the proximal and envelope forms of MMSE denoisers
\cite{Gribonval2011,GribonvalNikolova2021Bayes,PritchardParhi2025}, the matched-point Hessian relation \cite{GribonvalMachart2013},
mean shift \cite{Cheng1995,ComaniciuMeer2002,CarreiraPerpinan2007} and covariance thresholds \cite{BardetGozlanMalrieuZitt2018}.

The distinction between PnP and denoising equilibria is observed in \cite{ReehorstSchniter2019}.
Observation-dependent posterior-mean penalties \cite{GribonvalMachart2013} address a different problem from transferring a fixed
denoising prior to a new measurement.
Suzuki and Ono \cite{SuzukiOno2026} formulate PnP and RED with firmly nonexpansive denoisers as monotone inclusions at a fixed
noise level.
They combine them into a hybrid regularizer with convergent splitting algorithms.
See also \cite{PesquetRepettiTerrisWiaux2021}.
Our analysis includes the nonconvex regime excluded by firm nonexpansiveness.

The dependence on scale links our analysis to work on posterior means and implicit regularizers.
For compactly supported priors, the scale derivative used in \Cref{sec:preference} is due to \cite{Liu2026TwoLevels}.
There it serves to show that no single regularizer reproduces the posterior means at two noise levels. That work also shows
that only Gaussian priors admit a common regularizer on a set of scales with an accumulation point \cite[Thm.~F.1]{Liu2026TwoLevels};
\Cref{thm:rigidity}(ii) restates this for the prior force.
The viscous Hamilton--Jacobi structure of posterior-mean estimators \cite{DarbonLanglois2021} and its nonconvex extension
\cite{DarbonLangloisMeng2023} are close antecedents of the low-temperature analysis of \Cref{sec:bridge}.
Related analyses treat MMSE PnP convergence \cite{XuMMSE2020,ParkEtAl2023}, prior mismatch \cite{ShoushtariEtAl2024},
implicit denoising priors \cite{KadkhodaieSimoncelli2021,LaumontEtAl2022}, learned proximal regularizers
\cite{FangBuchananSulam2024} and recovery with MMSE denoisers \cite{WolfFadiliGuoHe2026}.
A population-likelihood optimal regularizer is studied in \cite{LeongOReillySohChandrasekaran2025}.

The proximal hull, the Moreau envelope and quadratic envelopes are classical \cite{RockafellarWets1998,Carlsson2019}.
The Laplace limit of the smoothed energy is known \cite{DarbonLanglois2021,TibshiraniFungHeatonOsher2025}.
We have not found the statement that the induced energy of a Bayes denoiser is, at leading order, the proximal hull of the MAP energy.
It can, however, be assembled formally from these ingredients and the upper-envelope form of \cite{PritchardParhi2025}.
The scores of \Cref{thm:mixture-selection} use Gaussian component evidence constants familiar from restoration with mixture models
\cite{YuSapiroMallat2012}.

We connect this structure to selection across scales and under observation. \Cref{tab:results} lists the new results and
their relation to the closest known statements.

\begin{table}[t]
\centering\footnotesize
\caption{Results of this paper. ``New (statement)'': the statement is new, but it can be assembled formally from known ingredients,
which are cited next to it.}
\label{tab:results}
\setlength{\tabcolsep}{4pt}
\begin{tabularx}{\textwidth}{@{}>{\raggedright}p{3.6cm}>{\raggedright}X>{\raggedright}p{2.3cm}l@{}}
\toprule
result & statement & status & here \\
\midrule
variance-ordering law & gap derivative $-(V_t(u_1)-V_t(u_2))/(2t^2)$ along branches, on fibers & new; pointwise \cite{Liu2026TwoLevels} & \Cref{thm:switching} \\
fiber stability & Schur-complement criterion $\Sigma_{K\mid N}\prec tI_K$ & new & \Cref{thm:fibre-selection} \\
proximal hull & $R_t=h_t(G_\lambda)+o(\lambda^2)$; core or bridge & new (statement) & \Cref{cor:hull} \\
proximal hull, mixtures & $\abs{R_t-h_t(-\log\pi)}\le C$, every separation & new & \Cref{cor:hull} \\
height against mass & PnP score $=$ MAP score $+$ penalty on narrow modes & new & \Cref{thm:mixture-selection,cor:mixture-offsets} \\
selection bias & bounds by $P_KW_t(x^*)$, local and statistical & new & \Cref{thm:bias,rem:random-truth} \\
scale bias and oracle scale & error $-tb$ and force error, $t^*$ & new & \Cref{thm:sharp-local-reconstruction,cor:sharp-scale} \\
\bottomrule
\end{tabularx}
\end{table}

\subsection*{Organization}
We begin with the forces supplied by a denoiser: \Cref{sec:object} introduces the prior-force family and its known structure.
\Cref{sec:preference} establishes the baseline without observation.
\Cref{sec:selection} then shows how measurement fibers change the decisions.
\Cref{sec:recovery} relates these decisions to recovery of a truth.
\Cref{sec:experiments} reports numerical evidence, and \Cref{sec:conclusion} discusses open problems.
Known results are collected in \Cref{sec:known} with their sources; all other results are new unless a source is given.
Proofs not given after the statements are in the appendix. We study fixed points rather than the convergence of particular
algorithms, and the decisions the prior makes in $\ker A$ rather than its effect along observed
directions.

%% file: sections/s2_family.tex
\section{The prior-force family}\label{sec:object}

We first identify the forces that a denoiser supplies, relate them to the generating prior, describe the energy of an
exact Bayes denoiser, and ask how the prior force changes with the noise level. \Cref{sec:known} collects this known structure with its
sources; the new results begin in \Cref{sec:preference}.

\subsection{Setting and notation}\label{sec:model}

A denoiser is defined by the auxiliary Gaussian channel
\begin{equation}\label{eq:aux-channel}
 X\sim\pi,\qquad Z_t=X+\sqrt t\,G,\qquad G\sim\mathcal N(0,I_n)\ \text{independent of }X,
 \qquad t=\sigma^2>0 ,
\end{equation}
and used in the inverse problem
\begin{equation}\label{eq:ip}
 y^\delta=Ax^*+v,\qquad \norm v\le\delta,
\end{equation}
with $A:\mathbb R^n\to\mathbb R^m$ linear. The prior $\pi$ may be non-Gaussian, multimodal or singular. Here ``Gaussian'' refers
to the channel. The noise variance $t$, the data error $\delta$ and the weight $\alpha$ of the prior are independent
parameters.

The measurement separates the signal space into observed and unobserved directions. We write $K=\ker A$ and $N=K^\perp$,
and denote by $P_K$ and $P_N$ the orthogonal projectors onto $K$ and $N$. The \emph{measurement fiber}
$S_y=\{u:Au=y\}$ is the set of signals consistent with exact data $y$. For $y=Ax^*$ this set is $x^*+K$.

All quantitative results concern the family $D_t(z)=\mathbb E[X\mid Z_t=z]$, $t>0$, of MMSE
denoisers of one prior $\pi$. We call it the \emph{exact family} of $\pi$, and each of its members a \emph{Bayes denoiser}.
A trained network is a general Gaussian denoiser. Its maps $D_t$ acquire the structure of the exact family only through
accuracy hypotheses, since training with the square loss does not make a network an exact MMSE map.

For the exact family we use the density and posterior moments of the auxiliary channel. We write $p_t=\pi*\mathcal N(0,tI)$ for
the smoothed density, $E_t=-\log p_t$ for the \emph{smoothed energy} and $s_t=\nabla\log p_t$ for the score. When $\pi$ has a density,
$g=-\log\pi$ is the \emph{prior energy}. The \emph{posterior covariance} is $\Sigma_t(z)=\Cov(X\mid Z_t=z)$.
Tweedie's identity \cite{Robbins1956,Miyasawa1961,Efron2011} relates these quantities to the denoiser:
\begin{equation}\label{eq:tweedie}
 D_t(z)=z+t\,s_t(z),\qquad \nabla D_t(z)=\Sigma_t(z)/t .
\end{equation}

\emph{Notation.} We collect here the symbols used throughout.
\begin{itemize}
\item $\PP_2(\mathbb R^n)$ is the set of Borel probability measures on $\mathbb R^n$ with finite second moment. A measure has
 \emph{full affine span} if it is not supported in a proper affine subspace. $\mathcal N(m,C)$ is the Gaussian law with mean $m$ and
 covariance $C$.
\item For a law $q$, $\mathbb E_q[f(X)]=\int f(x)\,q(dx)$ is the expectation of $f(X)$ for $X\sim q$, and $m_q=\mathbb E_q[X]$ is its mean.
 $\mathrm{KL}(q\|\pi)$ is the relative entropy and $\mathrm{TV}(\mu,\nu)=\sup_B\abs{\mu(B)-\nu(B)}$, over Borel sets $B$, is the
 total-variation distance.
\item For a symmetric matrix $M\succ0$, $\norm v_M^2=v^\top Mv$. The symbols $\preceq$ and $\prec$ denote the Loewner order of
 symmetric matrices, and $\lambda_{\min}$, $\lambda_{\max}$ the extreme eigenvalues.
\item For a set-valued map $T$ on $\mathbb R^n$, $\dom T=\{x:T(x)\neq\emptyset\}$, $\ran T=\bigcup_xT(x)$ is its range,
 $\gph T=\{(x,w):w\in T(x)\}$ its graph and $\zer T=\{x:0\in T(x)\}$ its zero set. For a map $D$, $\Fix D=\{x:D(x)=x\}$.
 For a function $f$ with values in $\mathbb R\cup\{+\infty\}$, $\dom f=\{f<\infty\}$.
\item \emph{Inverses are taken in the sense of graphs throughout:} $D_t^{-1}(u)=\{z:D_t(z)=u\}$. For the exact family of a prior with
 full affine span $D_t$ is injective (\Cref{thm:bayes-structure}(ii)). For $u\in\ran D_t$ we then write $z=D_t^{-1}(u)$ for the unique preimage.
\item $C^k(\mathbb R^n)$ is the space of $k$ times continuously differentiable functions, and $C_b^k$ the subspace with bounded
 derivatives of orders $0$ to $k$. $\nabla^2$ is the Hessian, and $\Delta$ the Laplacian.
\item Limits $t\to0$ are taken through $t>0$.
\end{itemize}

\emph{Truth, prior and data.} The truth $x^*$ is a deterministic signal and is not assumed to be a draw from $\pi$. The
decisions compared in \Cref{sec:selection} are related to a deterministic truth in \Cref{sec:recovery}, within the framework of regularization theory.

\subsection{Known structure}\label{sec:known}

This subsection collects the known results used later, each with its source; \Cref{tab:frame} relates them to the literature on
denoiser-based reconstruction. They are scattered over different settings, mostly stated at a single noise level, and mostly used
for convergence proofs.

\subsubsection{Denoising graph, prior force and score force}\label{sec:graph}

Both forces are obtained from denoiser evaluations. Let $D_t:\mathbb R^n\to\mathbb R^n$, $t\in\mathcal I\subseteq(0,\infty)$, be a family of maps.
Each evaluation records a noisy input, a denoised output and the normalized denoising residual \eqref{eq:graph-intro}. Evaluated at the
noisy input, the residual gives the \emph{score force}
\begin{equation}\label{eq:score-force}
 \nabla E_t(z):=t^{-1}\bigl(z-D_t(z)\bigr),
\end{equation}
which equals $-s_t(z)=\nabla(-\log p_t)(z)$ for the exact family by \eqref{eq:tweedie}. For a general denoiser $\nabla E_t$ denotes the map
\eqref{eq:score-force}, which need not be a gradient \cite{ReehorstSchniter2019}. Evaluated at the denoised output, the same residual gives
the \emph{prior force}
\begin{equation}\label{eq:Wt}
 W_t:=t^{-1}\bigl(D_t^{-1}-I\bigr),\qquad
 \gph W_t=\bigl\{\bigl(D_t(z),\,t^{-1}(z-D_t(z))\bigr):z\in\mathbb R^n\bigr\},
\end{equation}
with $D_t^{-1}$ the graph inverse. Thus $D_t=(I+tW_t)^{-1}$ and
\begin{equation}\label{eq:inverse-response}
 z=u+t\,w\qquad\text{for every }(u,w)\in\gph W_t,\ u=D_t(z).
\end{equation}
Every forward evaluation thus gives a point of the graph of $W_t$ without an inverse computation. For a general denoiser $W_t$ may be
set-valued. We call $(W_t)_{t\in\mathcal I}$ the \emph{prior-force family}; indexing by the standard deviation gives the same family,
$W_{\sigma}=\sigma^{-2}(D_{\sigma}^{-1}-I)=W_{t=\sigma^2}$. We use $t$ in the theory and state the axis of every figure.

In monotone-operator terms, $D_t$ is the resolvent of $W_t$ and $\nabla E_t$ its Yosida approximation with parameter $t$
\cite{BauschkeCombettes2017,PesquetRepettiTerrisWiaux2021}; for the exact family, $E_t$ is the Moreau envelope of the induced energy
$R_t$ of \Cref{thm:bayes-structure} up to a constant. The two forces are related by
\begin{equation}\label{eq:W-score-t}
 \nabla E_t(z)\in W_t\bigl(D_t(z)\bigr),\qquad\text{and, for the exact family,}\qquad W_t(u)=-s_t\bigl(D_t^{-1}(u)\bigr),
\end{equation}
with equality in the inclusion when $D_t$ is injective, as for the exact family of a prior with full affine span. The two forces
then take the same values but are evaluated at different arguments; in general $W_t(u)\neq\nabla E_t(u)$.

The factor $t^{-1}$ removes the $O(t)$ size of the residual for a smooth prior, so $W_t\to\nabla g$ as $t\to0$
(\Cref{prop:small-sigma}): the prior does not disappear in this limit, although at $t=0$ itself the denoiser is the identity.

Each reconstruction scheme uses one of the two forces as its prior term, with an effective prior weight. Put
\begin{equation}\label{eq:Talpha}
 T_{\alpha,t}(x):=A^*(Ax-y^\delta)+\alpha W_t(x),
\end{equation}
so that \eqref{eq:pnp-intro} reads $0\in T_{\alpha,t}(x)$.

\begin{proposition}[Equation correspondence; after \cite{Buzzard2018,Ryu2019,RomanoEladMilanfar2017,HuraultChambolleLeclairePapadakis2024}]\label{prop:noisy-coordinates}
Let $D_t$ be any single-valued denoiser and $f(x)=\tfrac12\norm{Ax-y^\delta}^2$.
\begin{enumerate}
\item[(i)] \emph{(PnP family.)} The fixed points of PnP forward--backward splitting $x\mapsto D_t(x-\gamma\nabla f(x))$, and the $x$-components
 of the fixed points of PnP Douglas--Rachford splitting with data step $\gamma$, of PnP-ADMM with penalty $\rho$ and of the consensus
 equilibrium of $\prox_{\gamma f}$ and $D_t$ with equal weights, solve $0\in T_{\alpha,t}(x)$ with $\alpha=t/\gamma$ ($\alpha=\rho t$ for ADMM).
 The denoiser input at the fixed point is $z=x+tw$, $w\in W_t(x)$.
\item[(ii)] \emph{(Score formulation.)} Fixed points of RED with weight $\lambda$, $A^*(Ax-y^\delta)+\lambda\bigl(x-D_t(x)\bigr)=0$, and critical
 points of $f+\alpha E_t$ for the exact family, solve \eqref{eq:red-intro} with $\alpha=\lambda t$, respectively $\alpha$.
\item[(iii)] \emph{(Noisy coordinate.)} $z\mapsto D_t(z)$ is a bijection from the zeros of
 $\mathcal F_{\alpha,t}(z):=A^*\bigl(AD_t(z)-y^\delta\bigr)+\tfrac{\alpha}{t}\bigl(z-D_t(z)\bigr)$ onto the solutions of $0\in T_{\alpha,t}(x)$,
 with inverse $x\mapsto x-(t/\alpha)A^*(Ax-y^\delta)$; no invertibility of $D_t$ is needed.
\end{enumerate}
\end{proposition}
\noindent Thus all PnP schemes share one equation and RED has a different one; to compare methods, translate their step sizes or
penalties into the effective weight $\alpha$, which also changes when the noise level changes at fixed algorithmic parameters. These
are statements about fixed points, not convergence guarantees. By (iii), the PnP equation can be solved in the noisy coordinate with
forward evaluations only; if $D_t$ is $C^1$ with invertible Jacobian, $\nabla\mathcal F_{\alpha,t}(z)=[A^*A+\alpha\nabla W_t(D_t(z))]\nabla D_t(z)$,
with $W_t$ the local branch through $D_t(z)$. The proof expands the fixed-point conditions (\Cref{app:proofs-object}).
Equation \eqref{eq:pnp-intro} is the non-variational regularization of \cite{ObmannHaltmeier2026} with the operator $W_t$, and PnP and
RED as monotone inclusions of the same kind at one noise level are known \cite{PesquetRepettiTerrisWiaux2021,SuzukiOno2026}. Along $K$
the data term of \eqref{eq:pnp-intro} vanishes, so $P_KW_t(x)\ni0$ at every solution: among signals with the same measurements, the
reconstruction is determined by the prior force alone.

\subsubsection{The Bayes denoiser: induced energy and posterior covariance}\label{sec:bayes-slice}

For an exact Bayes denoiser the prior force is the gradient of an energy whose curvature is given by the posterior covariance.
Throughout this subsection $\pi\in\PP_2(\mathbb R^n)$ has full affine span.

\begin{proposition}[Structure of a Bayes denoiser; \cite{Gribonval2011,GribonvalNikolova2021Bayes,GribonvalMachart2013,PritchardParhi2025,Stam1959,Blachman1965}]\label{thm:bayes-structure}
Let $t>0$ and define
\begin{equation}\label{eq:entropic-logpartition}
 \Psi_t(z)\coloneqq t\log\mathbb E_\pi\Bigl[\exp\bigl((\inn zX-\norm X^2/2)/t\bigr)\Bigr],\qquad
 R_t(u)\coloneqq t^{-1}\bigl(\Psi_t^*(u)-\norm u^2/2\bigr).
\end{equation}
\begin{enumerate}
\item[(i)] \emph{(Induced energy.)} $\Psi_t$ is smooth and strictly convex, the MMSE denoiser is $D_t=\nabla\Psi_t=\prox_{tR_t}$, and $R_t$ is
 $t^{-1}$-weakly convex. For every $u\in\ran D_t$,
 \begin{equation}\label{eq:entropic-primal}
  R_t(u)=\min_{q:\,m_q=u}\Bigl\{\mathrm{KL}(q\|\pi)+\frac{1}{2t}\,\mathbb E_q\norm{X-u}^2\Bigr\},
  \qquad \nabla R_t(u)=W_t(u),
 \end{equation}
 and the minimizer is the posterior $q_{t,z}$ of $X$ given $Z_t=z$, at $z=u+tW_t(u)$.
\item[(ii)] \emph{(Curvature.)} $D_t$ is a diffeomorphism of $\mathbb R^n$ onto the open set $\ran D_t$, and at $u=D_t(z)$
 \begin{equation}\label{eq:hessian-g}
  \nabla D_t(z)=t^{-1}\Sigma_t(z)\succ0,\qquad \nabla W_t(u)=\nabla^2R_t(u)=\Sigma_t(z)^{-1}-t^{-1}I .
 \end{equation}
 In particular, $R_t$ is convex on an open convex subset of $\ran D_t$ if and only if $\Sigma_t\preceq tI$ at the corresponding noisy
 points.
\item[(iii)] \emph{(Smoothed energy.)} For $u\in\ran D_t$ and its preimage $z=D_t^{-1}(u)$,
 \begin{equation}\label{eq:three-energies}
  R_t(u)=E_t(z)-\frac{\norm{z-u}^2}{2t}-\frac n2\log(2\pi t),\qquad
  \min_{v}\Bigl\{R_t(v)+\frac{\norm{v-z}^2}{2t}\Bigr\}=E_t(z)-\frac n2\log(2\pi t),
 \end{equation}
 and the minimum is attained exactly at $v=D_t(z)$. Thus $E_t-\tfrac n2\log(2\pi t)$ is the Moreau envelope of the induced energy
 with parameter $t$.
\item[(iv)] \emph{(Posterior-averaged prior force.)} If $\pi=e^{-g}$ with $g\in C^1$ and $\mathbb E_\pi\abs{\nabla g(X)}<\infty$, then
 $W_t(D_t(z))=\nabla E_t(z)=\mathbb E[\nabla g(X)\mid Z_t=z]$ for every $z$.
\end{enumerate}
\end{proposition}
\noindent By \eqref{eq:entropic-primal}, $R_t(u)$ is the smallest value, over distributions $q$ with mean $u$, of the relative entropy
to the prior plus the variance of $q$ weighted by $1/(2t)$; the weight decreases as the noise level grows. A distribution concentrated near
several modes can have its mean $u$ between them; its relative entropy and variance give an upper bound for $R_t(u)$, in which the
variance enters with the weight $1/(2t)$. As $t$ grows, points between modes can therefore attain lower values of $R_t$, the
mechanism studied on fibers below. $R_t$ is referenced to $\pi$ but is not $-\log\pi$, and $e^{-R_t}$ is in general not the generating
law. Under the hypotheses of (iv), its force averages the prior gradient over the posterior and in general differs from the MAP
force at its output. The minimizing posterior is $q_{t,z}(dx)=\exp\bigl((\inn zx-\norm x^2/2-\Psi_t(z))/t\bigr)\pi(dx)$. If $\pi$ is compactly supported,
\eqref{eq:entropic-primal} holds as an extended-real identity on $\mathbb R^n$, and $R_t(u)\ge\dist(u,\supp\pi)^2/(2t)\ge0$.

By (ii), a posterior covariance eigenvalue above the channel variance gives negative curvature in that direction; on an affine subspace the
compression of $\Sigma_t^{-1}-t^{-1}I$ to its direction space decides instead (\Cref{thm:fibre-selection}(b)). The covariance is
available from the Jacobian of the exact denoiser at the matched input, and (iii) evaluates the induced energy from the smoothed
density there. Part (i) is the proximal representation of \cite{Gribonval2011,GribonvalNikolova2021Bayes} written through the Gibbs
variational principle, (ii) is in \cite{Gribonval2011,GribonvalMachart2013}, (iii) is the upper Moreau-envelope form of
\cite[Thm.~III.2]{PritchardParhi2025}, and (iv) is the classical identity for the score of a sum of independent variables
\cite{Stam1959,Blachman1965}; \Cref{app:proofs-object} records only the steps that differ in the present notation.

By (iii), the score force is the Yosida approximation of the prior force. Differentiating (iii) at matched points $u=D_t(z)$ gives
\begin{equation}\label{eq:congruence}
 \nabla^2R_t(u)=\Sigma_t(z)^{-1}-t^{-1}I=t\,\Sigma_t(z)^{-1/2}\,\nabla^2E_t(z)\,\Sigma_t(z)^{-1/2},
 \qquad \nabla^2E_t(z)=t^{-1}I-t^{-2}\Sigma_t(z),
\end{equation}
a congruence implicit in \cite{GribonvalMachart2013}. The curvature criteria for the decisions studied below are therefore conditions on
the posterior covariance; under observation, on its blocks along $\ker A$ and their Schur complements. For a trained denoiser $\nabla D_t$ need be neither symmetric
nor positive semidefinite \cite{Gribonval2011}, so an induced energy need not exist. The following routine consequence of the
heat-semigroup expansion prepares the comparison with MAP on a fiber.

\begin{lemma}[Small-scale expansion]\label{prop:small-sigma}
Suppose $\pi$ has a density $f=e^{-g}$ with $g\in C^6(\mathbb R^n)$ and $f\in C_b^6$. Then every compact set is contained in $\ran D_t$ for small $t$, and
as $t\to0$, uniformly on compact sets, $E_t=g+\tfrac t2\bigl(\Delta g-\abs{\nabla g}^2\bigr)+O(t^2)$ in $C^2$ and
\begin{equation}\label{eq:small-sigma}
\begin{aligned}
 W_t&=\nabla g+\tfrac t2\nabla\Delta g&&+O(t^2),\\
 \nabla E_t&=\nabla g+\tfrac t2\nabla\Delta g-t\,\nabla^2g\,\nabla g&&+O(t^2)
\end{aligned}
\qquad\text{in }C^1.
\end{equation}
\end{lemma}
\noindent The hypothesis holds for Gaussian mixtures and, by integrability, whenever $\nabla g,\dots,\nabla^6g$ are bounded (proof in
\Cref{app:proofs-object}). The two forces share the first-order term $\tfrac t2\nabla\Delta g$ and differ by $t\nabla^2g\,\nabla g$, which
vanishes where $\nabla g=0$: the infinitesimal form of their agreement without observation (\Cref{thm:stationary-structure}) and of their
disagreement on fibers (\Cref{thm:fibre-selection}), where a constrained MAP point need not be critical for $g$. As $t\to0$ both forces
tend to $\nabla g$, in agreement with the annealing of MMSE denoisers toward the MAP \cite{PesmeMeantiArbelMairal2025}: decreasing the
noise level brings the force closer to the MAP force, but does not remove the prior term.

\subsubsection{Agreement without observation}\label{sec:stationary}

Without observation the two energies have the same critical points and minimizers. We record the part of this classical result
used later;
\Cref{sec:selection} shows how it fails on a fiber.

\begin{proposition}[Agreement without observation; \cite{Cheng1995,CarreiraPerpinan2007}]\label{thm:stationary-structure}
Let $\pi\in\PP_2(\mathbb R^n)$ have full affine span and $t>0$.
\begin{enumerate}
\item[(i)] \emph{(Equilibria.)}
 $\{u\in\ran D_t:\nabla R_t(u)=0\}=\zer W_t=\Fix D_t=\{u:\nabla p_t(u)=0\}$, and every such $u$ satisfies
 $u=\mathbb E[X\mid Z_t=u]$.
\item[(ii)] \emph{(Types.)} Let $u_*$ be a common critical point whose posterior covariance $\Sigma_*=\Sigma_t(u_*)$ has no eigenvalue $t$.
 Then $u_*$ is a strict local minimizer of $R_t$ $\iff$ a strict local maximizer of $p_t$ $\iff$ $\Sigma_*\prec tI$ $\iff$ a linearly
 attracting fixed point of $u_{k+1}=D_t(u_k)$, with factor $\lambda_{\max}(\Sigma_*)/t$. Mixed signs of $\Sigma_*-tI$ give a saddle of both
 energies, and $\Sigma_*\succ tI$ a strict local maximizer of $R_t$ and minimizer of $p_t$.
\item[(iii)] \emph{(Preferences.)} Without any nondegeneracy, the local minimizers of $R_t$ are the local maximizers of $p_t$, and
 $\argmin R_t=\argmax p_t$.
\end{enumerate}
\end{proposition}
\noindent So without observation the induced and smoothed energies have the same equilibria, with corresponding types. Their
local minima are the modes of the smoothed density, which need not be modes of the prior. Part (ii) follows from \eqref{eq:congruence} with $z=u_*$ and gives a local
test for repeated denoising: compare the largest posterior covariance eigenvalue with the channel variance; if it is smaller, their
ratio is the linear convergence factor. Part (iii) follows from the envelope identity of \Cref{thm:bayes-structure}(iii)
\cite[Thm.~III.2]{PritchardParhi2025} by a standard envelope argument. For an empirical prior, $u_{k+1}=D_t(u_k)$ is Gaussian mean shift
\cite{FukunagaHostetler1975,Cheng1995,ComaniciuMeer2002}, with linear rate at a mode given in
\cite{CarreiraPerpinan2007,CarreiraPerpinan2015review}; \Cref{app:proofs-object} indicates the changes for a general prior in $\PP_2$. The
test is inconclusive in the degenerate case: for $\pi=\tfrac12(\delta_{-1}+\delta_1)$ and $t=1$, the origin has $\Sigma_*=t$ and
$R_1(u)-R_1(0)=u^4/12+O(u^6)$.

On a fiber the congruence changes the directions tested: the compression $P_K\nabla^2R_t(u)P_K$ corresponds through
\eqref{eq:congruence} to the compression of $\nabla^2E_t(z)$ to $\Sigma_t(z)^{-1/2}K$, not to $K$, so restriction to the fiber need not
preserve inertia. This is the mechanism of \Cref{sec:fibre-selection}.

\begin{table}[t]
\centering\footnotesize
\caption{Known structure used in this paper: concepts of the literature and the corresponding objects here.}
\label{tab:frame}
\setlength{\tabcolsep}{4pt}
\begin{tabularx}{\textwidth}{@{}>{\raggedright}X>{\raggedright}p{3.5cm}>{\raggedright}p{3.3cm}l@{}}
\toprule
concept in the literature & object here & source & here \\
\midrule
MMSE denoiser is proximal & $D_t=\prox_{tR_t}$ & \cite{Gribonval2011,GribonvalNikolova2021Bayes} & \Cref{thm:bayes-structure} \\
denoiser as resolvent of a monotone operator & $D_t=(I+tW_t)^{-1}$ & \cite{PesquetRepettiTerrisWiaux2021} & \Cref{sec:graph} \\
PnP and RED as monotone inclusions & \eqref{eq:pnp-intro}, \eqref{eq:red-intro} & \cite{SuzukiOno2026} & \Cref{prop:noisy-coordinates} \\
fixed points of PnP-ADMM, DRS, consensus equilibrium & \eqref{eq:pnp-intro}, effective $\alpha$ & \cite{Buzzard2018,Ryu2019} & \Cref{prop:noisy-coordinates} \\
non-variational regularization & \eqref{eq:pnp-intro} & \cite{ObmannHaltmeier2026} & \Cref{prop:noisy-coordinates} \\
upper Moreau envelope, weak convexity & $E_t$ Moreau envelope of $R_t$ & \cite{PritchardParhi2025} & \Cref{thm:bayes-structure} \\
RED regularizer, Jacobian symmetry & $\nabla E_t$, not a gradient in general & \cite{RomanoEladMilanfar2017,ReehorstSchniter2019} & \Cref{sec:graph} \\
Tweedie's formula, score & $\nabla E_t=-s_t$ & \cite{Robbins1956,Miyasawa1961,Efron2011} & \eqref{eq:tweedie} \\
score of a sum is a conditional expectation & $W_t(D_tz)=\mathbb E[\nabla g(X)\mid Z_t=z]$ & \cite{Stam1959,Blachman1965} & \Cref{thm:bayes-structure} \\
Gaussian mean shift & $\Fix D_t=\{\nabla p_t=0\}$ & \cite{Cheng1995,CarreiraPerpinan2007} & \Cref{thm:stationary-structure} \\
MMSE annealing toward the MAP & $W_t\to\nabla g$ as $t\to0$ & \cite{PesmeMeantiArbelMairal2025} & \Cref{prop:small-sigma} \\
common regularizer only for Gaussian priors & $W_t$ constant in $t$ only for Gaussian $\pi$ & \cite{Liu2026TwoLevels} & \Cref{thm:rigidity} \\
covariance thresholds, log-concavity & monotonicity of $W_t$ & \cite{BardetGozlanMalrieuZitt2018,SaumardWellner2014} & \Cref{rem:threshold} \\
scale derivative of the induced penalty & $\partial_tR_t=-V_t/(2t^2)$ & \cite{Liu2026TwoLevels} (compact support) & \Cref{lem:dtR} \\
viscous Hamilton--Jacobi equation, annealed PnP & scale dependence of $R_t$ & \cite{DarbonLanglois2021,Hurault2026annealed} & \Cref{lem:dtR} \\
Laplace limit, Moreau envelope & $\lambda^{-2}E_t(\lambda\,\cdot)\to e_tg$ & \cite{DarbonLanglois2021,TibshiraniFungHeatonOsher2025} & \Cref{cor:hull} \\
proximal hull, quadratic envelopes & $h_tg$, contact & \cite{RockafellarWets1998,Carlsson2019} & \Cref{lem:hull} \\
noise level as implicit parameter & the family $(W_t)_{t>0}$ & \cite{Hauptmann2024} & \Cref{sec:intro,sec:recovery} \\
\bottomrule
\end{tabularx}
\end{table}

\subsubsection{Dependence on the noise level}\label{sec:family-structure}

At a single scale every MMSE denoiser is the proximal map of an energy (\Cref{thm:bayes-structure}(i)). Across scales the energy
can stay fixed only for a Gaussian prior; in terms of the prior force this is the following known characterization.

\begin{proposition}[{Gaussian rigidity; (i) \cite{Moreau1965,Rockafellar1966}, (ii) \cite[Thm.~F.1]{Liu2026TwoLevels}}]\label{thm:rigidity}
\begin{enumerate}
\item[(i)] \emph{(Proximal families.)} If $D_t=\prox_{tg}$ for all $t>0$ with $g$ proper, lower semicontinuous and convex, then
 $W_t=\partial g$ for every $t$. Conversely, a family with $W_t\equiv W$ maximally monotone for all $t$ is of this form if and only if
 $W$ is cyclically monotone.
\item[(ii)] \emph{(Bayes families.)} Let $\pi\in\PP_2(\mathbb R^n)$ have full affine span, with exact family $D_t$. If $W_t$ is
 independent of $t$ on a nonempty open interval of scales, then $\pi$ is Gaussian. Conversely, if $\pi=\mathcal N(m,\Sigma)$
 with $\Sigma\succ0$, then $W_t(u)=\Sigma^{-1}(u-m)=\nabla g(u)$ for every $t>0$.
\end{enumerate}
\end{proposition}
\begin{proof}
(i) $\prox_{tg}=(I+t\partial g)^{-1}$ \cite{Moreau1965}, so \eqref{eq:Wt} gives $W_t=\partial g$. The converse is Rockafellar's
characterization of subdifferentials as the maximally cyclically monotone operators \cite{Rockafellar1966}.

(ii) Let $W_t=W$ for $t$ in an open interval $\mathcal I$, and fix $t_0\in\mathcal I$. Then $\ran D_t=\dom W=:\Omega$ for all $t\in\mathcal I$,
a connected set that is open by \Cref{thm:bayes-structure}(ii), and $\nabla R_t=W=\nabla R_{t_0}$ on $\Omega$ by \Cref{thm:bayes-structure}(i), so $R_t-R_{t_0}$ is a constant
$c(t)$ on $\Omega$. Since $\partial\Psi_t^*=(\nabla\Psi_t)^{-1}$, $\dom\partial\Psi_t^*=\Omega$ lies between the relative interior of
$\dom\Psi_t^*$ and $\dom\Psi_t^*$; as $\Omega$ is open, $\Omega=\operatorname{int}\dom\Psi_t^*$ \cite[Thm.~23.4, Cor.~23.5.1]{Rockafellar1970}.
The closed convex function $\Psi_t^*$ equals on the boundary of $\Omega$ its limits along segments from $\Omega$
\cite[\S7]{Rockafellar1970}, and it is $+\infty$ outside the closure of $\Omega$. Hence $R_t=R_{t_0}+c(t)$ on $\mathbb R^n$, and
$D_t=\prox_{tR_t}=\prox_{tR_{t_0}}$ for every $t\in\mathcal I$ by \Cref{thm:bayes-structure}(i). The function $R_{t_0}$ is proper and
lower semicontinuous, so \cite[Thm.~F.1]{Liu2026TwoLevels} shows that $\pi$ is Gaussian; its covariance is nonsingular because
$\pi$ has full affine span. Conversely, for $\pi=\mathcal N(m,\Sigma)$, $D_t(z)=m+\Sigma(\Sigma+tI)^{-1}(z-m)$, so
$D_t^{-1}(u)=u+t\Sigma^{-1}(u-m)$.
\end{proof}

\noindent Part (ii) restates \cite[Thm.~F.1]{Liu2026TwoLevels} for the prior force: a $t$-independent force gives a common regularizer,
to which that theorem applies. The theorem itself is more general: it allows any probability measure, a proper lower semicontinuous
regularizer $R$, convex or not, with $D_t(y)\in\prox_{tR}(y)$, and any set of scales with an accumulation point in $(0,\infty)$. In
particular, under full affine span a proper, lower semicontinuous and convex $g$ with $\prox_{tg}=D_t$ on an open interval of scales
exists only if $\pi=\mathcal N(m,\Sigma)$ and $g(u)=\tfrac12\inn{u-m}{\Sigma^{-1}(u-m)}$ up to a constant. For compactly supported priors
that are not point masses, two scales already exclude a common regularizer \cite[Thm.~2]{Liu2026TwoLevels}.

For the prior-force family this means that the noise level of a non-Gaussian Bayes denoiser is more than a step size: changing it
changes the prior force even at fixed effective weight, and the decisions studied below can change with $t$ (\Cref{sec:preference,sec:selection}).
For a non-Gaussian prior, such as a distribution of natural images, the implicit regularizer $R_t$, with $D_t=\prox_{tR_t}$, must change
with the noise level. Single-scale tests cannot establish that a family comes from one prior: the MMSE maps $D_t(z)=v_0(t)z/(v_0(t)+t)$ of
$\mathcal N(0,v_0(t))$ are Bayes optimal and proximal at every $t$, but come from one prior only if $v_0$ is constant.

Conversely, an exact prior force at one scale determines the prior. If two priors with full affine span have $W_t=W'_t$ on a nonempty
open set $O\subseteq\ran D_t\cap\ran D'_t$ for one $t$, then $D_t=D'_t$ on $D_t^{-1}(O)$, since $D_t^{-1}(u)=u+tW_t(u)$. By Tweedie's identity the smoothed densities are
proportional there; being positive and real analytic, they are proportional everywhere, hence equal, and the Fourier transform of
$p_t=\pi*\mathcal N(0,tI)$ determines $\pi$. This identification is not stable.

\begin{example}[One scale identifies the prior, but not stably]\label{prop:prior-inverse-unstable}
In one dimension let $\pi_0=\mathcal N(0,s^2)$ and $\pi_k\propto(1+\epsilon\cos(kx))\pi_0$, $0<\epsilon<1$, $k\in\mathbb N$.
Then for every $t>0$, with $c=s^2t/(2(s^2+t))$ and $r=s^2/(s^2+t)$,
\[
 \norm{D_{t,k}-D_{t,0}}_\infty\le\frac{\epsilon\,t\,r\,k\,e^{-ck^2}}{1-\epsilon},\qquad
 \norm{W_{t,k}-W_{t,0}}_\infty\le\frac{\epsilon\,k\,e^{-ck^2}}{1-\epsilon},
\]
while $\mathrm{TV}(\pi_k,\pi_0)\to\epsilon/\pi>0$ as $k\to\infty$. Hence no uniformly continuous inverse exists from single-scale prior forces in
the supremum norm to priors in total variation. The computation is in \Cref{app:proofs-object}.
\end{example}

Nearly identical denoisers can thus belong to substantially different priors: in the family above, at fixed noise
variance $t$, an oscillation of frequency $k$ changes the denoiser and the prior force only by $O(k\,e^{-ck^2})$. This is a second reason to study the family rather than one member.

We now ask which signals minimize the induced energy as the scale varies, first without observation in
\Cref{sec:preference}, before imposing measurement fibers in \Cref{sec:selection}.

%% file: sections/s3_preference.tex
\section{The family without observation: minimizers across scale}\label{sec:preference}

Without observation the local minimizers of the induced energy are the modes of the smoothed density
(\Cref{thm:stationary-structure}). These points, and the order of their energy values, change with the noise level. A scale identity
for the induced energy (\Cref{lem:dtR}) gives a law for the energy difference of two minimizers, with or without observation
(\Cref{thm:switching}); \Cref{rem:threshold} records when the induced energy is convex. Throughout, $\pi\in\PP_2(\mathbb R^n)$ has full affine span, so $\Sigma_t(z)\succ0$, $D_t$ is a
diffeomorphism onto the open set $\ran D_t$, and $R_t$, $W_t$ are smooth there.

\subsection{The variance-ordering law}\label{sec:switching}

Increasing the scale lowers the variance penalty in the representation \eqref{eq:entropic-primal}. The next identity quantifies
this.

\begin{lemma}[Scale derivative of the induced energy; \cite{Liu2026TwoLevels} for compact support]\label{lem:dtR}
Let $\pi\in\PP_2(\mathbb R^n)$ have full affine span. On the open set $\mathcal D=\{(t,u):t>0,\ u\in\ran D_t\}$,
$R_t(u)$ is smooth in $(t,u)$ and
\begin{equation}\label{eq:dtR}
 \partial_tR_t(u)=-\frac{V_t(u)}{2t^2},\qquad V_t(u):=\Tr\Sigma_t\bigl(D_t^{-1}(u)\bigr)=\mathbb E_{q_{t,z}}\norm{X-u}^2 ,
\end{equation}
where $\Sigma_t$ is the posterior covariance and $q_{t,z}$ the posterior at the preimage $z=D_t^{-1}(u)$.
\end{lemma}
\noindent We call $V_t(u)$ the \emph{retained variance} at $u$: the energy decreases faster where the posterior retains more
variance. It is available from the Jacobian of the exact denoiser at the matched input. By the envelope theorem, $\partial_t$ in
\eqref{eq:entropic-primal} acts only on the factor $1/(2t)$ (proof in \Cref{app:proofs-preference}). For compactly supported priors the
identity and its integrated form are due to \cite[Prop.~B.1, Lemma~1]{Liu2026TwoLevels}; here the prior needs only a finite second
moment and full affine span. It is a pointwise analog, with constrained mean, of the relation between mutual information and MMSE
(I-MMSE) \cite{GuoShamaiVerdu2005}, and has the same origin as the scale derivative of the implicit regularizer in annealed PnP
\cite{Hurault2026annealed} and the viscous Hamilton--Jacobi structure of posterior-mean estimators \cite{DarbonLanglois2021}.

We compare minimizers of $R_t$ on an affine set $S$. Without observation $S=\mathbb R^n$, and the minimizers are those of the induced
energy of the denoiser itself. With observation $S$ is a fiber $S_y$: a penalized PnP reconstruction is stationary for $R_t$ on its own
fiber, and a global minimizer of the penalized energy minimizes $R_t$ there (\Cref{prop:own-fibre}). Let $P$ be the orthogonal projector
onto the direction space of $S$ ($P=I$ or $P=P_K$). A \emph{branch of nondegenerate local minimizers} of $R_t$ on $S$ over an interval
$J$ is a continuous map $t\mapsto u(t)\in S\cap\ran D_t$ with $P\nabla R_t(u(t))=0$ and $P\nabla^2R_t(u(t))P|_{\ran P}\succ0$ for $t\in J$.

\begin{theorem}[The variance-ordering law]\label{thm:switching}
Let $\pi\in\PP_2(\mathbb R^n)$ have full affine span, and let $u_1,u_2$ be two branches of nondegenerate local minimizers of $R_t$
on $S$ over an open interval $J\subset(0,\infty)$. Then
\begin{equation}\label{eq:switching}
 \frac{d}{dt}\Bigl[R_t\bigl(u_1(t)\bigr)-R_t\bigl(u_2(t)\bigr)\Bigr]=-\frac{V_t(u_1(t))-V_t(u_2(t))}{2t^2}.
\end{equation}
Hence, if $u_1$ retains more posterior variance than $u_2$ throughout $J$, the preference between the two branches changes at most
once on $J$, from $u_2$ to $u_1$ as $t$ increases. If moreover every global minimizer of $R_t$ on $S$ lies on one of the two
branches, the same holds for the global decision on $S$. Each branch is $C^1$, and its velocity is an ascent direction of the retained
variance $V_t$ on $S$ \textup{(}\eqref{eq:branch-ode} below\textup{)}.
\end{theorem}
\begin{proof}
By \Cref{lem:dtR}, $R_t(u)$ is smooth in $(t,u)$ on $\mathcal D$ with $\partial_tR_t=-V_t/(2t^2)$. The map
$(t,u)\mapsto P\nabla R_t(u)$ on the affine space $S$ has invertible derivative in $u$ along the branches, so the
implicit function theorem and continuity make each branch locally the unique smooth solution, hence $C^1$.
Along a branch, $\frac d{dt}R_t(u_i)=\partial_tR_t(u_i)+\inn{\nabla R_t(u_i)}{\dot u_i}$, and the second term vanishes
because $\dot u_i\in\ran P$ while $P\nabla R_t(u_i)=0$. Subtracting the two branches gives \eqref{eq:switching}. If
$V_t(u_1)>V_t(u_2)$ on $J$, the gap $R_t(u_1)-R_t(u_2)$ is strictly decreasing, so it changes sign at most once, from positive to
negative; the statement on global decisions follows. Differentiating $P\nabla R_t(u_i(t))=0$ in $t$ gives
$P\nabla^2R_t\,\dot u_i+P\,\partial_t\nabla R_t=0$, and $\partial_t\nabla R_t=\nabla\partial_tR_t=-\nabla V_t/(2t^2)$ since $R$ is $C^2$ on
$\mathcal D$. Inverting the constrained Hessian gives \eqref{eq:branch-ode} below.
\end{proof}

The proof also gives the motion of each branch,
\begin{equation}\label{eq:branch-ode}
 \frac{du_i}{dt}=\frac1{2t^2}\bigl(P\nabla^2R_t(u_i)P|_{\ran P}\bigr)^{-1}P\,\nabla V_t(u_i):
\end{equation}
with $H_S=P\nabla^2R_t(u_i)P|_{\ran P}$, $\inn{P\nabla V_t}{\dot u_i}=(2t^2)^{-1}\inn{P\nabla V_t}{H_S^{-1}P\nabla V_t}\ge0$, with
equality only if $P\nabla V_t(u_i)=0$. As $t$ increases, a selected point thus moves in an ascent direction of the retained
variance at fixed $t$, at a speed set by the inverse curvature of $R_t$ on $S$.

As $t$ increases, the energy of the branch with greater retained posterior variance thus decreases faster, so any switch of
the lower-energy branch under a persistent variance order is from a narrow minimum to a broad one; the variances at the current
points give the sign of the change of the energy gap without solving at neighboring scales. Three consequences go beyond
the pointwise identity. \emph{Fibers:} the law holds on a measurement fiber, where constrained minimizers of $R_t$ need not be critical
points of $p_t$ (\Cref{thm:fibre-selection}); without observation, \eqref{eq:switching} is the logarithmic heat equation at modes, the
classical evolution of mode heights in Gaussian scale space. \emph{Annealing:} \eqref{eq:branch-ode} describes how a selected point
moves when the noise level is annealed. How annealing affects the selected stationary point is an open question in
\cite{Hurault2026annealed}; the variance order determines which of two branches has the faster-decreasing energy, though not which
branch an iteration follows. \emph{Switching between modes:} applied to the component scores of \Cref{cor:mixture-offsets}, the law gives the switching
criterion of \Cref{rem:switching-scores} and underlies the height-to-mass transition of \Cref{thm:mixture-selection}.

The theorem compares two given branches; it does not classify bifurcations of one branch, and the number of modes of $p_t$ need
not decrease with $t$ in dimension two and higher \cite{CarreiraPerpinanWilliams2003}. Energy preference alone does not make an
iteration leave its current basin.

\subsection{Convexity of the induced energy}\label{sec:threshold}

If the induced energy is convex on an affine set, every local minimizer there is global, so two branches of nondegenerate
minimizers cannot coexist and no switching occurs. By \eqref{eq:hessian-g}, $R_t$ is convex on an open convex subset of $\ran D_t$
exactly when $\Sigma_t\preceq tI$ at the corresponding noisy points; contractivity hypotheses in convergence theories for
denoiser-based reconstruction \cite{Hauptmann2024,Ebner2023,GiltonOngieWillett2021} are of this kind. We record the known facts about
the global bound.

\begin{remark}[Monotonicity along the family; \cite{BardetGozlanMalrieuZitt2018,WolfFadiliGuoHe2026,Behboodian1970,BakryGentilLedoux2014,SaumardWellner2014}]\label{rem:threshold}
Let
\[
 t_c(\pi):=\inf\bigl\{t_0>0:\ \sup_z\lambda_{\max}(\Sigma_{t}(z))\le t\ \text{for all }t>t_0\bigr\}
\]
be the threshold above which the covariance bound holds everywhere.
\begin{enumerate}
\item[(i)] \emph{(Compact support.)} If $\supp\pi\subseteq\bar B(m,R)$, then $\Sigma_t\preceq R^2I$, and for $t>R^2$ the map $D_t$ is a
 contraction with $\Lip(D_t)\le R^2/t$, $W_t$ is maximally monotone and $(R^{-2}-t^{-1})$-strongly monotone, and $R_t$ has a unique
 critical point \cite[\S2.1]{BardetGozlanMalrieuZitt2018}, \cite[Lem.~4.3]{WolfFadiliGuoHe2026}. This is sharp: for $n=1$ and
 $\pi=\tfrac12(\delta_{-R}+\delta_R)$, $D_t(z)=R\tanh(Rz/t)$ and on $(-R,R)$
 \begin{equation}\label{eq:two-point}
  W_t(u)=R^{-1}\operatorname{artanh}(u/R)-u/t,\qquad W_t'(u)=(R^2-u^2)^{-1}-t^{-1},
 \end{equation}
 so $W_t$ is strongly monotone if and only if $t>R^2$ \cite{Behboodian1970}.
\item[(ii)] \emph{(Log-concavity.)} $t_c(\pi)=0$ if and only if $\pi$ is log-concave, by the Brascamp--Lieb inequality
 \cite{BakryGentilLedoux2014}.
\item[(iii)] \emph{(Upward inheritance.)} If $\Sigma_t\preceq t(1+ct)^{-1}I$ for some $c\ge0$, the same holds at every $t'>t$, which is the
 preservation of strong log-concavity under Gaussian convolution \cite{SaumardWellner2014}. The set of $t$ at which $W_t$ is monotone
 is therefore an interval with left endpoint $t_c(\pi)$. This fails for families that are Bayes optimal for scale-dependent
 priors: the MMSE map of $\mathcal N(0,v_0(t))$ has $W_t(u)=u/v_0(t)$, whose modulus decreases wherever $v_0$ increases.
\end{enumerate}
\end{remark}

For reconstruction the global bound is a benchmark. Above $R^2$ the contraction hypotheses of \cite[Thm.~5]{Hauptmann2024} hold
automatically, but not its linearity hypothesis: $D_t$ has bounded range and a positive definite Jacobian, whereas a bounded affine
map is constant. For images the support radius is of order $\sqrt n$, so $R^2$ is far above any useful noise level.
On a fiber only the curvature along $\ker A$ matters (\Cref{thm:fibre-selection}(b)).

%% file: sections/s4_decisions.tex
\section{Decisions on measurement fibers}\label{sec:selection}

Data determine $Au$ up to noise; among signals with the same measurements, the choice along $K=\ker A$ is made by the prior
alone. This matters when $\ker A$ is large, as in inpainting,
super-resolution, compressed sensing and limited-angle tomography. We compare three decisions built from the same $\pi$,
\begin{equation}\label{eq:three-selections}
 \text{(R)}\ \min_{S_y}R_t,\qquad
 \text{(E)}\ \min_{S_y}E_t,\qquad
 \text{(G)}\ \min_{S_y}g,
\end{equation}
made by the induced energy of PnP, the smoothed energy of RED and score-based methods, and the prior energy of MAP.

These constrained problems are not reconstruction methods; they isolate the choice made in the unobserved directions. A method
minimizes $F_\Phi(u)=\tfrac12\norm{Au-y^\delta}^2+\alpha\Phi(u)$, $\Phi\in\{R_t,E_t,g\}$, whose critical points solve \eqref{eq:pnp-intro},
\eqref{eq:red-intro} and \eqref{eq:map-intro}; for PnP this is
\begin{equation}\label{eq:recon}
 0\in A^*(Au-y^\delta)+\alpha W_t(u)=T_{\alpha,t}(u).
\end{equation}
The link to the constrained problems is exact on the reconstruction's own fiber. Projecting onto $\ker A$ removes the data term,
so a PnP solution $\hat x$ satisfies $P_KW_t(\hat x)=0$, an equation without $\alpha$ or data (\Cref{prop:own-fibre}). The three
decisions generally lie on nearby but different fibers; comparing them on a common fiber $S_y$ isolates the choice along $\ker A$.
\Cref{thm:mixture-selection} treats the penalized problems directly.

\begin{lemma}[A penalized solution is a decision on its own fiber]\label{prop:own-fibre}
Let $\alpha>0$, let $\Phi$ be differentiable at $\hat x$, and let $\hat x$ be a critical point (a global minimizer) of
$\tfrac12\norm{Ax-y}^2+\alpha\Phi(x)$. Then
$\hat x$ is a critical point (a global minimizer) of $\Phi$ on the fiber $S_{A\hat x}$.
\end{lemma}
\begin{proof}
At a critical point $A^*(A\hat x-y)+\alpha\nabla\Phi(\hat x)=0$. Applying $P_K$ and using $P_KA^*=0$ gives $P_K\nabla\Phi(\hat x)=0$,
the stationarity condition on $S_{A\hat x}=\hat x+K$. If $\hat x$ is a global minimizer and $x\in S_{A\hat x}$, then
$\norm{Ax-y}=\norm{A\hat x-y}$, so the data terms agree and $\Phi(x)\ge\Phi(\hat x)$.
\end{proof}

Thus the effective weight can change which fiber a penalized reconstruction reaches, but within that fiber the reconstruction is
determined by the prior alone. The fiber analysis therefore applies to penalized solutions, whatever their measured values.

Even a Gaussian prior separates the PnP and score decisions on a fiber: in the next example the PnP decision equals the MAP point
for every noise level, while the score decision changes with $t$ in the unmeasured coordinate.

\begin{example}[Gaussian prior: PnP exact, score biased]\label{ex:gaussian-selection}
Let $\pi=\mathcal N(0,C)$ with $C=\begin{pmatrix}1&1\\1&2\end{pmatrix}$, $A=(1\ 0)$ and $y\in\mathbb R$.
By \Cref{thm:rigidity}(ii), $R_t=g+\mathrm{const}$ for every $t$.
Thus the (R)- and MAP decisions on $S_y$ coincide at $(y,y)$ for every $t$.
The smoothed density is $\mathcal N(0,C+tI)$, whose fiber maximizer is $(y,\,y/(1+t))$.
The (E)-decision is therefore biased by $-ty/(1+t)$ in the unmeasured coordinate.
This agrees with \Cref{thm:fibre-selection}(c), where $x_G=(y,y)$, $\nabla g(x_G)=(y,0)$ and $P_K\nabla^2g\,\nabla g(x_G)=(0,-y)$.
A nonzero multiplier alone does not separate the decisions: for $\pi=\mathcal N(0,I)$ all three coincide on every fiber.
\end{example}

\subsection{Local decisions on a fiber}\label{sec:fibre-selection}

We first explain locally why the decisions can differ. On a fiber two mechanisms separate the prior force from the score
force. The Lagrange multiplier displaces the point at which the score is evaluated. Restricting the Hessians to $\ker A$ also changes
their comparison, because compression does not commute with the congruence \eqref{eq:congruence}.

The curvature comparison requires a covariance that accounts for the observed directions.
For a symmetric positive definite $\Sigma$ on $\mathbb R^n=K\oplus N$ we use its blocks
$\Sigma_{KK}=P_K\Sigma|_K$, $\Sigma_{KN}=P_K\Sigma|_N$, $\Sigma_{NK}=P_N\Sigma|_K$ and $\Sigma_{NN}=P_N\Sigma|_N$, which are linear maps
between the subspaces $K$ and $N$. If $U_K$ and $U_N$ have orthonormal columns spanning $K$ and $N$, their matrices are
$U_K^\top\Sigma U_K$, $U_K^\top\Sigma U_N$, $U_N^\top\Sigma U_K$ and $U_N^\top\Sigma U_N$. The \emph{Schur complement}
\[
 \Sigma_{K\mid N}:=\Sigma_{KK}-\Sigma_{KN}\Sigma_{NN}^{-1}\Sigma_{NK}
\]
is a symmetric positive definite map on $K$. Inequalities such as $\Sigma_{K\mid N}\prec tI_K$ refer to the Loewner order on $K$ and
depend only on the subspaces $K$ and $N$, not on the choice of bases. It is the error covariance of the best linear prediction of the $K$ component from the $N$
component. When the law is Gaussian, it equals the conditional covariance.

We compare stationarity, local stability and displacement from MAP.
A point $u\in S_y$ is \emph{(R)-stationary} if $P_KW_t(u)=0$ and \emph{(E)-stationary} if $P_K\nabla E_t(u)=0$.
For $u\in\ran D_t$ we write $z=D_t^{-1}(u)$ for the noisy input from which the denoiser produces $u$, and $\Sigma_{K\mid N}(z)$ for
the Schur complement of the posterior covariance $\Sigma_t(z)$.

\begin{theorem}[Decisions on a fiber]\label{thm:fibre-selection}
Let $\pi\in\PP_2(\mathbb R^n)$ have full affine span, $t>0$, $K\neq\{0\}$, and $u\in S_y\cap\ran D_t$.
\begin{enumerate}
\item[(a)] \emph{(PnP evaluates the score off the fiber.)} $u$ is (R)-stationary if and only if
 \begin{equation}\label{eq:transfer}
  P_K\nabla E_t(z)=0,\qquad\text{and then}\qquad z-u=tW_t(u)\in N .
 \end{equation}
 Thus $z$ is displaced from $u$ in the observed directions only, and $z=u$ exactly when $u$ is a critical point of $p_t$.
\item[(b)] \emph{(Stability.)} An (R)-stationary $u$ is a nondegenerate local minimizer of $R_t$ on $S_y$ if and only if
 \begin{equation}\label{eq:schur-condition}
  \Sigma_{K\mid N}(z)\prec tI_K,
 \end{equation}
 whereas an (E)-stationary $u$ is a nondegenerate local minimizer of $E_t$ on $S_y$ if and only if $\Sigma_t(u)_{KK}\prec tI_K$.
 Since $\Sigma_{K\mid N}\preceq\Sigma_{KK}$, with equality if and only if $\Sigma_{KN}=0$, the PnP condition is the weaker one at a point
 with $z=u$: there every stable score decision is a stable PnP decision, while the converse can fail when $\Sigma_{KN}\neq0$
 (\Cref{ex:two-mode}).
\item[(c)] \emph{(Small noise.)} Let $\pi=e^{-g}$ satisfy the hypotheses of \Cref{prop:small-sigma}, and let $x_G$ be a
 nondegenerate local minimizer of $g$ on $S_y$, with $H_K=P_K\nabla^2g(x_G)|_K$. For small $t$, the PnP and score decisions near
 $x_G$ are unique nondegenerate local minimizers $x_R(t)$, $x_E(t)$ on $S_y$, and
 \begin{align}
  x_R(t)&=x_G-t\,b+O(t^2),\qquad b:=\tfrac12H_K^{-1}P_K\nabla\Delta g(x_G),\label{eq:xR}\\
  x_E(t)&=x_R(t)+t\,H_K^{-1}P_K\nabla^2g(x_G)\nabla g(x_G)+O(t^2).\label{eq:xE-xR}
 \end{align}
 In particular, the PnP and score decisions differ at first order in $t$ if and only if $P_K\nabla^2g(x_G)\nabla g(x_G)\neq0$.
\end{enumerate}
\end{theorem}
\begin{proof}
(a) By \eqref{eq:W-score-t}, $W_t(u)=\nabla E_t(z)$, and $z-u=tW_t(u)$ by \eqref{eq:inverse-response}. Hence
$P_KW_t(u)=0$ if and only if $P_K\nabla E_t(z)=0$, and then $W_t(u)\in N$. If $W_t(u)=0$, then $z=u$ and
\Cref{thm:stationary-structure}(i) applies; conversely $z=u$ forces $W_t(u)=0$.

(b) By \eqref{eq:congruence}, $P_K\nabla^2R_t(u)|_K=(\Sigma_t(z)^{-1})_{KK}-t^{-1}I_K$, and block inversion gives
$(\Sigma^{-1})_{KK}=\Sigma_{K\mid N}^{-1}$. For $E_t$ use $\nabla^2E_t=t^{-1}I-t^{-2}\Sigma_t$. The Hessians of $R_t$ and $E_t$ on $S_y$ at $u$
are therefore
\begin{equation}\label{eq:fibre-hessian}
 \Sigma_{K\mid N}(z)^{-1}-t^{-1}I_K\qquad\text{and}\qquad t^{-1}I_K-t^{-2}\Sigma_t(u)_{KK},
\end{equation}
which gives both criteria.

(c) By \eqref{eq:small-sigma}, uniformly in $C^1$ near $x_G$, $P_KW_t=P_K\nabla g+\tfrac t2P_K\nabla\Delta g+O(t^2)$ and
$P_K\nabla E_t=P_K\nabla g+\tfrac t2P_K\nabla\Delta g-tP_K\nabla^2g\,\nabla g+O(t^2)$, using
$\tfrac12\nabla\abs{\nabla g}^2=\nabla^2g\,\nabla g$. On $S_y=x_G+K$ both maps have derivative $H_K+O(t)$ at $x_G$, and
$P_K\nabla g(x_G)=0$; the implicit function theorem gives unique zeros $x_G+d$, $d\in K$, with
$H_Kd=-\tfrac t2P_K\nabla\Delta g(x_G)+O(t^2)$, respectively $H_Kd=-\tfrac t2P_K\nabla\Delta g(x_G)+tP_K\nabla^2g(x_G)\nabla g(x_G)+O(t^2)$.
The constrained Hessians are $H_K+O(t)\succ0$.
\end{proof}

The theorem separates two effects.
\emph{Where the score is evaluated} (a). $W_t(u)$ is the normal force of $\min_{S_y}R_t$, so the matched input $z$ lies on the parallel
fiber $S_{y+tAW_t(u)}$: PnP evaluates the score off its own fiber. For a computed PnP fixed point the displacement is observable,
$z-\hat x=-(t/\alpha)A^*(A\hat x-y^\delta)$ (\Cref{prop:noisy-coordinates}), that is, $t/\alpha$ times the negative gradient of the
data term. It is noted in
\cite[\S VI]{ReehorstSchniter2019}; part (a) gives its exact form on a fiber. A displacement alone does not separate PnP from RED
(\Cref{ex:gaussian-selection} with $C=I$).
\emph{Which curvature determines stability} (b). PnP stability uses the uncertainty left after linear prediction from the observed
directions, $\Sigma_{K\mid N}$; the score decision uses the marginal uncertainty $\Sigma_{KK}$. PnP can therefore decide stably on a fiber
where the prior is nonconvex and the score decision is unstable. The ratio $\lambda_{\max}(\Sigma_{K\mid N})/t$ is a local diagnostic of
ambiguity along $\ker A$; in \Cref{ex:two-mode} and in the image experiments it decreases as $t$ grows.
\emph{Small noise} (c). Both decisions differ from the MAP decision by $O(t)$, and they differ from each other at first order exactly
when $P_K\nabla^2g(x_G)\nabla g(x_G)\neq0$, that is, when the prior Hessian maps the normal force $\nabla g(x_G)$ of the MAP decision to a
vector with a nonzero component along the fiber. Part (c) is tested in \Cref{fig:mechanisms}(a), and $b$ is the scale bias of \Cref{thm:sharp-local-reconstruction}.

\subsection{An extreme example: a phase transition on a fiber}\label{sec:phase-transition}

The local mechanisms can also change the global decision. In the next model two separated modes lie along a measured
direction, and the data fix the measured coordinate halfway between them (\Cref{fig:transition}). It is an extreme case of the
Schur-complement criterion and of the bridge mechanism of \Cref{cor:hull}, not a general law.

\begin{example}[Two modes, one fiber through the middle]\label{ex:two-mode}
Let $n=2$, $a^2>\varepsilon>0$, $\pi=\tfrac12\mathcal N(-a(1,1),\varepsilon I)+\tfrac12\mathcal N(a(1,1),\varepsilon I)$,
$A=(1\ 0)$ and $y=0$, so that $S_0=\{(0,k)\}$, and put
\[
 t_R:=\frac{\varepsilon(a^2-\varepsilon)}{a^2+\varepsilon},\qquad t_E:=a^2-\varepsilon,\qquad
 \kappa_a(t):=\frac{a^2(\varepsilon-t)}{\varepsilon(\varepsilon+t)} .
\]
The following statements describe the transition. The proofs are in \Cref{app:proofs-selection}.
\begin{enumerate}
\item[(i)] Without observation, the origin is a nondegenerate saddle of both $R_t$ and $p_t$ for $0<t<2a^2-\varepsilon$.
\item[(ii)] \emph{(PnP.)} For $t\ge t_R$, $R_t|_{S_0}$ is strictly convex and the origin is its unique global minimizer.
 For $t<t_R$ it has exactly two global minimizers $(0,\pm as_t)$, where $s_t\in(0,1)$ is the positive root of
 $s=\tanh(\kappa_a(t)s)$. The selected point is continuous in $t$ and leaves the origin at $t_R$ in a supercritical pitchfork.
\item[(iii)] \emph{(Score.)} For $t<t_E$ the origin is a strict local minimizer of $p_t|_{S_0}$, and the global maximizers of
 $p_t|_{S_0}$ lie off the origin, symmetrically about it.
\item[(iv)] \emph{(MAP.)} The maximizers of $\pi|_{S_0}$ are $(0,\pm k_G)$, where $k_G\in(0,a)$ is the positive root of
 $k=a\tanh(ak/\varepsilon)$, and $\pi(0,0)/\pi(0,k_G)\le2e^{-k_G^2/(2\varepsilon)}$.
\item[(v)] \emph{(Consequence.)} For $t_R\le t<t_E$ the PnP decision on $S_0$ is the origin, the conditional mean of $\pi$ given
 $x_1=0$. There $p_t|_{S_0}$ has a strict local minimum, and the prior density is at most $2e^{-k_G^2/(2\varepsilon)}$ times its maximum
 on $S_0$.
\end{enumerate}
The mechanism is \Cref{thm:fibre-selection}(b): at the origin the posterior is elongated along the diagonal, so its Schur complement
is small while its marginal variance along the fiber is large. The fiber curvature of the induced energy depends on the former, that of the smoothed energy on the latter.
Low prior density does not mean large risk: the PnP decision is of mean type rather than mode type.
The change at $t_R$ is a symmetry-breaking bifurcation on the fiber, while the unconstrained structure (i) stays the same.
\end{example}

\subsection{The proximal hull as the leading-order induced energy}\label{sec:bridge}

When can a fiber minimizer of the induced energy lie between modes? We answer by relating it to a relaxation of the MAP energy.
For a Gaussian mixture we first give an exact finite-dimensional representation that makes such a blend explicit. Let
$\pi=\sum_{j=1}^Lw_j\mathcal N(m_j,C_j)$ with $L$ components (the letter $K$ is reserved for $\ker A$), $w_j>0$ and $C_j\succ0$, put
$M_j=C_j+tI$, and for $p$ in the probability simplex $\Delta_L=\{p\in\mathbb R^L:p_j\ge0,\ \sum_jp_j=1\}$ define
\begin{gather*}
 \bar M_p=\Bigl(\sum_jp_jM_j^{-1}\Bigr)^{-1},\qquad \bar C_p=\bar M_p-tI,\\
 \bar m_p=\bar M_p\sum_jp_jM_j^{-1}m_j,\qquad \delta_p=\tfrac12\sum_jp_j\norm{m_j-\bar m_p}^2_{M_j^{-1}},
\end{gather*}
the effective covariances, the precision-weighted barycenter of the smoothed components and its spread. The constants of the
$j$-th component in the prior, induced and smoothed energies are the \emph{component offsets}
\begin{equation}\label{eq:mixture-offsets}
\begin{gathered}
 c_j^G=-\log w_j+\tfrac12\log\det(2\pi C_j),\qquad
 c_j^R(t)=-\log w_j+\tfrac12\log\det(I+C_j/t),\\
 c_j^E(t)=-\log w_j+\tfrac12\log\det\bigl(2\pi(C_j+tI)\bigr),
\end{gathered}
\end{equation}
and we put $H_R(p)=\sum_jp_j\bigl(c_j^R(t)+\log p_j\bigr)$. Below, $M^\dagger$ denotes the Moore--Penrose pseudoinverse of a matrix $M$.

\begin{lemma}[Exact representation for Gaussian mixtures]\label{thm:simplex}
For every $u\in\mathbb R^n$
\begin{equation}\label{eq:simplex}
 R_t(u)=\min_{p\in\Delta_L}\Bigl\{\delta_p+\tfrac12\norm{u-\bar m_p}^2_{\bar C_p^{-1}}+H_R(p)\Bigr\},
\end{equation}
the minimum being attained uniquely at the posterior component probabilities $p_j(z)$, $z=D_t^{-1}(u)$. If $y\in\ran A$, then on the fiber $S_y$
\begin{equation}\label{eq:simplex-fibre}
 \min_{S_y}R_t=\min_{p\in\Delta_L}\Bigl\{\delta_p+\tfrac12r_p^\top(A\bar C_pA^\top)^\dagger r_p+H_R(p)\Bigr\},\qquad r_p=y-A\bar m_p,
\end{equation}
and every fiber minimizer is $\bar m_p+\bar C_pA^\top(A\bar C_pA^\top)^\dagger r_p$ for a minimizing $p$.
\end{lemma}
\noindent The lemma turns reconstruction on a fiber into an exact problem in the $L-1$ free proportions, whatever the signal
dimension, and the displayed formula then gives the reconstruction. Since the barycenter $\bar m_p$ may lie between the component
means, where every component density is small, the lemma computes a blend without assuming that the answer lies in a component
core. What distinguishes the induced energy is the nonlinear dependence on $p$: the smoothed and prior energies have analogous
representations in which the component quadratics enter linearly in $p$, so their leading-order minimum over $p$ is at a vertex.
The proof (\Cref{app:proofs-bridge}) combines \Cref{lem:R-sup} with the entropy duality
$-\log\sum_je^{-b_j}=\min_{p\in\Delta_L}\sum_jp_j(b_j+\log p_j)$ and a minimax interchange.

To describe the leading geometry of these decisions we use two classical regularizations of the prior energy.

\begin{lemma}[Proximal hull and Moreau envelope; \cite{RockafellarWets1998,Carlsson2019}]\label{lem:hull}
Let $f:\mathbb R^n\to\mathbb R$ be bounded below and $s>0$. The \emph{Moreau envelope} and the \emph{proximal hull} of $f$ with
parameter $s$ are, with $F^{**}$ the biconjugate of $F$, that is, the largest lower semicontinuous convex function below $F$,
\begin{equation}\label{eq:hull}
 e_sf(u)=\inf_x\Bigl\{f(x)+\frac{\norm{x-u}^2}{2s}\Bigr\},\qquad
 h_sf=\Bigl(f+\frac{\norm\cdot^2}{2s}\Bigr)^{**}-\frac{\norm\cdot^2}{2s}.
\end{equation}
\begin{enumerate}
\item[(i)] \emph{(Order.)} $e_sf\le h_sf\le f$, and $h_sf(u)=\sup_z\bigl\{e_sf(z)-\norm{z-u}^2/(2s)\bigr\}$. The hull is the largest
 function below $f$ whose sum with $\norm\cdot^2/(2s)$ is convex. It is in general not convex.
\item[(ii)] \emph{(Contact.)} $h_sf(u)=f(u)$ if and only if $u$ minimizes $x\mapsto f(x)+\norm{x-z}^2/(2s)$ for some $z$. If $f$ is
 differentiable at $u$, this holds if and only if $f(x)\ge f(u)+\inn{\nabla f(u)}{x-u}-\norm{x-u}^2/(2s)$ for all $x$. If $f\ge0$ is lower
 semicontinuous, then $\{h_sf=0\}=\{f=0\}$.
\item[(iii)] \emph{(Core or bridge.)} On an affine subspace $S$, $\inf_Sh_sf\le\inf_Sf$. If equality holds, every minimizer of $f$ on
 $S$ also minimizes $h_sf$ on $S$ (\emph{core}). If the inequality is strict, every minimizer of $h_sf$ on $S$ lies where $h_sf<f$
 (\emph{bridge}).
\end{enumerate}
\end{lemma}
\noindent Parts (i) and (ii) are classical \cite[Ex.~1.44, \S11]{RockafellarWets1998}. Part (iii) is the fiber analog of a property
of quadratic envelopes: a minimizer of the relaxed problem minimizes the original one where the envelope touches the function
\cite[Thm.~5.1]{Carlsson2019}. The relaxation leaves the global minimizers unchanged, and creates no new local minimizers, when the data term is weak compared with $1/s$
\cite[\S5]{Carlsson2019}; a hard fiber constraint is the opposite regime, which is why bridges occur. For a convex $f$ the hull is $f$.
Proofs, with rules for scaling and for bounded perturbations (\Cref{lem:hull-calc}), are in \Cref{app:proofs-bridge}.

\emph{Three length scales.} A multimodal prior has two lengths, the width of its modes and their separation; the denoiser adds a
third, $\sqrt t$. If $\sqrt t$ is small compared with the width, PnP decides as the MAP does (\Cref{prop:small-sigma},
\Cref{thm:fibre-selection}(c)); if it is comparable with the separation, the smoothed modes merge. We study the regime in between:
$\sqrt t$ comparable with or larger than the width, and the modes far apart compared with both. It is not a small-noise regime.

We model it by priors whose mode shapes and noise variance are fixed while the distances between the modes tend to infinity. For fixed $t>0$ let
\begin{equation}\label{eq:prior-family}
 \pi^\lambda(dx)=Z_\lambda^{-1}\rho_\lambda(x/\lambda)\,e^{-\lambda^2g(x/\lambda)}\,dx,\qquad
 Z_\lambda=\int\rho_\lambda(x/\lambda)\,e^{-\lambda^2g(x/\lambda)}\,dx,\qquad \lambda\ge1,
\end{equation}
where $g$ is continuous with $\min g=0$ and $g(v)\ge c_0\norm v^2-c_1$ for some $c_0>0$, and the measurable prefactor
$0<\rho_-\le\rho_\lambda\le\rho_+$ lets the heights of the modes differ by bounded amounts. The MAP energy is
$G_\lambda=-\log(d\pi^\lambda/dx)=\lambda^2g(\cdot/\lambda)-\log\rho_\lambda(\cdot/\lambda)+\log Z_\lambda$, and $R_t$, $E_t$ are the energies of
$\pi^\lambda$ at variance $t$. Near a global minimizer $v_k$ at which $g$ is twice differentiable with Hessian $H_k=\nabla^2g(v_k)\succ0$,
$\lambda^2g(x/\lambda)\approx\tfrac12(x-\lambda v_k)^\top H_k(x-\lambda v_k)$: the modes keep their shape while their distances grow like
$\lambda$, so $1/\lambda$ is the ratio of width to separation. The statements concern the leading order as $\lambda\to\infty$ with $t$
fixed, in the sense of Laplace's method.

The substitution $x=\lambda v$, $\varepsilon=\lambda^{-2}$ turns $\pi^\lambda$ into the law of $V=X/\lambda$,
\begin{equation}\label{eq:low-temp}
 \pi_\varepsilon(dv)=Z_\varepsilon^{-1}\rho_\varepsilon(v)\,e^{-g(v)/\varepsilon}\,dv,\qquad \rho_\varepsilon=\rho_\lambda,\qquad
 Z_\varepsilon=\int\rho_\varepsilon(v)e^{-g(v)/\varepsilon}dv=\lambda^{-n}Z_\lambda,
\end{equation}
a low-temperature family with MAP energy $G^\varepsilon=g/\varepsilon-\log\rho_\varepsilon+\log Z_\varepsilon$, whose denoiser is used at variance
$\varepsilon t$. The induced energy is invariant, $R_t(\lambda v)=R^\varepsilon(v)$, and $E_t(\lambda v)=E^\varepsilon(v)+n\log\lambda$, where
$R^\varepsilon$, $E^\varepsilon$ are the energies of $\pi_\varepsilon$ at variance $\varepsilon t$ (\Cref{app:proofs-bridge}). The proofs use
\eqref{eq:low-temp}; the statements use \eqref{eq:prior-family}.

The theorem compares the energies of \eqref{eq:prior-family} with the proximal hull and the Moreau envelope of the MAP energy.

\begin{theorem}[The proximal hull as the leading-order induced energy]\label{cor:hull}
Let $t>0$.
\begin{enumerate}
\item[(i)] \emph{(Leading order.)} For the family \eqref{eq:prior-family}, uniformly for $v$ in compact sets, as $\lambda\to\infty$,
 \[
  R_t(\lambda v)=h_t(G_\lambda)(\lambda v)+o(\lambda^2),\qquad E_t(\lambda v)=e_t(G_\lambda)(\lambda v)+o(\lambda^2).
 \]
\item[(ii)] \emph{(Decisions: core or bridge.)} For $\nu\in\ran A$, the minimizers of $R_t$ on $S_{\lambda\nu}$ exist, and after
 division by $\lambda$ all their accumulation points as $\lambda\to\infty$ minimize $h_tg$ on $\{Av=\nu\}$. Consequently, for the limit problems
 on $\{Av=\nu\}$: if $h_tg$ and $g$ have the same minimum value, every minimizer of $g$ also minimizes $h_tg$ (\emph{core}); otherwise
 every minimizer of $h_tg$ lies where $h_tg<g$ (\emph{bridge}). Among tied limit minimizers, lower-order terms decide.
\item[(iii)] \emph{(Gaussian mixtures, every separation.)} If $\pi=\sum_{j=1}^Lw_j\mathcal N(m_j,C_j)$ with $w_j>0$ and $C_j\succ0$, then
 on $\mathbb R^n$
 \begin{equation}\label{eq:hull-mixture-bound}
  \begin{gathered}
  \abs{R_t-h_t(-\log\pi)}\le\max_{i,j}\abs{c_i^R(t)-c_j^G}+\log L,\\
  \abs{E_t-e_t(-\log\pi)}\le\max_{i,j}\abs{c_i^E(t)-c_j^G}+\log L ,
  \end{gathered}
 \end{equation}
 with the offsets \eqref{eq:mixture-offsets}; the bounds do not depend on the means.
\end{enumerate}
\end{theorem}
\noindent In words: for separated modes, PnP uses at leading order the proximal hull of the MAP energy with parameter $t$, and the
score decision its Moreau envelope; equivalently $\lambda^{-2}R_t(\lambda v)\to h_tg(v)$ and $\lambda^{-2}E_t(\lambda v)\to e_tg(v)$. The dichotomy in (ii) is
\Cref{lem:hull}(iii) applied to $f=g$.

For the limiting potential $g$, the hull satisfies $h_tg=g$ exactly on the set of proximal points
$\bigcup_z\prox_{tg}(z)$, which contains the global minimizers of $g$, and $h_tg<g$ elsewhere, typically on the barriers between
modes. If the fiber contains a global minimizer of $g$, the minimizers of $h_tg$ on the fiber are the global minimizers of $g$ on
it, as for the MAP, and lower-order terms decide among them; otherwise PnP can select a blend. The set where $h_tg<g$ is
nondecreasing in $t$: the hull is close to the MAP energy for $\sqrt t$ small compared with the width, and close to its convex
envelope for $t$ large, as long as $\sqrt t$ stays small compared with the separation. When $\sqrt t$ is comparable with the
separation the theorem does not apply, but for Gaussian mixtures \eqref{eq:simplex-fibre} and \Cref{thm:fibre-selection}(a)(b) still
hold at every $t$.

Part (iii) complements (i) for Gaussian wells. Its bound holds at every separation and is informative for separated modes,
because the barriers grow with the square of the separation while the constants do not. Separated mixtures are instances of
\eqref{eq:prior-family}: for $\pi=\sum_jw_j\mathcal N(\lambda\mu_j,C_j)$, $\pi(x)=(2\pi)^{-n/2}\rho_\lambda(x/\lambda)e^{-\lambda^2\Gamma_\mu(x/\lambda)}$ with
\[
 \Gamma_\mu=\min_j\gamma_j,\qquad \gamma_j(v)=\tfrac12\norm{v-\mu_j}^2_{C_j^{-1}},\qquad
 \rho_\lambda=\sum_jw_j\det(C_j)^{-1/2}e^{-\lambda^2(\gamma_j-\Gamma_\mu)},
\]
and $\min_jw_j\det(C_j)^{-1/2}\le\rho_\lambda\le\sum_jw_j\det(C_j)^{-1/2}$. The piecewise quadratic $\Gamma_\mu$ is the leading-order MAP
energy. The proof of (iii) and \Cref{lem:hull-calc} give $\lambda^{-2}R_t(\lambda v)=h_t\Gamma_\mu(v)+O(\lambda^{-2})$ uniformly on $\mathbb R^n$,
with the same rate for $E_t$ and $-\log\pi$, and \Cref{lem:mixture-contact} decides explicitly where the hull touches $\Gamma_\mu$.

The proof is in \Cref{app:proofs-bridge}: a uniform Laplace principle for $\Psi_t$ and one conjugation give (i) and (ii), and
\Cref{thm:simplex} with the bounds of the log-sum-exponential gives (iii). The limit of the smoothed energy is the classical Laplace
(vanishing-viscosity) limit \cite{DarbonLanglois2021,TibshiraniFungHeatonOsher2025}, explicit for Gaussian mixtures in
\cite{DarbonLangloisMeng2023}. For the induced energy, \cite{DarbonLanglois2021} identify $R_t$ for log-concave priors and
\cite{PritchardParhi2025} write it as an upper Moreau envelope of $E_t$; neither treats separated modes with nonconvex $g$. Formally,
(i) combines that representation with the Laplace limit and \Cref{lem:hull}(i); the work lies in the uniformity.

\begin{remark}[Priors close to a mixture]\label{rem:perturbation}
The energies are stable under bounded changes of the density. If a second prior $\pi'$ satisfies $c_1\le d\pi'/d\pi\le c_2$ with
constants $0<c_1\le c_2$, its energies satisfy on $\mathbb R^n$
\[
 -\log c_2\le R'_t-R_t\le-\log c_1,\qquad -\log c_2\le E'_t-E_t\le-\log c_1 .
\]
Indeed $t\log c_1\le\Psi'_t-\Psi_t\le t\log c_2$ in \eqref{eq:entropic-logpartition}, conjugation reverses these bounds, and
$c_1p_t\le p'_t\le c_2p_t$. The MAP energies differ by the same interval, and so do their hulls by \Cref{lem:hull-calc}. Since
$c_1\le1\le c_2$, \Cref{cor:hull}(iii) therefore extends to every prior whose density lies within bounded factors of a Gaussian
mixture, with the constants in \eqref{eq:hull-mixture-bound} increased by $\log(c_2/c_1)$.
\end{remark}

\emph{Scope for images.} The family \eqref{eq:prior-family} is an analytic device, not a model of images. It contains separated
Gaussian mixtures, which are established patch priors \cite{ZoranWeiss2011,YuSapiroMallat2012}, and image data appear to lie on a
disconnected union of sets of different intrinsic dimensions \cite{BrownEtAl2023}, the kind of multimodality it describes. Priors
concentrated near low-dimensional manifolds \cite{PopeEtAl2021,FeffermanMitterNarayanan2016} are covered by \Cref{cor:hull}(i)(ii)
when they satisfy the assumptions of \eqref{eq:prior-family}, since isolated minimizers are not required, but not by \Cref{thm:mixture-selection} under \Cref{ass:compatible}(b); this is the second
open problem of \Cref{sec:conclusion}. Heavy-tailed components violate the quadratic growth. The image prior of \Cref{sec:exp-gmm} is a
fitted mixture outside the asymptotic regime. That the scores still predict most of its decisions indicates that the mechanism acts at
moderate separations, not that the regime describes real images.

\begin{remark*}
The dichotomy of \Cref{lem:hull}(iii) concerns the limit problem. For a mixture at finite separation every optimal $p$ in
\eqref{eq:simplex} is interior. The decision is given by the exact problem \eqref{eq:simplex-fibre}.
\end{remark*}

The simplex representation also recovers \Cref{ex:two-mode}: with $s=p_+-p_-$, \eqref{eq:simplex-fibre} becomes one-dimensional with
stationarity condition $\operatorname{atanh}s=\kappa_as$. At $t=0.0418$ and $0.0439$ in \Cref{fig:transition} ($a=1$,
$\varepsilon=0.05$), its root gives $\abs k=as_t=0.93$ and $0.75$, as computed (\Cref{sec:exp-selection}).

A bridge need not lie at the midpoint. For $\mu_\pm=\pm(1,1)$, $C_-=\tfrac12I$, $C_+=\tfrac32I$, equal
weights, $t=2$ and the fiber $\{v_1=0\}$, the leading problem $\min_{\{v_1=0\}}h_t\Gamma_\mu$ reduces to $\min_m\ell(m)$ with
$\ell(m)=\frac{1-m^2}{3+m/2}+\frac{m^2}{2(1+m/2)}$.
Its unique minimizer is $m_*=0.23595$, with minimum value $0.32776$, below the minimum $\tfrac13$ of $\Gamma_\mu$ on the fiber.
At leading order the induced energy selects the bridge $(0,\lambda m_*)$, while the MAP and the smoothed energy select $(0,\lambda)$.

The analysis thus has two orders. At order $\lambda^2$ the hull $h_t\Gamma_\mu$ determines whether the decision is a core or a
bridge. When the fiber
passes through compatible cores, $\nu=A\mu_j$ for some $j$, the leading minimum is zero and is attained at those cores, and the
order-one terms decide among them. We study this choice next.

\subsection{Decisions among compatible modes: height against mass}\label{sec:mixtures}

Modes that remain compatible with the data are tied at leading order. Explicit scores, which add a data fit and an offset that
depends on the decision rule, resolve this tie. Let
\begin{equation}\label{eq:mixture}
 \pi_\lambda=\sum_{k=1}^Lw_k\,\mathcal N(m_k,C_k),\qquad m_k=\lambda\mu_k ,
\end{equation}
with $L\ge2$ components fixed, $w_k>0$, distinct $\mu_k$ and fixed $C_k\succ0$.
Write $c_\pm$ for the extreme eigenvalues of the $C_k$ and $\delta_\mu=\min_{j\ne k}\abs{\mu_j-\mu_k}$.

The scores come from the reconstruction problem for each component considered separately.
For a Gaussian $\mathcal N(m,C)$ the induced energy is exact,
\begin{equation}\label{eq:gaussian-R}
 R_t(u)=\tfrac12(u-m)^\top C^{-1}(u-m)+\tfrac12\log\det(I+C/t),
\end{equation}
since the minimizing law in \eqref{eq:entropic-primal} is $\mathcal N(u,tC(C+tI)^{-1})$. For $s\ge0$ write
$S_k=C_k+sI$, $r_k=y-Am_k$, and
\begin{equation}\label{eq:mixture-fit}
 Q_k^{(s)}(y)=\tfrac\alpha2\,r_k^\top\bigl(AS_kA^\top+\alpha I\bigr)^{-1}r_k,\qquad
 x_k^{(s)}=m_k+S_kA^\top\bigl(AS_kA^\top+\alpha I\bigr)^{-1}r_k ,
\end{equation}
the minimum value and minimizer of $\tfrac12\norm{Au-y}^2+\tfrac\alpha2(u-m_k)^\top S_k^{-1}(u-m_k)$. The component offsets
$c_k^G$, $c_k^R(t)$ and $c_k^E(t)$ are those of \eqref{eq:mixture-offsets}, and we set $s_G=s_R=0$, $s_E=t$.
We call $Q_k^{(s_\Phi)}+\alpha c_k^\Phi$ the \emph{score} of mode $k$ for the decision $\Phi$.
Here a score is a scalar objective value used to rank modes, distinct from the density score used to define the score force.

The next theorem treats separated Gaussian mixtures, with exponentially small errors, and general priors with isolated
nondegenerate global modes at low temperature, where the same scores use the local mass, location and curvature of each mode.

\begin{assumption}[Compatible modes]\label{ass:compatible}
One of the following holds.
\begin{enumerate}
\item[(a)] \emph{(Separated Gaussian mixtures.)} $\pi=\pi_\lambda$ is as in \eqref{eq:mixture}, and ranges $0<t_0\le t_1$, $\alpha_1>0$
 and $B>0$ are fixed; $t\in[t_0,t_1]$, $0<\alpha\le\alpha_1$ and $y\in\ran A$. A nonempty set $J$ of components is compatible with the
 data, $\abs{y-Am_k}\le B$ for $k\in J$, the other components are excluded by the data in the sense of \eqref{eq:mixture-exclusion},
 and $\lambda\ge\lambda_0$, with $\lambda_0$ given in \eqref{eq:lambda0-eps}.
\item[(b)] \emph{(Global modes at low temperature.)} $\pi=\pi_\varepsilon$ is the rescaled family \eqref{eq:low-temp}, and $\varepsilon$ is
 small. The function $g$ has finitely many global minimizers $x_k$ and is $C^3$ near each of them with $H_k=\nabla^2g(x_k)\succ0$, and
 $\log\rho_\varepsilon$ is bounded in $C^2$ near each $x_k$, uniformly in $\varepsilon$ (it encodes height differences of order
 $\varepsilon$). Further, for constants $\tau>0$ and $a_1>0$ fixed independently of $\varepsilon$, $t=\varepsilon\tau$, $0<\alpha\le a_1\varepsilon$, $y=y_0+\sqrt\varepsilon\,b_\varepsilon$ with $b_\varepsilon$ bounded,
 and $J=\{k:Ax_k=y_0\}\neq\emptyset$. The components are $m_k=x_k$, $C_k=\varepsilon H_k^{-1}$ and $w_k=\pi_\varepsilon(U_k)$ for fixed small
 disjoint balls $U_k\ni x_k$, and the prior energy is $G^\varepsilon$.
\end{enumerate}
\end{assumption}

\begin{theorem}[Decisions among compatible modes]\label{thm:mixture-selection}
Under \Cref{ass:compatible}, let $\Phi$ be the prior energy, $R_t$ or $E_t$, and let $S_k^\Phi=Q_k^{(s_\Phi)}+\alpha c_k^\Phi$ be the score of
mode $k$. There are error bounds $\theta$ and $\ell$, with $\theta/\alpha$ and $\ell$ exponentially small in $\lambda^2$ in setting (a) and of
order $\sqrt\varepsilon$ and $\varepsilon$ in setting (b), such that the following hold.
\begin{enumerate}
\item[(i)] \emph{(Local decisions.)} For each compatible mode $k\in J$, $F_\Phi$ has exactly one minimizer $v_{k,\Phi}$ on a ball about $m_k$,
 and it is a strict local minimizer. It lies within $\ell$ of the component reconstruction $x_k^{(s_\Phi)}$, and $F_\Phi(v_{k,\Phi})$ lies
 within $\theta$ of the score $S_k^\Phi$.
\item[(ii)] \emph{(Global decision.)} Every global minimizer of $F_\Phi$ is one of the $v_{k,\Phi}$, and its score is within $2\theta$ of
 $\min_{j\in J}S_j^\Phi$. In particular, if one score is smaller than all others by more than $2\theta$ (by more than $\theta$ in setting (a)), its
 mode is selected and the global minimizer is unique.
\item[(iii)] \emph{(Hard constraint.)} The same holds for $\min_{S_y}\Phi$ (in (b) for $y\in\ran A$), with the scores and component
 reconstructions replaced by their limits as $\alpha\to0$,
 \[
  \bar S_k^\Phi=\lim_{\alpha\downarrow0}\frac{S_k^\Phi}\alpha=\tfrac12r_k^\top(AS_kA^\top)^\dagger r_k+c_k^\Phi,\qquad
  \bar x_k^{(s_\Phi)}=\lim_{\alpha\downarrow0}x_k^{(s_\Phi)}=m_k+S_kA^\top(AS_kA^\top)^\dagger r_k,
 \]
 with $S_k=C_k+s_\Phi I$, and $\theta$ replaced by $\theta/\alpha$.
\end{enumerate}
\end{theorem}
\noindent The theorem reduces global selection to a comparison of explicit scores and locates the reconstruction within the
selected component. In practice, compare the lowest scores and check their gap against the margin of (ii) (divided by $\alpha$ for the hard constraint)
before inferring a unique choice.
The theorem concerns global minimizers; an algorithm can stop at the local minimizer $v_{k,\Phi}$ of another compatible mode. Several
components remain compatible as $\lambda\to\infty$ only if $A\mu_j=A\mu_k$, so the theorem concerns modes that differ along $\ker A$. The
constants are far from sharp; in examples the scores are exact already at moderate separation.

Setting (b) is stated for \eqref{eq:low-temp}, where the proof works; since
$D^{\pi^\lambda}_\tau(\lambda\zeta)=\lambda D^{\pi_\varepsilon}_{\varepsilon\tau}(\zeta)$, the variance $t=\varepsilon\tau$ there corresponds to the fixed
variance $\tau$ for $\pi^\lambda$. The main step of the proofs (\Cref{app:proofs-mixtures,app:proofs-general}) controls where the
denoiser input lies, not just its output. In (a), for $u$ near the core of component $k$, the input $z=D_t^{-1}(u)$ stays where
component $k$ dominates the posterior, and the three-energy identity gives
\begin{equation}\label{eq:mixture-exact}
 \begin{aligned}
 &R_t(u)-\Bigl[\tfrac12(u-m_k)^\top C_k^{-1}(u-m_k)+c_k^R(t)\Bigr]\\
 &\qquad=-\log\Bigl(1+\sum_{j\ne k}\rho_{jk}(z)\Bigr)-\tfrac1{2t}(z-z_k)^\top C_k(C_k+tI)^{-1}(z-z_k),
 \end{aligned}
\end{equation}
with $\rho_{jk}$ the posterior odds of component $j$ against $k$ and $z_k=m_k+(I+tC_k^{-1})(u-m_k)$ the single-component inverse of
$u$. In (b) the modes are not exactly Gaussian, the errors are $O(\sqrt\varepsilon)$, and
$D_t^{-1}(x_k+\sqrt\varepsilon\xi)=x_k+\sqrt\varepsilon(I+\tau H_k)\xi+O(\varepsilon)$.

Setting (b) makes the role of height explicit. The local masses may be replaced by their Laplace values
$w_k\propto\rho_\varepsilon(x_k)\det(2\pi C_k)^{1/2}$ without changing the error order, so the MAP offset is
$-\log\rho_\varepsilon(x_k)+\mathrm{const}$. For a fixed Gibbs potential, $\rho_\varepsilon\equiv1$, all global wells have the same height and
the MAP decides by data fit alone, whereas PnP adds $\tfrac\alpha2\log\det(I+\tau H_k)$, which is increasing in $H_k$ in the Loewner order
(\Cref{cor:mixture-offsets}). Height differences come only from $\rho_\varepsilon$; they are of order one in $G^\varepsilon$, that is, of order $\varepsilon$
relative to $g/\varepsilon$, while the masses also depend on the curvatures through $\det(H_k)^{-1/2}$. The restriction to global modes is essential. A higher well $x_k$ need not be a
proximal point of $g$: if $g(x)+\norm{x-x_k}^2/(2\tau)<g(x_k)$ for some $x$, which can happen at another well when $\tau$ is large, then
$h_\tau g(x_k)<g(x_k)$ by \Cref{lem:hull}(ii), and a single-well score for $x_k$ fails at leading order.

Among compatible modes, PnP uses the data fit of the MAP but penalizes narrow modes by an amount growing with $t$. As $t$ grows
its score approaches $Q_k^{(0)}-\alpha\log w_k$, which selects the mode of largest mass when the compatible modes fit the data
equally well. Raising the noise level can therefore switch the
reconstruction to another mode, in a predictable direction. The next corollary makes this explicit.

\begin{corollary}[Three scores]\label{cor:mixture-offsets}
Up to constants common to all components:
\begin{enumerate}
\item[(i)] the PnP score is the MAP score plus a penalty on narrow modes,
 \begin{equation}\label{eq:offset-E}
  S_k^{R}=S_k^{G}+\tfrac\alpha2\log\det\bigl(I+tC_k^{-1}\bigr),
 \end{equation}
 while the score decision uses the same offsets as PnP with the widened data fit $Q_k^{(t)}$;
\item[(ii)] every strict ordering of the components by height $w_k\det(2\pi C_k)^{-1/2}$, as used by the MAP, is the ordering of the
 PnP offsets for small $t$, and as $t$ grows the PnP offsets order the components by mass: $c_k^R(t)=-\log w_k+\Tr C_k/(2t)+O(t^{-2})$;
\item[(iii)] if exactly two modes are compatible and their MAP and PnP score differences both exceed $2\theta$ in absolute value, the MAP
 and PnP decisions select different modes when the two differences have opposite signs, and the same mode otherwise.
\end{enumerate}
\end{corollary}
\begin{proof}
(i) $\tfrac12\log\det(I+C_k/t)=\tfrac12\log\det(2\pi C_k)+\tfrac12\log\det(I+tC_k^{-1})-\tfrac n2\log(2\pi t)$, the last term is common to
all components, and $S_k^R$ and $S_k^G$ share the fit $Q_k^{(0)}$. (ii) By the same identity $c_k^R(t)=c_k^G+O(t)$ up to a common constant
as $t\to0$; the expansion as $t\to\infty$ is direct. (iii) By \Cref{thm:mixture-selection}(ii), every global minimizer belongs to
$J=\{1,2\}$, and a mode whose score exceeds the other by more than $2\theta$ is not selected.
\end{proof}

For the score decision, $\alpha c_k^E(t)=\alpha c_k^R(t)+\tfrac{\alpha n}2\log(2\pi t)$, and its data fit $Q_k^{(t)}$ uses the covariance
$C_k+tI$. In setting (b), $tC_k^{-1}=\tau H_k$. The PnP offsets decrease monotonically in $t$,
\begin{equation}\label{eq:offset-derivative}
 \frac{d}{dt}c_k^R(t)=-\frac{V_k(t)}{2t^2},\qquad V_k(t)=\Tr\bigl(tC_k(C_k+tI)^{-1}\bigr),
\end{equation}
where $V_k(t)$ is the variance that component $k$ retains in the channel.

The curvature contribution to the offsets vanishes as $t\to\infty$: the MAP offset contains $\tfrac12\log\det C_k$, which is
sensitive to small eigenvalues, whereas the PnP offset contains $\tfrac12\sum_i\log(1+\lambda_i(C_k)/t)$, in which an eigenvalue
$\lambda_i\ll t$ contributes approximately $\lambda_i/(2t)$. For the exact mixture these limits hold with $\lambda\to\infty$ first, on a fixed interval $[t_0,t_1]$, since at fixed
$\lambda$ a large $t$ merges the components. Part (iii) turns the comparison into a diagnostic of agreement with the MAP; near a score
tie the error bounds do not decide.

The offsets are familiar evidence constants of smoothed Gaussian components. They appear in the log-partition of Gaussian
mixtures \cite{DarbonLangloisMeng2023} and in component selection for restoration with Gaussian-mixture models
\cite{YuSapiroMallat2012}, whose criterion with $\log\det C_k$ is the small-$t$ endpoint of (ii). For a single Gaussian they only shift
the induced energy by a constant \cite{Gribonval2011,Liu2026TwoLevels}. What is new is their role as a selection rule for the PnP
decision among modes.

\begin{remark}[Switching and two switches]\label{rem:switching-scores}
The retained variances determine the direction of a switch. For two components and fixed $y,\alpha$, the difference
$\Delta_R(t)=S_1^R-S_2^R$ of their PnP scores satisfies $\Delta_R'(t)=-\alpha(V_1(t)-V_2(t))/(2t^2)$ by \eqref{eq:offset-derivative}: this is
\Cref{thm:switching} for the scores, and every crossing transfers the selection to the component with the larger retained variance.
If $C_1\preceq C_2$, $C_1\neq C_2$, then $V_1<V_2$ for all $t>0$, and the selection switches at most once, from component $1$ to
component $2$, at the root of
\begin{equation}\label{eq:switch-equation}
 \det(C_1+tI)=\rho\det(C_2+tI),\qquad \rho=(w_1/w_2)^2\exp\bigl(2(Q_2^{(0)}-Q_1^{(0)})/\alpha\bigr);
\end{equation}
For $C_i=v_iI$ the root is $t_*=(hv_2-v_1)/(1-h)$, $h=\rho^{1/n}$, when $h\neq1$ and $t_*>0$.
In dimension one and for isotropic covariances the retained variances are always ordered.

Without such an order two switches can occur.
For $n=2$, $C_1=\operatorname{diag}(1,9)$, $C_2=4I$, $w_1/w_2=5/(2\sqrt6)$ and data at both means,
$c_1^R-c_2^R=\tfrac12\log\bigl(24(t+1)(t+9)/(25(t+4)^2)\bigr)$ vanishes at $t=20\pm6\sqrt6$.
The score selection is $1\to2\to1$.
Here $V_2>V_1$ for small $t$ since $\Tr C_2^{-1}<\Tr C_1^{-1}$, and $V_1>V_2$ for large $t$ since $\Tr C_1>\Tr C_2$.
By \Cref{thm:mixture-selection} and the transversality of these crossings, the exact selection
on a fixed interval containing both has the same pattern for large $\lambda$, with switching scales $20\pm6\sqrt6+O(\varepsilon_\lambda)$.
\end{remark}

Observation thus changes both the location and the stability of the preferred reconstruction. The covariance test, the
proximal hull and the component scores describe these changes at their respective scales. They concern the decisions of the prior,
not their distance from the truth, which is the subject of \Cref{sec:recovery}.

%% file: sections/s5_recovery.tex
\section{Recovery}\label{sec:recovery}

We now ask when the decisions of \Cref{sec:selection} recover a deterministic truth $x^*$, and how the noise level affects the
error, in the framework of regularization theory. Let $y:=Ax^*$, $\norm{y^\delta-y}\le\delta$ and $\mathcal S:=S_y=x^*+K$. A point
$x^\dagger_t\in\mathcal S$ is \emph{selected by the prior force} if $W_t(x^\dagger_t)\cap N\neq\emptyset$, that is,
$0\in W_t(x^\dagger_t)+N_{\mathcal S}(x^\dagger_t)$ with the normal cone $N_{\mathcal S}=N$. This is the condition that remains in the
exact-data, vanishing-weight limit of \eqref{eq:recon}, since every solution of \eqref{eq:recon} satisfies $W_t(x)\cap N\neq\emptyset$
(\Cref{prop:nullspace}(i)). Its \emph{selection bias} is $\rho_t(x^*):=\norm{x^\dagger_t-x^*}$. The scale $t$, the weight $\alpha$ and the
noise $\delta$ are independent parameters, and statements such as ``$t=O(\delta)$'' are asymptotic rates. As $t\to0$ the prior force
tends to $\nabla g$; it does not disappear (\Cref{prop:small-sigma}).

\subsection{Selection bias}

We first bound the error of the prior's choice on the exact-data fiber, before data noise and weights enter. On a convex set $U\subseteq\mathcal S$ on which $W_t$ is single-valued, we say that $W_t$ is \emph{$c_K$-strongly monotone along the fiber} if
\begin{equation}\label{eq:local-fibre-monotone}
 \inn{P_K\bigl(W_t(x)-W_t(x')\bigr)}{x-x'}\ge c_K\norm{x-x'}^2\qquad(x,x'\in U),\qquad c_K>0 .
\end{equation}

\begin{theorem}[Selection bias]\label{thm:bias}
\begin{enumerate}
\item[(i)] \emph{(Global.)} If $W_t$ is maximally monotone and $c$-strongly monotone, with $\mathcal S\cap\operatorname{int}\dom W_t\neq\emptyset$,
 then the selected point $x^\dagger_t$ exists and is unique, and $\rho_t(x^*)\le c^{-1}\dist\bigl(0,P_KW_t(x^*)\bigr)$ if $x^*\in\dom W_t$.
\item[(ii)] \emph{(Local, on the fiber.)} If $U\subseteq\mathcal S$ is convex, contains $x^*$ and a selected point $x^\dagger_t$, and $W_t$ is
 $c_K$-strongly monotone along the fiber on $U$, then $x^\dagger_t$ is the only selected point in $U$ and
 $\rho_t(x^*)\le\norm{P_KW_t(x^*)}/c_K$. For a Bayes denoiser of a prior $\pi\in\PP_2(\mathbb R^n)$ with full affine span and
 $U\subseteq\ran D_t$, one may take
 \[
  c_K=\inf_{z\in D_t^{-1}(U)}\Bigl(\lambda_{\max}\bigl(\Sigma_{K\mid N}(z)\bigr)^{-1}-t^{-1}\Bigr)
 \]
 whenever this number is positive.
\end{enumerate}
\end{theorem}
\noindent The selection bias is at most the norm of the component of the prior force at the truth along the unobserved directions,
$\norm{P_KW_t(x^*)}$, divided by the curvature bound along the fiber; the effective weight plays no role. The global hypothesis of (i) essentially requires the convex
regime $t>t_c(\pi)$ of \Cref{rem:threshold}. For multimodal priors at useful noise levels the local version (ii) is the one to use,
and its constant is the Schur-complement curvature of \Cref{thm:fibre-selection}(b): positive curvature along the fiber controls the
bias without global monotonicity. The proof is in \Cref{app:proofs-recovery}.

\begin{remark}[Random truths]\label{rem:random-truth}
Since (ii) holds truth by truth, it can be averaged over a class of truths without assuming a relation to the training
distribution. Let $x^*\sim\mu$, a law unrelated to $\pi$, and suppose that (ii) applies for $\mu$-almost every $x^*$ with a constant
$c_K(x^*)>0$, where $c_K(x^*)$, $\norm{P_KW_t(x^*)}$ and a selected point $x^\dagger_t(x^*)$ depend measurably on $x^*$. Taking expectations,
and using $\{\norm{x^\dagger_t-x^*}>\varepsilon\}\subseteq\{\norm{P_KW_t(x^*)}>c_K(x^*)\varepsilon\}$ up to a null set,
\[
 \mathbb E_\mu\norm{x^\dagger_t-x^*}\le\mathbb E_\mu\frac{\norm{P_KW_t(x^*)}}{c_K(x^*)},\qquad
 \mathbb P_\mu\bigl(\norm{x^\dagger_t-x^*}>\varepsilon\bigr)\le\mathbb P_\mu\bigl(\norm{P_KW_t(x^*)}>c_K(x^*)\,\varepsilon\bigr).
\]
\end{remark}

When $P_K\nabla g(x^*)\neq0$, the prior force in the bound can be computed, to leading order, from one denoiser evaluation at the truth, since
$W_t(x^*)=(x^*-D_t(x^*))/t+O(t)$ (\Cref{prop:small-sigma}). For fiber-critical truths both $P_KW_t(x^*)$ and this estimate are $O(t)$ and differ at
that order, by the gap between the PnP and score decisions in \Cref{thm:fibre-selection}(c), so the forward evaluation does not give
the leading bias. Matching the denoiser to the distribution of truths does not in general make the bounds vanish, since they include the
intrinsic non-recoverability of $\ker A$. Nor can they be estimated from the data, because $P_KW_t$ vanishes at every computed
solution:

\begin{lemma}[What the data determine]\label{prop:nullspace}
\begin{enumerate}
\item[(i)] Every zero $x$ of $T_{\alpha,t}$ satisfies $W_t(x)\cap K^\perp\neq\emptyset$, for every $\alpha>0$ and every datum: the
 component of the reconstruction in $K$ satisfies $P_KW_t(x)\ni0$, an equation in which the data do not appear.
\item[(ii)] The residual $\norm{Ax-y^\delta}$ is invariant under $x\mapsto x+v$, $v\in K$.
\end{enumerate}
\end{lemma}
\begin{proof}
(i) If $0=A^*(Ax-y^\delta)+\alpha w$ with $w\in W_t(x)$, then $w\in\ran A^*\subseteq K^\perp$.
(ii) $A(x+v)=Ax$.
\end{proof}

When $K\neq\{0\}$, the data cannot distinguish truths on the same fiber, so no reconstruction rule recovers all of them
\cite{MicchelliRivlin1977}: if $v\in K$, the truths $x^*$ and $x^*+v$ give the same exact data, and any map from $(A,y^\delta,\delta,\{D_t\})$ to a reconstruction
errs by at least $\tfrac12\norm v$ on one of them. Stability and convergence to a selected solution \cite{ObmannHaltmeier2026}
therefore do not imply recovery of the truth, and a residual-only rule for the noise level cannot observe the selection bias.
Which truths can be recovered as the parameters vanish?

\begin{proposition}[Only fiber-critical truths can be recovered]\label{prop:only-critical}
Let $\pi=e^{-g}$ satisfy the hypotheses of \Cref{prop:small-sigma}. Let $x_k$ solve $0\in T_{\alpha_k,t_k}(x_k)$ with data
$y^{\delta_k}$, where $(t_k,\alpha_k,\delta_k)\to0$, and let $(x_k)$ be bounded. Then every limit point $\bar x$ satisfies
$A\bar x=y$ and $P_K\nabla g(\bar x)=0$. Hence $x_k\to x^*$ is possible only if $x^*$ is a critical point of $g$ on its fiber
$x^*+K$.
\end{proposition}
\begin{proof}
Pass to a subsequence $x_k\to\bar x$ and let $V$ be a compact neighborhood of $\bar x$. By \Cref{prop:small-sigma}, $V\subseteq\ran D_t$
for small $t$, where $W_t$ is single-valued, and $W_t\to\nabla g$ uniformly on $V$. Hence $W_{t_k}(x_k)\to\nabla g(\bar x)$. By
\Cref{prop:nullspace}(i), $P_KW_{t_k}(x_k)=0$, so $P_K\nabla g(\bar x)=0$. Moreover
$A^*(Ax_k-y^{\delta_k})=-\alpha_kW_{t_k}(x_k)\to0$, so $A^*(A\bar x-y)=0$. Since $A\bar x-y\in\ran A\cap\ker A^*=\{0\}$, $A\bar x=y$.
\end{proof}

Thus, if the truth is not a critical point of $g$ on its fiber, no bounded sequence of reconstructions converges to it, and
annealing toward the MAP force does not remove the prior's choice along the fiber. The condition $P_K\nabla g(x^*)=0$ concerns the
deterministic truth and plays the role of a source condition. For a random $X\sim\pi$ it typically fails almost surely: for a
nondegenerate Gaussian prior, $P_K\nabla g(X)=P_K\Sigma^{-1}(X-m)$ has a nondegenerate Gaussian law on $K$ whenever $K\neq\{0\}$. No global
bound on $\nabla g$ is needed, so Gaussian priors and Gaussian mixtures are covered.

For multimodal priors, selecting a wrong component, or the bridge of \Cref{ex:two-mode}, causes errors that persist as the data
noise and the weight vanish. The next corollary follows from
\Cref{thm:mixture-selection} and \Cref{ex:two-mode} by the triangle inequality.

\begin{corollary}[Wrong mode or bridge]\label{prop:wrong-core}
\begin{enumerate}
\item[(i)] In setting (a) of \Cref{ass:compatible}, let the truth lie within distance $\rho_*$ of the mean $m_{k^*}$ of a component $k^*$. If a
 global minimizer $u$ of $F_\Phi$ (or of $\Phi$ on $S_y$) belongs to another component, then
 \[
  \norm{u-x^*}\ge\delta_\mu\lambda-\rho_*-C,
 \]
 with $C$ independent of $\lambda$, $t\in[t_0,t_1]$ and $\alpha\le\alpha_1$. The error is of the order of the separation and does not
 decrease as $\delta,\alpha\to0$.
\item[(ii)] In \Cref{ex:two-mode}, for the truths $x^*=(0,\pm k_G)$ and exact data, the hard-constrained PnP decision has
 error $k_G$ at every $t\ge t_R$.
\end{enumerate}
\end{corollary}
\begin{proof}
(i) By \Cref{thm:mixture-selection}, $u=v_{k,\Phi}$ for some $k\in J$, $k\neq k^*$, with $\norm{u-x_k^{(s_\Phi)}}\le\ell\le\tfrac14$, and
$x_k^{(s_\Phi)}\in\bar B(m_k,r)$ with the constant $r$ of Step 2 of \Cref{app:proofs-mixtures} (for the hard constraint, $\bar x_k^{(s_\Phi)}$).
Since $\norm{m_k-m_{k^*}}\ge\delta_\mu\lambda$, the triangle inequality gives
$\norm{u-x^*}\ge\norm{m_k-m_{k^*}}-\norm{u-m_k}-\norm{m_{k^*}-x^*}\ge\delta_\mu\lambda-r-\rho_*-\ell$, with
$\ell=(c_++t_1)\varepsilon_\lambda\le\tfrac14$, so $C=r+\tfrac14$ suffices.
(ii) By \Cref{ex:two-mode}(ii) the decision is the origin for $t\ge t_R$.
\end{proof}

Reducing the data noise or the weight cannot remove these errors while the wrong component or the bridge remains selected.

\subsection{Local reconstruction and the choice of the noise level}
\label{sec:sharp-local}

Near a fiber-critical truth we can expand the error. The measurement can stabilize directions in which the prior has negative
curvature, so positive curvature is needed only along $\ker A$. Let $B=A^*A|_N\succ0$ (if $N=\{0\}$, terms with $B^{-1}$ are omitted),
and for a truth with $P_K\nabla g(x^*)=0$ put
\[
 H=\nabla^2g(x^*),\ \ H_K=P_KH|_K\succ0,\ \
 \mathcal L n=n-H_K^{-1}P_KHn\ \ (n\in N),\quad b=\tfrac12H_K^{-1}P_K\nabla\Delta g(x^*).
\]
The map $\mathcal L$ augments an observed displacement $n$ with the unobserved displacement $-H_K^{-1}P_KHn$ that it induces
through the prior curvature, and $b$
is the scale bias of \eqref{eq:xR} at $x_G=x^*$; when $K=\{0\}$, $b=0$ and $\mathcal L=I$. We allow the prior force used to be
$\widehat W_t=W_t+\psi_t$, with a perturbation $\psi_t$ that is $C^1$ near $x^*$, and measure all perturbations together by
$q:=\delta+\alpha+t+\varepsilon_W(t)$, where $\varepsilon_W(t)=\norm{\psi_t}_{C^1(U)}$ on the ball $U$ of the theorem.

\begin{theorem}[Local error expansion and the choice of $t$]\label{thm:sharp-local-reconstruction}
Let $\pi=e^{-g}$ satisfy \Cref{prop:small-sigma}, and let the truth be a nondegenerate critical point of $g$ on its fiber,
$P_K\nabla g(x^*)=0$ and $H_K\succ0$. There are a closed ball $U$ about $x^*$ and constants $C,q_0>0$ such that, whenever $\alpha>0$ and
$q\le q_0$:
\begin{enumerate}
\item[(i)] the equation $A^*(Ax-y^\delta)+\alpha\widehat W_t(x)=0$ has exactly one solution $\hat x$ in $U$, a strict local minimizer
 of $F_{R_t}$ when $\psi_t=0$, and
 \begin{equation}\label{eq:sharp-local-expansion}
  \hat x-x^*=\mathcal L B^{-1}A^*v-\alpha\mathcal L B^{-1}P_N\nabla g(x^*)-t\,b-H_K^{-1}P_K\psi_t(x^*)+\mathcal R,
  \qquad \norm{\mathcal R}\le Cq^2 ;
 \end{equation}
\item[(ii)] for the exact prior force and $\alpha\asymp\delta$, the error is $O(\delta)$ if and only if $t=O(\delta)$ when $b\neq0$, and it is
 $O(\delta+t^2)$ when $b=0$;
\item[(iii)] an approximate solution $\tilde x\in U$, with $A^*(A\tilde x-y^\delta)+\alpha\widehat W_t(\tilde x)=\zeta$, satisfies the same expansion
 up to an additional error $C(\norm{P_N\zeta}+\norm{P_K\zeta}/\alpha)$: the residual along $\ker A$ is amplified by $1/\alpha$.
\end{enumerate}
\end{theorem}
\noindent The expansion separates the four sources of error. Data noise and the prior force in observed directions propagate
through $\mathcal L$; the denoiser scale shifts the reconstruction along $\ker A$ by $-tb$; a force error contributes
$-H_K^{-1}P_K\psi_t(x^*)$. No positivity of $H$ on $N$ or monotonicity of $W_t$ is assumed, and the constants depend on $A$, including its
smallest nonzero singular value (proof in \Cref{app:proofs-recovery}). Part (ii) gives the scale needed to keep the error at the
level of the data noise. Part (iii) concerns solver accuracy: the residual along $\ker A$ is divided by $\alpha$, so a small unscaled
residual is compatible with a large reconstruction error.

A trained denoiser has errors, and they are amplified by $t^{-1}$ in the prior force. For $\widehat D_t=D_t+\eta\phi$, $\eta\ge0$,
$\phi\in C_b^2$, used with its local inverse branch, the force error enters \eqref{eq:sharp-local-expansion} as $(\eta/t)e$ with
$e=H_K^{-1}P_K\phi(x^*)$ (\Cref{lem:perturbed-force}). The error along $\ker A$ thus balances the scale bias $-tb$ against the model error
$(\eta/t)e$. The next corollary bounds the resulting error and finds the best balance.

\begin{corollary}[Denoiser errors and the oracle noise level]\label{cor:sharp-scale}
In the setting of \Cref{thm:sharp-local-reconstruction}, let the denoiser used be $\widehat D_t=D_t+\eta\phi$ as above, and $\alpha\asymp\delta$.
\begin{enumerate}
\item[(i)] For $t\asymp\sqrt\eta$ and a residual $\norm\zeta\le\alpha\varepsilon$, the approximate reconstruction satisfies
 $\norm{\tilde x-x^*}\le C(\delta+\sqrt\eta+\varepsilon)$ for small parameters.
\item[(ii)] Let $b\neq0\neq e$ and $\delta=o(\sqrt\eta)$. As $\eta\to0$, every minimizer $t^*$ of $\norm{\hat x-x^*}$ over the window $I_\eta$
 defined below satisfies
\begin{equation}\label{eq:oracle-scale}
 \frac{t^*}{t_\eta}\to1,\qquad t_\eta:=\Bigl(\frac{\eta\norm e}{\norm b}\Bigr)^{1/2},\qquad
 \norm{\hat x-x^*}\big|_{t^*}=\sqrt{2\eta\bigl(\norm b\norm e-\inn be\bigr)}+o(\sqrt\eta).
\end{equation}
 In particular, the best noise level is set by the denoiser accuracy $\eta$, not by the data noise $\delta$, and for fixed $s\neq1$
 the noise level $s\,t_\eta$ gives a larger error at leading order.
\end{enumerate}
\end{corollary}
\noindent Here $I_\eta=\{0<t\le t_2:\ t+\eta/t\le\varepsilon_0\}$, with $t_2$ and $\varepsilon_0$ fixed and small enough for
\eqref{eq:sharp-local-expansion} to hold on $I_\eta$. The proof is in \Cref{app:proofs-recovery}.

Residual-based rules cannot determine the oracle scale (\Cref{prop:nullspace}(ii)).
Since $-tb$ and $(\eta/t)e$ enter with opposite signs, they partly cancel when $b$ and $e$ point in similar directions, which the term
$\inn be$ in \eqref{eq:oracle-scale} records. The oracle scale needs the unknown truth and is a benchmark. A similar balance between
smoothing bias and inverse-scale amplification sets early stopping in a Wasserstein bound for diffusion posterior sampling
\cite{Chang2025ProvableDPS}; here \eqref{eq:oracle-scale} gives a deterministic leading constant with the vector cancellation. The local
result assumes a fixed $A$, gives no discretization-uniform rate, and does not select a basin.

\begin{remark}[Force errors of power type]\label{rem:power-force-error}
When only a bound on the force error is known, the same balance gives a scale. If $\norm{H_K^{-1}P_K\psi_t(x^*)}\le\kappa t^{\gamma-1}$
and $\varepsilon_W(t)\le\kappa't^{\gamma-1}$ on a window of scales, with $\gamma<1$, then \Cref{thm:sharp-local-reconstruction}(i) with
$\alpha\asymp\delta$ gives $\norm{\hat x-x^*}\le\norm b\,t+\kappa t^{\gamma-1}+C(\delta+q^2)$. For $b\ne0$ the first two terms are minimal at
$t^*=\bigl((1-\gamma)\kappa/\norm b\bigr)^{1/(2-\gamma)}$, which for $\gamma=0$ and $\kappa=\eta\norm e$ is the scale $t_\eta$ of
\Cref{cor:sharp-scale}. The bound ignores the cancellation between the two displacements, which can only lower the error.
\end{remark}

We next test the fiber decisions and their predicted recovery errors numerically.

%% file: sections/s6_experiments.tex
\section{Numerical evidence}\label{sec:experiments}

The experiments test the preceding analysis of fiber decisions and recovery. We ask whether the identities,
coefficients and thresholds of \Cref{sec:object,sec:preference,sec:selection,sec:recovery} hold quantitatively for exact priors,
and whether the decisions of one prior can be computed and predicted on images.

\emph{Setting and protocol.} Exact priors let us isolate the effect of the decision rule. For Gaussian mixtures,
$p_t,D_t,\Sigma_t,E_t$ have closed forms. For other low-dimensional priors we use Gauss--Hermite quadrature.
We evaluate $R_t$ by \eqref{eq:three-energies}, computing the inverse by damped Newton iteration with Hessian $\Sigma_t/t$ and residual
$\le10^{-9}$. Alternatively, we use the noisy coordinate of \Cref{prop:noisy-coordinates}(iii) to avoid inversion. The image prior has $20$
full-covariance components fitted to $20\,000$ MNIST digits in a $60$-dimensional principal subspace, lifted to $\mathbb R^{784}$ with
complementary variance $0.0106$. All three decisions use this same exact mixture.

Computed minimizers are the best of several local runs, without global certification. For small $\alpha$ the null-space part of
the reconstruction equations has size $O(\alpha)$; we solve the equations with that part divided by $\alpha$, which have the same zeros.

\subsection{Structure and preference along the family}\label{sec:exp-structure}

\begin{figure}[t]
\centering
\includegraphics[width=\textwidth]{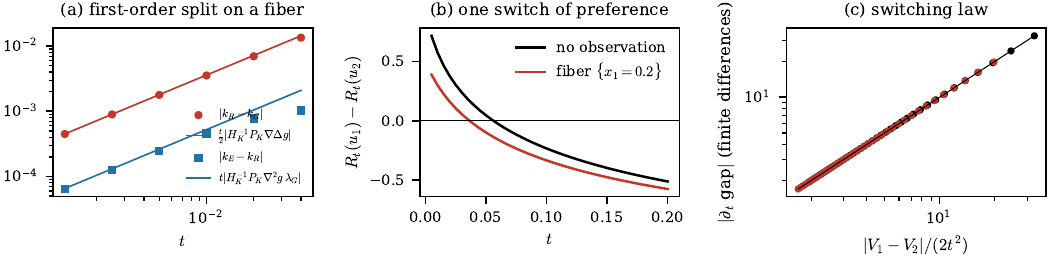}
\caption{Decisions and preferences along the family; all horizontal axes are the noise variance $t$ or functions of it. (a)
Separation of the fiber decisions of $R_t$, $E_t$ and $g$ on $\{x_1=0.4\}$ for the mixture with weights $0.6$, $0.4$, means
$(0,0)$, $(1.5,1)$ and covariances $\bigl(\begin{smallmatrix}1&0.3\\0.3&0.6\end{smallmatrix}\bigr)$,
$\bigl(\begin{smallmatrix}0.5&-0.2\\-0.2&0.8\end{smallmatrix}\bigr)$, against the first-order
predictions \eqref{eq:xR}--\eqref{eq:xE-xR} (lines, slope one); axis $t$. (b) Energy gap between two branches of minimizers of
$R_t$ against $t$, without observation (one-dimensional mixture with weights $0.8$, $0.2$ and variances $1$, $0.01$) and on the fiber $\{x_1=0.2\}$ of
the mixture with weights $0.8$, $0.2$, means $(0,-2)$, $(0.3,2)$ and covariances
$\bigl(\begin{smallmatrix}0.5&0.3\\0.3&1\end{smallmatrix}\bigr)$, $\bigl(\begin{smallmatrix}0.5&0.1\\0.1&0.04\end{smallmatrix}\bigr)$; each
switches once. (c) Measured $t$-derivative of the gap against the variance law \eqref{eq:switching},
$40$ values of $t$ each; the line is the identity.}
\label{fig:mechanisms}
\end{figure}

We begin with the structural identities and the change of preference across scales. \Cref{tab:closed-form} compares the
identities with exact values. In mixtures with a light narrow and a heavy broad mode, the variance-ordering law matches
the derivatives in $t$ of the energy gap both without observation and on $\{x_1=0.2\}$.
The constrained minimizers are not critical points of $p_t$, yet each preference switches once, toward larger retained
variance (\Cref{fig:mechanisms}(b,c)).
Retained variance predicts which branch has the faster-decreasing energy as the noise level grows, also on a fiber.
It does not guarantee that an iteration will leave its current basin.

\begin{table}[t]
\centering\small
\caption{Structural identities against exact values. Relative errors unless stated.}
\label{tab:closed-form}
\begin{tabularx}{\textwidth}{@{}>{\raggedright}X>{\raggedright}p{4.1cm}>{\raggedright\arraybackslash}p{5.0cm}@{}}
\toprule
statement & tested against & result \\
\midrule
$\nabla D_t=\Sigma_t/t$, \eqref{eq:tweedie} & finite differences & $1.9\cdot10^{-10}$ \\
$\Lip D_t=R^2/t$, attained at $0$ (\Cref{rem:threshold}) & two-point prior, $t<R^2$ & $\Lip D_t=2.0864=R^2/t$ \\
rigidity (\Cref{thm:rigidity}(ii)) & drift of $W_t$ in $t$ & $\le1.5\cdot10^{-13}$ Gaussian; $0.3639$ otherwise \\
small-scale expansion \eqref{eq:small-sigma} & halving $\sigma=\sqrt t$ & error ratios $16.2$, $14.8$ (predicted $16$) \\
$\Fix D_t=\operatorname{Crit}p_t$ (\Cref{thm:stationary-structure}(i)) & $24$ critical points & residual $1.7\cdot10^{-13}$ \\
congruence \eqref{eq:congruence}, equal inertia & finite-difference Hessians & $1.5\cdot10^{-7}$; inertia equal at all $24$ \\
variance law \eqref{eq:switching} & central differences in $t$ & $9.1\cdot10^{-11}$; on a fiber $2.2\cdot10^{-10}$ \\
\bottomrule
\end{tabularx}
\end{table}

\subsection{Decisions on fibers}\label{sec:exp-selection}

We next test how observation separates the decisions and changes their stability. On a two-component mixture with fiber
$\{x_1=0.4\}$, the transfer identity \eqref{eq:transfer} holds to $4.2\cdot10^{-14}$. At
$t=0.00125$ the measured coefficients of \eqref{eq:xR} and \eqref{eq:xE-xR} are $0.35912$ and $-0.05150$ against the predicted
$0.35978$ and $-0.05237$. The error decreases as $t$ is halved (\Cref{fig:mechanisms}(a)). We also test $189$ noisy points at three
values of $t$, including points in nonconvex regions. The Schur-complement criterion \eqref{eq:schur-condition} decides the sign of the
fiber curvature of $R_t$ in every case.
The relevant curvature diagnostic therefore accounts for the observed directions through the Schur complement, rather
than using only the marginal covariance along the fiber.

The transition in \Cref{ex:two-mode} shows how this curvature difference changes the selected point.
For $a=1$, $\varepsilon=0.05$, the thresholds are $t_R=0.045238$, $t_E=0.95$.
The eleven curvature checks agree. By \Cref{ex:two-mode}(i), the origin remains an unconstrained saddle for $0<t<2a^2-\varepsilon$. Its prior density ratio is $9.08\cdot10^{-5}$,
and the largest inverse residual is $4.7\cdot10^{-10}$.
We compute the decisions across $53$ values of $t$ in $[0.004,2]$, with a $k$-grid of spacing $0.004$ and inverse residual
$\le1.0\cdot10^{-11}$. The induced energy $R_t$ selects zero for $t\ge1.03t_R$ and $|k|\ge0.05$ below $0.97t_R$.
The smoothed energy $E_t$ makes the corresponding choices at the $t_E$ bounds, while $g$ selects $|k|=1.000$
(\Cref{fig:transition}(d)).
At every one of the $53$ values the computed point selected by the induced energy agrees with the exact solution $\abs k=as_t$ of
\Cref{ex:two-mode}(ii) to within $0.0013$, below the grid spacing.
Thus changing the denoiser scale can select a bridge with low prior density, even while the unconstrained origin remains a saddle.

\subsection{One prior, three decisions: inpainting with an exact mixture prior}\label{sec:exp-gmm}

We now ask whether the component scores predict decisions in image inpainting. We observe the bottom $12$ of the $28$ rows
of a test digit, with Gaussian noise of standard deviation $0.02$, and reconstruct the
$448$ unobserved pixels. For $40$ test images and $t\in\{0.0025,0.01,0.04,0.09,0.25\}$ we compute three decisions of the same $\pi$:
the MAP $x_G$, the score decision $x_E$ with $E_t$ in place of $g$, and the PnP decision $x_R$ with $R_t$. The posterior of
$\pi$ given the data is again a Gaussian mixture; $H$ denotes the entropy of its component probabilities. We set $\alpha=0.02^2$ equal to the noise variance and hold this effective weight fixed as the
denoiser scale varies. The test images are not samples of the prior.

\begin{figure}[t]
\centering
\includegraphics[width=0.66\textwidth]{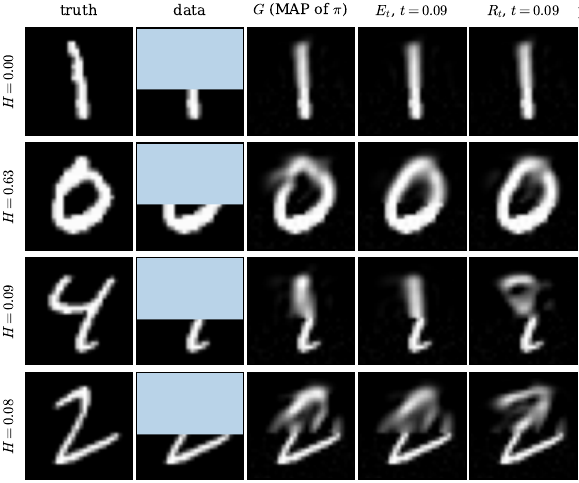}
\caption{Height against mass. Columns: truth, data (blue rows unobserved), MAP of $\pi$, and the score and PnP decisions at
noise variance $t=0.09$. Top: the decisions agree in the dominant posterior component. Below: the PnP
decision has a different dominant component from the narrow MAP peak, while the score decision stays in the component of the
peak. $H$
is the entropy of the posterior component probabilities.}
\label{fig:gmm-images}
\end{figure}

\emph{Two groups.} The location of the MAP within the posterior separates the images into groups.
For $32$ images the MAP lies in the dominant posterior component, and the PnP decision reproduces it.
Up to $t=0.04$ the median distance, $\norm{x_R-x_G}_2/\sqrt{784}$ over all pixels, is at most $1.6\cdot10^{-5}$, with maximum
$0.022$.
No component changes occur below $t=0.25$. The median distance of the score decision, however, grows from $0.0088$ to $0.093$.

For the other eight images the MAP lies in a narrow peak with little mass, at most $1.8\,\%$ on five of them.
At the five values of $t$, the PnP decision lies in a different component from the MAP on $0,2,3,6,8$ of these images.
The score decision changes component on at most three of all forty at any $t$ (\Cref{fig:gmm-images}).
This is the behavior of \Cref{thm:mixture-selection,cor:mixture-offsets,rem:switching-scores},
although the prior is far from the asymptotic regime: the penalty on narrow modes increases with $t$ and changes the PnP choice
from the highest peak toward the component of largest mass.

The covariance diagnostics relate these changes to retained variance.
All $21$ PnP components that differ from the MAP's have larger retained variance.
Random alternative components would have larger variance in $12.1$ cases on average, so this proxy is weak.
The largest Schur ratio $\lambda_{\max}(\Sigma_{K|N})/t$ over all images decreases from $0.9988$ to $0.9428$ as $t$ grows.
Agreement with the MAP does not give the same ranking as the mean peak signal-to-noise ratio (PSNR), computed over all pixels
with intensities in $[0,1]$. The score decision has the highest mean PSNR for
$t\in[0.01,0.09]$. The experiment separates decisions and does not rank them.

\emph{Predicting the decisions.} The scores of \Cref{thm:mixture-selection} need only the mixture parameters, the mask and the data.
On the $40$ instances they predict the component of $195$ of $200$ PnP decisions and of all $40$ MAP decisions.
We compare them with the rule ``the decision stays in the component of the MAP'', which is right in $179$ PnP cases.
The scores predict $20$ of the $21$ departures with $4$ false alarms. For the score decision they are not more accurate than the rule.

To remove possible starting-point bias and test unseen data, we recompute all decisions from starting points in all
twenty components and one further data-dependent starting point. We use the $40$ images and $100$ unseen instances: $40$ new test images, $30$ with the top
$12$ rows observed instead, and $30$ with noise standard deviation $0.05$. The scales are $t\in\{0.01,0.09,0.25\}$.
On the original images the dominant components agree with the earlier ones in $95.0\,\%$ (PnP), $99.2\,\%$ (score)
and $100\,\%$ (MAP) of the cases.

The scores also predict departures from the MAP on unseen instances (\Cref{tab:heldout}).
They predict the component of the PnP decision in $96.3\,\%$ of $300$ cases, against $77.7\,\%$ for the trivial rule.
They predict all $67$ departures from the MAP component with $3.4\,\%$ false alarms.
The bootstrap lower $95\,\%$ bound over images is $94\,\%$.
The score decision is predicted in $88.3\,\%$, against $84.7\,\%$ for the trivial rule.

Predicting the switching scale is less reliable than predicting the selected component.
The predicted scale lay in the observed bracket for $7$ of $10$ switching
images on the original grid.
It lay within a factor $1.5$ of the observed scale for $8$ of $11$ switching images on a grid of twelve values of $t$.

\begin{table}[t]
\centering\small
\caption{Prediction of the PnP decision by the scores of \Cref{thm:mixture-selection} on $100$ unseen instances ($300$
cases), decisions recomputed from all component candidates. ``Trivial'': the decision stays in the component of the MAP.
A departure is a case in which the PnP decision lies in a different component from the MAP decision; ``departures predicted'' counts
the departures that the scores flag, and ``false alarms'' is the fraction of cases without a departure in which the scores flag
one.}
\label{tab:heldout}
\begin{tabular}{@{}lccccc@{}}
\toprule
condition & instances & scores & trivial & departures predicted & false alarms \\
\midrule
new images & $40$ & $95.0\,\%$ & $73.3\,\%$ & $32/32$ & $4.5\,\%$ \\
top rows observed & $30$ & $97.8\,\%$ & $81.1\,\%$ & $17/17$ & $2.7\,\%$ \\
noise $0.05$ & $30$ & $96.7\,\%$ & $80.0\,\%$ & $18/18$ & $2.8\,\%$ \\
all & $100$ & $96.3\,\%$ & $77.7\,\%$ & $67/67$ & $3.4\,\%$ \\
\bottomrule
\end{tabular}
\end{table}

\subsection{Recovery and the choice of the noise level}\label{sec:exp-recovery}

We now test the recovery error and its dependence on the denoiser scale.

\begin{figure}[t]
\centering
\includegraphics[width=\textwidth]{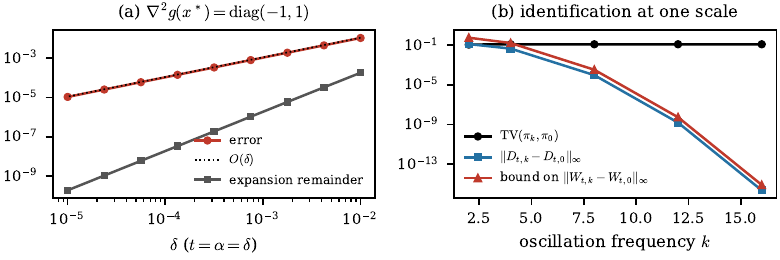}
\caption{Recovery and identification. (a) Error of \Cref{thm:sharp-local-reconstruction} against $\delta$, with $t=\alpha=\delta$ (so the
axis is also the noise variance $t$), for a prior whose Hessian at the truth is $\operatorname{diag}(-1,1)$: linear in $\delta$, with
quadratic expansion remainder. (b) One scale does not identify the prior stably: priors at total-variation distance bounded away from zero with
exponentially close denoisers and prior forces (\Cref{prop:prior-inverse-unstable}); axis: oscillation frequency $k$.}
\label{fig:recovery}
\end{figure}

\emph{Local reconstruction.} The local error expansion allows negative prior curvature in measured directions.
We test this with $A=(1,0)$ and $g(x,k)=\sqrt{1+x^2}+e^{-x^2}+\sqrt{1+k^2}+0.12k^3e^{-k^2}$.
The Hessian at the truth is $\operatorname{diag}(-1,1)$, so the measured direction has negative prior curvature.
With $\alpha=t=\delta$, the error of \Cref{thm:sharp-local-reconstruction} has log--log slope $1.00002$,
and the expansion remainder is quadratic (\Cref{fig:recovery}(a)).
A scale that decays too slowly reduces the convergence rate. With $t=\delta^{1/4}$ the scale bias $\norm b\,t$ dominates, and as
$\delta$ decreases from $10^{-8}$ to $10^{-12}$ the measured log--log slope increases from $0.229$ to $0.246$, approaching the value
$\tfrac14$ given by the expansion of \Cref{thm:sharp-local-reconstruction}(i).

We also test the predicted coefficient of scale bias for the prior
$g(x)=\sqrt{1+x_1^2}+\sqrt{1+x_2^2}+0.8\log\cosh(x_1-2x_2+0.5)$ in $\mathbb R^2$ on the fibers $\{x_1=0.3\}$ and
$\{(x_1+x_2)/\sqrt2=-0.8\}$, and for
$g(x)=\sum_{i=1}^3\sqrt{1+x_i^2}-1.5\,e^{-\norm{x-m}^2/2}$, $m=(0.4,-0.3,0.2)$, in $\mathbb R^3$, which is not log-concave, with
$A=(1,1,1)/\sqrt3$ and the truth $x^*\approx(0.3854,-0.0391,0.2600)$, a local minimizer of $g$ on $\{Ax=0.35\}$. At the smallest scale, the coefficient differs from $-b$ by a relative
$2.0\cdot10^{-3}$, $1.7\cdot10^{-3}$ and $1.0\cdot10^{-3}$.
The difference decays with orders $1.988$, $1.989$ and $1.978$ in $\sigma$, so the remainder is $O(t^2)$.

\emph{The oracle noise level.} We next test the balance between scale bias and amplified denoiser error.
We perturb the denoiser of the prior in $\mathbb R^3$ by the model error $\eta(\sin2z_2,\cos1.5z_1,\sin(z_1+z_3))$, with
$\alpha=\delta\in\{10^{-9},10^{-10}\}$ and datum $0.35+\delta$; here $\dim\ker A=2$.
The cosine between $b$ and $e$ is $-0.915$, so the two error terms reinforce rather than cancel.
The best of $64$ scales is $\sigma=0.1182$, $0.0658$, $0.0367$ for $\eta=10^{-4},10^{-5},10^{-6}$,
against $\sqrt{t_\eta}=0.1166$, $0.0656$, $0.0369$ from \eqref{eq:oracle-scale}.
The best error is $4.282\cdot10^{-3}$, $1.346\cdot10^{-3}$, $4.249\cdot10^{-4}$ against the predicted
$4.245\cdot10^{-3}$, $1.342\cdot10^{-3}$, $4.245\cdot10^{-4}$.
This tests a benchmark that uses the truth: decreasing the noise level below the balance point need not improve reconstruction.

\emph{Only fiber-critical truths.} Reducing the parameters cannot remove the bias when the truth is not fiber-critical.
A truth moved off the fiber-critical point by $0.3$ is not recovered.
The reconstructions converge to the fiber-critical point, with distance $2.1\cdot10^{-5}$ at $\delta=10^{-7}$.
The error tends to $0.3000$ (\Cref{prop:only-critical}).

\emph{Identification at one scale.} A small difference between denoisers need not mean a small difference between their priors.
For the family of \Cref{prop:prior-inverse-unstable}, the total-variation distance of the priors tends to $\epsilon/\pi$.
The sup-distances of the denoisers and of the prior forces decay exponentially in the frequency (\Cref{fig:recovery}(b)).

%% file: sections/s7_conclusion.tex
\section{Conclusions and open problems}\label{sec:conclusion}

The analysis and experiments describe what a Gaussian denoiser selects when used in reconstruction.
A Gaussian denoiser defines a prior force on clean signals at each noise level. The PnP family uses this force.
RED and score-based methods use the same denoising residual, evaluated at the noisy input.
The MAP uses the MAP force $\nabla g$, the gradient of the negative log-density of the prior. Without observation the induced and smoothed energies of an exact Bayes denoiser have the same critical points and minimizers.
On measurement fibers the three decisions can differ, in ways governed by the posterior covariance and depending on the noise
level.

The prior-force family brings the known structure of Bayes denoisers into a common description across scales.
It is the gradient field of an induced energy whose curvature is the posterior precision minus $t^{-1}$.
The smoothed energy on the score side is its Moreau envelope. As is known \cite[Thm.~F.1]{Liu2026TwoLevels}, the prior-force family is constant in $t$ only for Gaussian priors.
Thus, for a non-Gaussian Bayes denoiser, changing the noise level changes the prior force even when its effective weight is fixed.

Observation determines how this structure enters the reconstruction.
On a fiber, the Schur complement of the posterior covariance determines the stability of the PnP decision.
When the modes of the prior are far apart compared with the noise level and their width, the induced energy is the proximal
hull of the MAP energy at leading order, with parameter the noise variance. For Gaussian mixtures this holds at every separation
up to an explicit constant. At leading order PnP minimizes this hull on the fiber, so its decisions lie at MAP cores or where the hull is strictly below the
MAP energy.
Among compatible modes it uses the data fit of the MAP with a penalty on narrow modes that grows with $t$.
Its offsets order the modes by peak height for small $t$ and by mass for large $t$; these orderings decide the selection
when the data fits agree and the score gaps exceed the error bounds.

These selection effects enter the reconstruction error through the selection bias and an oracle noise level set by denoiser
accuracy.
The oracle balance explains why reducing the noise level need not improve reconstruction with an approximate denoiser.

Several questions remain open, in order of importance.
\begin{enumerate}
\item \emph{Trained denoisers.} A network with non-symmetric Jacobian has no induced energy, and its posterior covariance is not
 available. It remains open which conclusions remain valid for such networks.
 Possible approaches use the symmetric part of the Jacobian or a bound on the error of the prior force as in
 \Cref{thm:sharp-local-reconstruction}.
\item \emph{Low-rank and manifold priors.} Image distributions are often described as concentrated near low-dimensional manifolds.
 The proximal-hull limit of \Cref{cor:hull}(i)(ii) does not require isolated minimizers and covers such priors, but the selection
 among compatible modes of \Cref{thm:mixture-selection} under \Cref{ass:compatible}(b) requires finitely many nondegenerate global minimizers. For priors
 concentrated near low-dimensional sets the MAP degenerates, while the offsets of the PnP scores remain finite.
 For singular priors or nonisolated minima the present component-score theorem need not apply, and an extension requires a
 different local model; the posterior covariance stays positive definite as long as the prior has full affine span.
\item \emph{Beyond leading order and global modes.} For general multimodal priors the proximal-hull limit holds at leading order,
 without a rate. A non-asymptotic version for general priors remains open; for Gaussian mixtures, and for densities within bounded
 factors of them, \Cref{cor:hull}(iii) and \Cref{rem:perturbation} give non-asymptotic bounds. The scores of
 \Cref{thm:mixture-selection} are proved for compatible global modes. They fail at leading order for higher wells that are not
 proximal points, and a selection theory that includes higher local modes remains open.
\item \emph{Selection under annealing.} How annealing the noise level affects which stationary point is selected is raised in
 \cite{Hurault2026annealed}. The branch equation of \Cref{thm:switching} describes the variation within a branch, while the energy-gap identity
 determines changes in the order of two branches. It may provide a starting point for this question.
\item \emph{The prior along observed directions.} When the weighted prior force is comparable in size with the data gradient, the prior also affects
 the reconstruction along observed directions. This regime is outside the present analysis.
\item \emph{Switching scales.} The predicted switching scales were less accurate on the image prior than the predicted components. Whether this is a
 finite-separation effect or a missing term in the scores is not settled.
\end{enumerate}

%% file: sections/app_proofs.tex
\section{Proofs for \texorpdfstring{\Cref{sec:object}}{Section 2}}\label{app:proofs-object}

For the known statements of \Cref{sec:known} we record only what is needed in the present notation.

\begin{proof}[Proof of \Cref{prop:noisy-coordinates}]
Write $\nabla f(x)=A^*(Ax-y^\delta)$; by \eqref{eq:Wt}, $x=D_t(z)$ means $(z-x)/t\in W_t(x)$.
(i) Forward--backward: $z=x-\gamma\nabla f(x)$, so $0\in\nabla f(x)+(t/\gamma)W_t(x)$. ADMM, with updates
$x^{k+1}=\argmin_x\{f(x)+\tfrac\rho2\norm{x-v^k+u^k}^2\}$, $v^{k+1}=D_t(x^{k+1}+u^k)$, $u^{k+1}=u^k+x^{k+1}-v^{k+1}$: at a fixed point $x=v$,
$\nabla f(x)+\rho u=0$ and $x=D_t(x+u)$, so $u/t\in W_t(x)$ and $0\in\nabla f(x)+\rho tW_t(x)$. Douglas--Rachford, with
$x^{k+1/2}=D_t(z^k)$, $x^{k+1}=\prox_{\gamma f}(2x^{k+1/2}-z^k)$, $z^{k+1}=z^k+x^{k+1}-x^{k+1/2}$: a fixed point has $x=D_t(z)$ and $x-z=\gamma\nabla f(x)$.
Consensus equilibrium \cite{Buzzard2018}: $\prox_{\gamma f}(x+u_1)=x$, $D_t(x+u_2)=x$, $u_1+u_2=0$ give $u_1=\gamma\nabla f(x)$, $u_2\in tW_t(x)$.
(ii) For RED, $x-D_t(x)=t\nabla E_t(x)$ by \eqref{eq:score-force}.
(iii) If $\mathcal F_{\alpha,t}(z)=0$ and $x=D_t(z)$, then $\mathcal F_{\alpha,t}(z)=A^*(Ax-y^\delta)+\alpha t^{-1}(z-x)$ with $t^{-1}(z-x)\in W_t(x)$.
Conversely, if $0=A^*(Ax-y^\delta)+\alpha w$, $w\in W_t(x)$, then $z=x+tw=x-(t/\alpha)A^*(Ax-y^\delta)$ satisfies $D_t(z)=x$ and $\mathcal F_{\alpha,t}(z)=0$.
The Jacobian follows from $\nabla W_t(D_t(z))=t^{-1}((\nabla D_t(z))^{-1}-I)$.
\end{proof}

\begin{proof}[Proof of \Cref{thm:bayes-structure}]
We record the steps that are not stated in this form in the sources.
(i) The relative entropy identity
$t\mathrm{KL}(q\|\pi)+\tfrac12\mathbb E_q\norm X^2=\inn zu-\Psi_t(z)+t\mathrm{KL}(q\|q_{t,z})$ for $m_q=u=D_t(z)$, minus $\norm u^2/2$, gives
\eqref{eq:entropic-primal} and uniqueness. For compact support, $J_t(u)=\inf_{m_q=u}\{t\mathrm{KL}(q\|\pi)+\tfrac12\mathbb E_q\norm X^2\}$ is
proper, convex and lower semicontinuous with attained infimum, by weak compactness of probabilities on $\supp\pi$ and lower
semicontinuity of entropy. The Gibbs variational formula gives $J_t^*=\Psi_t$, so $J_t=\Psi_t^*$, which is the extended-real identity.
Every feasible $q$ is supported on $\supp\pi$, so its variance is at least $\dist(u,\supp\pi)^2$.
(ii) $D_t=\nabla\Psi_t$ is injective by strict convexity, and $\nabla^2\Psi_t^*(u)=t\Sigma_t(z)^{-1}$ gives \eqref{eq:hessian-g}.
(iii) By definition of $p_t$, $\Psi_t(z)=\tfrac12\norm z^2+t\log p_t(z)+\tfrac{nt}2\log(2\pi t)$. Inserting this into
$\Psi_t^*(u)=\inn zu-\Psi_t(z)$ gives the first identity, and
$\min_v\{R_t(v)+\norm{v-z}^2/(2t)\}=t^{-1}(\norm z^2/2-\Psi_t(z))$, attained exactly at $v=\nabla\Psi_t(z)$, gives the second.
(iv) $\nabla p_t=-(\nabla g\,e^{-g})*\varphi_t$ for the weak gradient of $e^{-g}\in W^{1,1}$; divide by $p_t$.
\end{proof}

\begin{proof}[Proof of \Cref{prop:small-sigma}]
We expand the Gaussian smoothing by the heat semigroup, then transfer the expansion to the denoised coordinate through the local inverse.

Write $P_t=e^{t\Delta/2}$, so that $p_t=f*\varphi_t=P_tf$. For $\abs\beta\le2$ the semigroup remainder gives
$\norm{\partial^\beta(P_tf-f-\tfrac t2\Delta f)}_\infty\le\tfrac{t^2}8\norm{\Delta^2\partial^\beta f}_\infty$. On a compact set
$f$ is bounded below, so taking logarithms gives the $C^2$ expansion of $E_t=-\log p_t$, since
$\Delta f/f=\abs{\nabla g}^2-\Delta g$.
Differentiating gives the expansion of $\nabla E_t$ in $C^1$, with $\tfrac12\nabla\abs{\nabla g}^2=\nabla^2g\,\nabla g$.

We next control the inverse on compact sets. On a compact neighborhood $\nabla E_t$ and $\nabla^2E_t$ are bounded uniformly for small $t$, so $z=u+t\nabla E_t(z)$ has a unique
solution near $u$. By injectivity, this solution is $D_t^{-1}(u)$.
In particular, the compact set lies in $\ran D_t$, and
$D_t^{-1}=I+t\nabla g+O(t^2)$ in $C^1$. By \eqref{eq:W-score-t},
\[
 W_t(u)=\nabla E_t\bigl(u+t\nabla g(u)\bigr)+O(t^2)
 =\nabla g+t\,\nabla^2g\,\nabla g-t\,\nabla^2g\,\nabla g+\tfrac t2\nabla\Delta g+O(t^2)
\]
in $C^1$, which is the stated expansion.
\end{proof}

\begin{proof}[Proof of \Cref{thm:stationary-structure}]
For an empirical prior these are the classical properties of Gaussian mean shift; for a prior in $\PP_2$ with full affine span the
argument is the following.
(i) For $u\in\ran D_t$, $W_t(u)=t^{-1}(D_t^{-1}(u)-u)=\nabla R_t(u)$, which vanishes if and only if $D_t(u)=u$. By Tweedie's identity
$D_t(u)=u+t\nabla\log p_t(u)$, so $D_t(u)=u$ if and only if $\nabla p_t(u)=0$.
(ii) By Sylvester's law of inertia applied to \eqref{eq:congruence}, the eigenvalues $s_i$ of $\Sigma_*$ give
eigenvalues $s_i^{-1}-t^{-1}$ and $t^{-1}-t^{-2}s_i$ of the two Hessians, with the same signs, and
$\nabla^2p_t=-p_t\nabla^2E_t$ at a critical point. Since $\nabla D_t(u_*)=\Sigma_*/t$ is symmetric positive
definite with norm $\lambda_{\max}(\Sigma_*)/t$, the mean-value inequality gives local contraction with any factor
$q\in(\lambda_{\max}(\Sigma_*)/t,1)$ when $\Sigma_*\prec tI$, and a repelling direction otherwise.
\end{proof}

\begin{proof}[Computation for \Cref{prop:prior-inverse-unstable}]
Put $a_k=e^{-ck^2}\le1$ and $b_k=rk$, with $c$ and $r$ as in the example. Gaussian conditioning gives the smoothed density
$\varphi_{s^2+t}(z)(1+\epsilon a_k\cos(b_kz))/(1+\epsilon e^{-s^2k^2/2})$, and Tweedie's formula gives
$D_{t,k}(z)-D_{t,0}(z)=-t\epsilon a_kb_k\sin(b_kz)/(1+\epsilon a_k\cos(b_kz))$, so $\norm{D_{t,k}-D_{t,0}}_\infty\le t\epsilon a_kb_k/(1-\epsilon a_k)$, which is
at most the stated bound. Since $D_{t,0}(z)=rz$, $u=D_{t,k}(z)$ implies
$|z-u/r|\le\norm{D_{t,k}-D_{t,0}}_\infty/r$.
Both denoisers are onto and injective because their densities are bounded above and below by positive multiples of a Gaussian.
This gives the force bound. Finally $\mathbb E_{\pi_0}\abs{\cos(kX)}\to2/\pi$, which gives the limit of the total
variation.
\end{proof}

\section{Proofs for \texorpdfstring{\Cref{sec:preference}}{Section 3}}\label{app:proofs-preference}

\begin{proof}[Proof of \Cref{lem:dtR}]
We differentiate the conjugate representation at its maximizing noisy input, where the derivative with respect to that input vanishes.

Smoothness follows from the inverse function theorem applied to $(t,z)\mapsto(t,D_t(z))$.
Its $z$-derivative $\Sigma_t/t$ is invertible.

To compute the scale derivative, write $\Psi_t=t\log I_t$ with $I_t(z)=\mathbb E_\pi\bigl[e^{(\inn zX-\norm X^2/2)/t}\bigr]$, and let
$z=D_t^{-1}(u)$, $q=q_{t,z}$. Then $\partial_t\Psi_t(z)=\Psi_t(z)/t-\mathbb E_q[\inn zX-\norm X^2/2]/t$. Since
$\Psi_t^*(u)=\inn zu-\Psi_t(z)$ and the derivative in $z$ of $\inn zu-\Psi_t(z)$ vanishes there,
$\partial_t\Psi_t^*(u)=-\partial_t\Psi_t(z)=(\Psi_t^*(u)-\tfrac12\mathbb E_q\norm X^2)/t$, using $\mathbb E_qX=u$. Differentiating
$R_t=(\Psi_t^*-\norm u^2/2)/t$ gives $\partial_tR_t=(\norm u^2-\mathbb E_q\norm X^2)/(2t^2)=-V_t/(2t^2)$.
\end{proof}

\section{Proofs for \texorpdfstring{\Cref{sec:selection}}{Section 4}}\label{app:proofs-selection}

The variational form below prepares the exact representation in \Cref{sec:bridge} and the analysis of the symmetric mixture.
For a Gaussian mixture
$\pi=\sum_kw_k\mathcal N(m_k,C_k)$ write $M_k=C_k+tI$, $B_k=C_kM_k^{-1}$, $e_k(z)=\tfrac12\norm{z-m_k}_{M_k^{-1}}^2+c_k^R(t)$ and
$z_k(u)=m_k+B_k^{-1}(u-m_k)$, the inverse of $u$ under the $k$-th component denoiser.

\begin{lemma}[Variational form of the induced energy]\label{lem:R-sup}
For every $u\in\mathbb R^n$,
\begin{equation}\label{eq:R-sup}
 R_t(u)=\sup_{z\in\mathbb R^n}\Bigl\{-\log\sum_je^{-e_j(z)}-\frac{\norm{z-u}^2}{2t}\Bigr\},
\end{equation}
attained uniquely at $z=D_t^{-1}(u)$; for $z=z_k(u)$ the $k$-th term reproduces the single-component energy
$\tfrac12\norm{u-m_k}_{C_k^{-1}}^2+c_k^R(t)$, and \eqref{eq:mixture-exact} holds.
\end{lemma}

\begin{proof}
We rewrite the conjugate definition as a supremum and use the mixture Jacobian to identify its unique maximizer.

For every prior, $\Psi_t(z)=\tfrac12\norm z^2+t\log p_t(z)+\tfrac{nt}2\log(2\pi t)$ and the definition of $R_t$ give, for all $u$,
\begin{equation}\label{eq:R-sup-general}
 R_t(u)=\sup_z\Bigl\{E_t(z)-\frac{\norm{z-u}^2}{2t}\Bigr\}-\frac n2\log(2\pi t),
\end{equation}
so that $R_t\ge E_t-\tfrac n2\log(2\pi t)$. With $\Psi_t$ as in \Cref{thm:bayes-structure}, $R_t=(\Psi_t^*-\norm\cdot^2/2)/t$, and for the mixture
$-\log\sum_je^{-e_j(z)}=E_t(z)-\tfrac n2\log(2\pi t)=\norm z^2/(2t)-\Psi_t(z)/t$. Hence the supremum equals
$t^{-1}\sup_z\{\inn zu-\Psi_t(z)\}-\norm u^2/(2t)=R_t(u)$. The function maximized has Hessian $-\nabla D_t(z)/t$, and
$\nabla D_t=\sum_jp_jB_j+t^{-1}\Cov_p(d_j)\succeq\min_j\lambda_{\min}(B_j)I\succ0$ with $d_j(z)=m_j+B_j(z-m_j)$ and $p$ the posterior
component weights. Thus the supremum is attained exactly at $D_t^{-1}(u)$, and $D_t$ is a bijection of $\mathbb R^n$.

We recover the remaining identities by considering a single component.
For a single component $\sup_z\{e_k(z)-\norm{z-u}^2/(2t)\}$ is a concave quadratic problem with maximizer $z_k(u)$ and value
$\tfrac12\norm{u-m_k}^2_{C_k^{-1}}+c_k^R(t)$, which gives the remaining statements.
Writing the log-sum relative to its $k$-th term then gives the identity \eqref{eq:mixture-exact}.
\end{proof}

\begin{proof}[Proof of \Cref{ex:two-mode}]
We compute the posterior covariance to compare stability, then reduce the fiber minimization to the component proportions to locate the selected points.

Write $c=t/(\varepsilon+t)$, so $t=\varepsilon c/(1-c)$. Given $Z_t=z$, $X$ is the mixture of the two component posteriors
$\mathcal N(\pm ca(1,1)+(1-c)z,\ \varepsilon cI)$ with weights $w(z),1-w(z)$, so
\[
 \Sigma_t(z)=\varepsilon c\,I+\beta(z)\,(1,1)^{\!\top}(1,1),\qquad \beta(z)=4w(z)(1-w(z))c^2a^2\in[0,c^2a^2],
\]
with $\beta(0)=c^2a^2$.

(i) We begin with stability in the full space. By the symmetry $u\mapsto-u$, the origin is a critical point of $p_t$ and of $R_t$.
The eigenvalues of $\Sigma_t(0)$ are $\varepsilon c<t$ along $(1,-1)$ and $\varepsilon c+2c^2a^2$ along $(1,1)$, and the latter exceeds $t$ if and only if
$2a^2(1-c)>\varepsilon$, that is, $t<2a^2-\varepsilon$. We apply \Cref{thm:stationary-structure}(ii).

(ii) We next restrict the curvature test to the fiber. The prior has a positive density, so $\ran D_t=\mathbb R^2$. By \eqref{eq:fibre-hessian} the second derivative of
$k\mapsto R_t(0,k)$ is $\Sigma_{2\mid1}(z)^{-1}-t^{-1}$ at $z=D_t^{-1}(0,k)$, where
\[
 \Sigma_{2\mid1}=\frac{\varepsilon c(\varepsilon c+2\beta)}{\varepsilon c+\beta},\qquad
 \frac{d}{d\beta}\Sigma_{2\mid1}=\frac{(\varepsilon c)^2}{(\varepsilon c+\beta)^2}>0 .
\]
Hence $\Sigma_{2\mid1}(z)\le\Sigma_{2\mid1}(0)$ for every $z$, with equality only where $w(z)=\tfrac12$. A direct computation shows
$\Sigma_{2\mid1}(0)<t$ if and only if $c>(a^2-\varepsilon)/(2a^2)$, i.e.\ $t>t_R$, and $\Sigma_{2\mid1}(0)=t$ at $t=t_R$. For $t>t_R$ the
second derivative is positive everywhere. For $t=t_R$ it is nonnegative and vanishes only where $w(z)=\tfrac12$, i.e.\ on
$\{z_1+z_2=0\}$, whose preimage on $S_0$ is the origin alone, because $D_t$ maps this line into itself by symmetry. In both cases
$k\mapsto R_t(0,k)$ is strictly convex and even, so the origin is its unique global minimizer.

To locate the selected points below $t_R$, we use \Cref{thm:simplex}. In the model $\bar C_p=\varepsilon I$, $\bar m_p=s(a,a)$ and
$\delta_p=a^2(1-s^2)/(\varepsilon+t)$ with $s=p_+-p_-$. Minimizing over the fiber in \eqref{eq:simplex-fibre} gives the point $(0,as)$ and
the reduced objective $a^2\ell(s)+\log2+\log(1+\varepsilon/t)+b(s)$, with $\ell(s)=\frac1{\varepsilon+t}+\frac{t-\varepsilon}{2\varepsilon(\varepsilon+t)}s^2$
and $b$ the negative binary entropy of $(1\pm s)/2$. Its derivative vanishes if and only if $\operatorname{atanh}s=\kappa_as$. Since
$\operatorname{atanh}(s)/s$ increases strictly from $1$ on $(0,1)$, the minimizer is $s=0$ if and only if $\kappa_a\le1$, i.e.\
$t\ge t_R$, and otherwise the two roots $\pm s_t$ of $s=\tanh(\kappa_as)$ are the minimizers. The branch $s_t$ is continuous in $t$.
The expansion of $\tanh$ gives $s_t^2=3(\kappa_a-1)/\kappa_a^3+O((\kappa_a-1)^2)$, so the pitchfork at $t_R$ is supercritical.

(iii) For the score decision, we use the marginal variance. The second derivative of $k\mapsto E_t(0,k)$ at $0$ is $t^{-1}-t^{-2}(\varepsilon c+c^2a^2)$, negative if and only if
$a^2(1-c)>\varepsilon$, i.e.\ $t<t_E$. Since $E_t(0,k)\to\infty$ as $\abs k\to\infty$ and is even, its global minimizers exist,
are symmetric, and exclude the origin.

(iv) Finally, we compute the MAP points and the density ratio.
On $S_0$, $-\log\pi(0,k)=(2a^2+k^2)/(2\varepsilon)-\log\cosh(ak/\varepsilon)+\mathrm{const}$.
Its critical points solve $k=a\tanh(ak/\varepsilon)$.
Since $a^2>\varepsilon$, there is a unique positive root $k_G<a$, and the origin is a local
maximum of the energy of $\pi|_{S_0}$. The ratio is $\pi(0,0)/\pi(0,k_G)=e^{k_G^2/(2\varepsilon)}/\cosh(ak_G/\varepsilon)$, and
$\cosh(ak_G/\varepsilon)\ge\tfrac12e^{ak_G/\varepsilon}\ge\tfrac12e^{k_G^2/\varepsilon}$ because $k_G<a$.

(v) Combine (ii), (iii) and (iv). By the symmetry $(x_1,x_2)\mapsto(-x_1,-x_2)$ of $\pi$, the conditional law of $x_2$ given $x_1=0$ is
symmetric about $0$, so its mean is the origin.
\end{proof}

\section{Proof of \texorpdfstring{\Cref{thm:mixture-selection}}{Theorem 4.10} in setting (a)}\label{app:proofs-mixtures}

\emph{Strategy.} We localize global minimizers near compatible components and compare them with the explicit Tikhonov minimizers.
Every global minimizer has bounded energy and therefore lies near the core of a compatible component.
There the three energies agree with their single-component versions up to exponentially small errors in $C^2$.

\emph{Constants.} Throughout, $t\in[t_0,t_1]$, $0<\alpha\le\alpha_1$, $\Lambda :=c_++t_1$, $b:=c_-/(c_-+t_1)$, and
$\ell_0:=\min\{0,\tfrac n2\log(2\pi c_-)\}$. For $s\ge0$, the single-component energies are $\varphi_{k,\Phi}(u)=\tfrac12(u-m_k)^\top(C_k+s_\Phi I)^{-1}(u-m_k)+c_k^\Phi$.
Put $d(u)=\min_k\abs{u-m_k}$. We use the following constants for the energy bounds:
\[
 C_0=\log\tfrac1{w_*}+\tfrac n2\max\Bigl\{0,\log(2\pi \Lambda),\log\bigl(1+\tfrac{c_+}{t_0}\bigr)\Bigr\},\quad
 M=C_0+\frac{\norm{A^\dagger}^2B^2}{2c_-},\quad r=\sqrt{2\Lambda(M-\ell_0)},
\]
Here $w_*=\min_kw_k$. These constants bound the energies at the feasible points $m_k+A^\dagger(y-Am_k)$, $k\in J$. The exclusion condition is
\begin{equation}\label{eq:mixture-exclusion}
 \min_{j\notin J}\abs{y-Am_j}>\norm A\,r+\sqrt{2\alpha_1(M-\ell_0)}
 \qquad\bigl(\text{hard constraint: }>\norm A\,r\bigr).
\end{equation}
With $H=r+1$, $Z=H/b+1$, $\Delta_\mu=\max_{j,k}\abs{\mu_j-\mu_k}$ and $P_\lambda=\Delta_\mu\lambda+2Z$ put
\begin{equation}\label{eq:lambda0-eps}
\begin{gathered}
 \eta_\lambda=\frac1{w_*}\Bigl(\frac \Lambda {c_-}\Bigr)^{n/2}\exp\Bigl(\frac{Z^2}{2c_-}-\frac{\delta_\mu^2\lambda^2}{8\Lambda}\Bigr),\qquad
 \varepsilon_\lambda=10\bigl(1+c_-^{-1}+t_0^{-1}+b^{-1}\bigr)^4(1+P_\lambda^2)\,\eta_\lambda ,\\
 \lambda_0=\inf\bigl\{\lambda'>0:\ \delta_\mu\lambda\ge2Z,\ \eta_\lambda\le1,\ P_\lambda\eta_\lambda\le b,\ \varepsilon_\lambda\le1/(4\Lambda)\ \text{for all }\lambda\ge\lambda'\bigr\}.
\end{gathered}
\end{equation}
Each of the four conditions holds for all large $\lambda$, so $\lambda_0<\infty$. Since $1+P_\lambda^2\le\max\{2\Delta_\mu^2,1+8Z^2\}(1+\lambda^2)$,
\[
 \varepsilon_\lambda\le C_*(1+\lambda^2)e^{-\delta_\mu^2\lambda^2/(8\Lambda)},
\]
\[
 C_*=\frac{10}{w_*}\bigl(1+c_-^{-1}+t_0^{-1}+b^{-1}\bigr)^4\max\{2\Delta_\mu^2,1+8Z^2\}\Bigl(\frac \Lambda {c_-}\Bigr)^{n/2}e^{Z^2/(2c_-)},
\]
so $\varepsilon_\lambda=O(e^{-c\lambda^2})$ for every $c<\delta_\mu^2/(8\Lambda)$. All constants depend on the chosen ranges $[t_0,t_1]$,
$\alpha_1$ and $B$ and on the mixture. They are far from sharp. In \Cref{thm:mixture-selection} for setting (a),
$\theta=\alpha\varepsilon_\lambda$ and $\ell=\Lambda\varepsilon_\lambda\le\tfrac14$, and the balls in (i) have radius $H=r+1$. Step 4 shows moreover
that $F_\Phi(v_{k,\Phi})$ lies between $S_k^\Phi-\theta$ and $S_k^\Phi$, so that in setting (a) a margin $\theta$ already suffices in (ii).

\emph{Step 1: posterior formulas and a global inverse.} A uniform lower bound on the denoiser Jacobian controls its inverse.
Let $B_k=C_k(C_k+tI)^{-1}$, $d_k(z)=m_k+B_k(z-m_k)$ and $\rho_k(z)$ the
posterior component probabilities. Then $D_t=\sum_k\rho_kd_k$ and
\[
 \nabla D_t(z)=\Sigma_t(z)/t=\sum_k\rho_k(z)B_k+t^{-1}\Cov_{\rho(z)}\bigl(d_k(z)\bigr)\succeq bI .
\]
Hence $\Psi_t$ is $b$-strongly convex, $D_t$ is a diffeomorphism of $\mathbb R^n$ onto itself, and
$\abs{D_t^{-1}(u)-D_t^{-1}(u')}\le b^{-1}\abs{u-u'}$. In particular the representation of \Cref{thm:bayes-structure} holds on all of
$\mathbb R^n$.

\emph{Step 2: the minimizers lie in the cores.} We compare a global lower bound with the energy at feasible component points.
Since every component density is at most $(2\pi c_-)^{-n/2}e^{-d(u)^2/(2\Lambda)}$
after smoothing by $sI$, $s\le t_1$, we have $g,E_t\ge d^2/(2\Lambda)+\ell_0$. For the induced energy, dropping the mean constraint in
the Gibbs variational principle gives $R_t(u)\ge E_t(u)-\tfrac n2\log(2\pi t)$, and
$(2\pi t)^{n/2}p_t(u)=\sum_kw_k\det(I+C_k/t)^{-1/2}e^{-\frac12(u-m_k)^\top(C_k+tI)^{-1}(u-m_k)}\le e^{-d(u)^2/(2\Lambda)}$.
Thus $\Phi\ge d^2/(2\Lambda)+\ell_0$ for all three energies.

For the upper bound, $\Phi\le\varphi_{k,\Phi}$ for every $k$.
For $g$ and $E_t$, this follows from $\pi\ge w_k\pi_k$.
For $R_t$, we minimize $\mathrm{KL}(q\|\pi)\le\mathrm{KL}(q\|\pi_k)-\log w_k$ over $q$ of mean $u$ and use \eqref{eq:gaussian-R}.
At the feasible points of the compatible components $\Phi\le M$, so every global minimizer satisfies
$\Phi(u)\le M$, $\abs{Au-y}\le\sqrt{2\alpha(M-\ell_0)}$ and $d(u)\le r$. By \eqref{eq:mixture-exclusion} it cannot lie within $r$ of an
excluded mean. The same bounds place the single-component minimizers $x_k^{(s)}$, $k\in J$, in $\bar B(m_k,r)$ and show that
excluded components cannot have the smallest score.

\emph{Step 3: exponential $C^2$ approximation on the cores.} We bound the contribution of the other components near a given mean.
Let $r^{(s)}_{jk}(z)$ be the ratio of the weighted component
densities smoothed by $sI$. If $\abs{z-m_k}\le Z$ then $\abs{z-m_j}\ge\delta_\mu\lambda/2$ for $j\ne k$, and the eigenvalue bounds give
$\sum_{j\ne k}r^{(s)}_{jk}(z)\le\eta_\lambda$, $\abs{\nabla\log r^{(s)}_{jk}}\le P_\lambda/c_-$ and $\norm{\nabla^2\log r^{(s)}_{jk}}\le2/c_-$. Since
$g$ and $E_t$ equal $\varphi_{k,\Phi}-\log\bigl(1+\sum_{j\ne k}r^{(s)}_{jk}\bigr)$, their value, gradient and Hessian differ from those
of $\varphi_{k,\Phi}$ by at most $\eta_\lambda$, $\eta_\lambda P_\lambda/c_-$ and $\eta_\lambda(2P_\lambda^2/c_-^2+2/c_-)$ on $B(m_k,Z)$.

For the induced energy, we must also control the noisy input corresponding to a point in the core.
Fix $\abs{u-m_k}\le H$ and let $z_k=m_k+B_k^{-1}(u-m_k)$ be the inverse of $u$ under the $k$-th component denoiser.
Then $\abs{z_k-m_k}\le H/b=Z-1$.
On $B(m_k,Z)$ we use two elementary bounds.
Since $d_j(z)-z=(I-B_j)(m_j-z)$ and $\norm{I-B_j}\le1$, we have $\abs{d_j(z)-d_k(z)}\le\abs{z-m_j}+\abs{z-m_k}\le P_\lambda$.
A covariance is at most the second moment about any point, so $\Cov_\rho(d)\preceq\sum_{j\ne k}\rho_j(d_j-d_k)(d_j-d_k)^\top\preceq\eta_\lambda P_\lambda^2I$.
Hence
$\abs{D_t(z)-d_k(z)}\le P_\lambda\eta_\lambda$ and $\norm{\nabla D_t(z)-B_k}\le\eta_\lambda(1+P_\lambda^2/t_0)$. With the global inverse bound,
$z=D_t^{-1}(u)$ satisfies $\abs{z-z_k}\le b^{-1}\abs{D_t(z_k)-u}\le b^{-1}P_\lambda\eta_\lambda\le1$, so $z$ stays where component $k$ dominates
the posterior.

This control of the input lets us compare the energies.
We use the three-energy identity and $E_t=E_t^{(k)}-\log w_k-\log(1+\sum_{j\ne k}r^{(t)}_{jk})$.
The function $z\mapsto E_t^{(k)}(z)-\abs{z-u}^2/(2t)$ is a concave quadratic with Hessian $-B_k/t$.
Its maximum is attained at $z_k$ and equals $\varphi_{k,R_t}(u)+\log w_k+\tfrac n2\log(2\pi t)$ by the Gaussian case.
These facts give the exact identity \eqref{eq:mixture-exact}. Using
$\nabla R_t(u)-\nabla\varphi_{k,R_t}(u)=(z-z_k)/t$ and
$(\nabla D_t(z))^{-1}-B_k^{-1}=(\nabla D_t(z))^{-1}(B_k-\nabla D_t(z))B_k^{-1}$ with both inverses bounded by $b^{-1}$, we obtain
\[
 \abs{R_t-\varphi_{k,R_t}}\le\eta_\lambda+\frac{P_\lambda^2\eta_\lambda^2}{2t_0b^2},\quad
 \abs{\nabla R_t-\nabla\varphi_{k,R_t}}\le\frac{P_\lambda\eta_\lambda}{t_0b},\quad
 \norm{\nabla^2R_t-C_k^{-1}}\le\frac{\eta_\lambda(1+P_\lambda^2/t_0)}{t_0b^2}.
\]
All bounds of this step are at most $\varepsilon_\lambda$ on $B(m_k,H)$.

\emph{Step 4: perturbation of the Gaussian minima.} The single-component soft objective has Hessian
$A^\top A+\alpha(C_k+s_\Phi I)^{-1}\succeq\alpha \Lambda^{-1}I$. By Step 3 the mixture objective has Hessian $\succeq\alpha(2\Lambda)^{-1}I$ on
$B(m_k,H)$. As $x_k^{(s_\Phi)}\in\bar B(m_k,r)$, every point of $\partial B(m_k,H)$ is at distance at least $1$ from it and has
objective value at least the score $Q_k^{(s_\Phi)}+\alpha c_k^\Phi$ plus $\alpha/(4\Lambda)$, while the value at $x_k^{(s_\Phi)}$ is at most the
score. The minimum over the ball is therefore interior and unique.
At the minimizer $v_k$,
$(A^\top A+\alpha(C_k+s_\Phi I)^{-1})(v_k-x_k^{(s_\Phi)})=\alpha(\nabla\varphi_{k,\Phi}-\nabla\Phi)(v_k)$, so
$\abs{v_k-x_k^{(s_\Phi)}}\le \Lambda\varepsilon_\lambda$, and its value lies between the score minus $\alpha\varepsilon_\lambda$ and the score.

We can now compare the local minima globally.
Every global minimizer lies in a core of a compatible component by Step 2.
It is therefore one of the $v_k$, which proves (i) and (ii).
For the hard constraint the same argument runs on the affine space $S_y$, where the single-component Hessian restricted to
$\ker A$ is at least $\Lambda^{-1}I$. This proves (iii). \qed

\section{Proof of \texorpdfstring{\Cref{thm:mixture-selection}}{Theorem 4.10} in setting (b)}\label{app:proofs-general}

\emph{Strategy.} We extend the Gaussian comparison by rescaling near each global mode and controlling the error in the resulting quadratic problems.
The rescaling $x=x_k+\sqrt\varepsilon\xi$ turns the three energies into the single-Gaussian energies of
\eqref{eq:gaussian-R} plus $O(\sqrt\varepsilon)$ in $C^2$.
The noisy coordinate of the induced energy stays near the mode.
Quadratic growth localizes the global minimizers near compatible modes.
A perturbation of the explicit quadratic problems then gives the scores.

We use the following notation for the Gaussian offsets. We write
$c^G(C)=\tfrac12\log\det(2\pi C)$, $c^R(C,t)=\tfrac12\log\det(I+C/t)$ and $c^E(C,t)=\tfrac12\log\det(2\pi(C+tI))$, so that
$c_k^\Phi=-\log w_k+c^\Phi(C_k,t)$, and $\xi_k^*=(x_k^{(s_\Phi)}-x_k)/\sqrt\varepsilon$. Here $C$ denotes constants independent of $\varepsilon$.

Write $G^\varepsilon=-\log(d\pi_\varepsilon/dx)=g/\varepsilon-\log\rho_\varepsilon+\log Z_\varepsilon$, and $R^\varepsilon$, $E^\varepsilon$ for the induced and smoothed
energies at $t=\varepsilon\tau$. Let $d(x)=\min_k\norm{x-x_k}$ over the global minimizers, $a_\varepsilon=\alpha/\varepsilon\le a_1$, and
$T_G=T_R=H_k^{-1}$, $T_E=H_k^{-1}+\tau I$ (for each mode $k$). All constants are independent of $\varepsilon$.

\emph{Step 1: quadratic control and masses.} Continuity, quadratic growth and nondegeneracy of the finitely many zeros give $c>0$ with
$g\ge c\,d^2$ on $\mathbb R^n$. Near $x_k$, $g(x_k+\sqrt\varepsilon\eta)/\varepsilon=\tfrac12\eta^\top H_k\eta+O(\sqrt\varepsilon\abs\eta^3)$ for $\sqrt\varepsilon\abs\eta\le r_0$,
and $g(x_k+h)\ge c\abs h^2$ there. With the $C^2$ bound on $\log\rho_\varepsilon$ this gives
$\abs{\rho_\varepsilon(x_k+\sqrt\varepsilon\eta)e^{-g(x_k+\sqrt\varepsilon\eta)/\varepsilon}-\rho_\varepsilon(x_k)e^{-\eta^\top H_k\eta/2}}\le C\sqrt\varepsilon(\abs\eta+\abs\eta^3)e^{-c\abs\eta^2}$,
and integration gives $w_k=(2\pi\varepsilon)^{n/2}\rho_\varepsilon(x_k)Z_\varepsilon^{-1}\det H_k^{-1/2}(1+O(\sqrt\varepsilon))$.
The mass outside $\bigcup U_k$
is $O(e^{-c'/\varepsilon})$ relative to $Z_\varepsilon$.

\emph{Step 2: the prior and smoothed energies near a mode.} We use Taylor expansion for the prior energy and Gaussian integration for the smoothed energy.
Since $-\log w_k+\tfrac12\log\det(2\pi C_k)=\log Z_\varepsilon-\log\rho_\varepsilon(x_k)+O(\sqrt\varepsilon)$
and $\nabla_\xi\log\rho_\varepsilon(x_k+\sqrt\varepsilon\xi)=O(\sqrt\varepsilon)$, Taylor's formula gives, on every ball $B_M$,
$G^\varepsilon(x_k+\sqrt\varepsilon\xi)=\tfrac12\xi^\top H_k\xi-\log w_k+c^G(C_k)+O_{C^2(B_M)}(\sqrt\varepsilon)$, with $c^G(C)=\tfrac12\log\det(2\pi C)$.

For the smoothed energy, write $z=x_k+\sqrt\varepsilon\zeta$. Then
\[
 e^{-E^\varepsilon(z)}=\frac{\varepsilon^{n/2}}{Z_\varepsilon(2\pi\varepsilon\tau)^{n/2}}\int\rho_\varepsilon(x_k+\sqrt\varepsilon\eta)\,
 \exp\Bigl[-\frac{g(x_k+\sqrt\varepsilon\eta)}{\varepsilon}-\frac{\abs{\eta-\zeta}^2}{2\tau}\Bigr]d\eta .
\]
By Step 1 the integrand may be replaced by its Gaussian approximation with an error $O(\sqrt\varepsilon)$ in $L^1$, uniformly for $\zeta\in B_M$.
Derivatives up to order two in $\zeta$ only insert polynomials of degree at most two in $\eta-\zeta$, controlled by the same Gaussian bound.
Contributions from outside a fixed neighborhood of $x_k$, including the other modes, are exponentially small, since there
$g\ge c\,d^2$ and $\abs{\eta-\zeta}\ge c/\sqrt\varepsilon$ or $g\ge c$. The Gaussian integral is bounded below on $B_M$, so by the quotient rule
the logarithm and its first two derivatives inherit the $O(\sqrt\varepsilon)$ error:
$E^\varepsilon(x_k+\sqrt\varepsilon\zeta)=\tfrac12\zeta^\top T_E^{-1}\zeta-\log w_k+c^E(C_k,t)+O_{C^2(B_M)}(\sqrt\varepsilon)$.

\emph{Step 3: the noisy coordinate of the induced energy.} We invert the local Gaussian approximation before applying the energy identity.
By Tweedie's identity, the rescaled denoiser
$\mathcal D(\zeta)=(D_t(x_k+\sqrt\varepsilon\zeta)-x_k)/\sqrt\varepsilon=\zeta-\tau\nabla_\zeta[E^\varepsilon(x_k+\sqrt\varepsilon\zeta)]$ satisfies
$\mathcal D(\zeta)=B_k\zeta+O_{C^1(B_M)}(\sqrt\varepsilon)$ with $B_k=(I+\tau H_k)^{-1}$, by Step 2. For bounded $\xi$, the equation $\mathcal D(\zeta)=\xi$
reads $\zeta=B_k^{-1}\xi-B_k^{-1}r_\varepsilon(\zeta)$ with $r_\varepsilon(\zeta)=\mathcal D(\zeta)-B_k\zeta=O_{C^1}(\sqrt\varepsilon)$ on a fixed ball $B_{M'}$
containing the unit balls about all $B_k^{-1}\xi$, $\xi\in B_M$.
For small $\varepsilon$, the right-hand side is a contraction of the unit ball about $B_k^{-1}\xi$ into itself.
Its fixed point satisfies $\zeta_\varepsilon(\xi)=B_k^{-1}\xi+O(\sqrt\varepsilon)$ and $D_\xi\zeta_\varepsilon=B_k^{-1}+O(\sqrt\varepsilon)$.
The denoiser $D_t$ is a bijection of $\mathbb R^n$ by Step 2 of the proof of \Cref{cor:hull} and positivity of the posterior covariance.
Thus this fixed point gives the actual inverse: $D_t^{-1}(x_k+\sqrt\varepsilon\xi)=x_k+\sqrt\varepsilon(I+\tau H_k)\xi+O(\varepsilon)$.

We now transfer the approximation to the induced energy.
We combine the three-energy identity
\eqref{eq:three-energies} with $\tfrac12\zeta_0^\top T_E^{-1}\zeta_0-\abs{\zeta_0-\xi}^2/(2\tau)=\tfrac12\xi^\top H_k\xi$ for $\zeta_0=B_k^{-1}\xi$.
We also use $c^E(C,t)-\tfrac n2\log(2\pi t)=c^R(C,t)$.
Together these give the value of $R^\varepsilon(x_k+\sqrt\varepsilon\xi)$.
For the derivatives,
$\nabla_\xi R^\varepsilon(x_k+\sqrt\varepsilon\xi)=\sqrt\varepsilon W_t=(\zeta_\varepsilon(\xi)-\xi)/\tau$ and $\nabla_\xi^2R^\varepsilon=(D_\xi\zeta_\varepsilon-I)/\tau=H_k+O(\sqrt\varepsilon)$. Hence
$R^\varepsilon(x_k+\sqrt\varepsilon\xi)=\tfrac12\xi^\top H_k\xi-\log w_k+c^R(C_k,t)+O_{C^2(B_M)}(\sqrt\varepsilon)$.

\emph{Step 4: global minimizers lie near compatible modes.} With $b_G=b_E=\log Z_\varepsilon$ and $b_R=\log Z_\varepsilon-\tfrac n2\log(2\pi t)$, all three
energies satisfy $\Phi-b_\Phi\ge c'd^2/\varepsilon-C$.
For $G^\varepsilon$ this follows from Step 1.
For $E^\varepsilon$ we convolve $\rho_\varepsilon e^{-g/\varepsilon}\le\rho_+\sum_ke^{-c\abs{x-x_k}^2/\varepsilon}$ with $\mathcal N(0,tI)$.
For $R^\varepsilon$ we use $R_t\ge E_t-\tfrac n2\log(2\pi t)$, which holds for every prior by evaluating the supremum in \eqref{eq:R-sup-general} at $z=u$.

We can remove the part of the datum that is orthogonal to the measurement range.
Let $P$ be the projector onto $\ran A$. Since
$\ran(AS_kA^\top)\subseteq\ran A$, $(AS_kA^\top+\alpha I)^{-1}$ acts as $\alpha^{-1}$ on $(\ran A)^\perp$, so $Q_k^{(s)}(y)=\tfrac12\norm{(I-P)y}^2+Q_k^{(s)}(Py)$,
and $\tfrac12\norm{Au-y}^2=\tfrac12\norm{(I-P)y}^2+\tfrac12\norm{Au-Py}^2$: the common term cancels and we may take $b_\varepsilon\in\ran A$.

Feasible test points now localize the minimizers. For $k\in J$ the point $x_k+\sqrt\varepsilon A^\dagger b_\varepsilon$ is feasible with $F_\Phi/\alpha-b_\Phi\le C$ by Steps 2--3. Hence every global minimizer
has $d(u)\le C\sqrt\varepsilon$ and $\norm{Au-y}\le C\sqrt\alpha\le C\sqrt\varepsilon$.
Its nearest mode $x_j$ has $\norm{Ax_j-y_0}\le C\sqrt\varepsilon$, so
$j\in J$ for small $\varepsilon$. The same lower bounds give coercivity and existence.

\emph{Step 5: perturbation of the quadratic problems.} Near $x_k$, $k\in J$, using $Ax_k=y_0$,
\[
 \alpha^{-1}F_\Phi(x_k+\sqrt\varepsilon\xi)=\frac{\norm{A\xi-b_\varepsilon}^2}{2a_\varepsilon}+\tfrac12\xi^\top T_\Phi^{-1}\xi-\log w_k+c^\Phi(C_k,t)+O_{C^2(B_M)}(\sqrt\varepsilon),
\]
whose quadratic part has Hessian $A^\top A/a_\varepsilon+T_\Phi^{-1}\succeq T_\Phi^{-1}$, uniformly even as $a_\varepsilon\to0$. With $C_k=\varepsilon H_k^{-1}$ and
$s_E=\varepsilon\tau$, the minimum value and minimizer of the quadratic part are $Q_k^{(s_\Phi)}/\alpha$ and $(x_k^{(s_\Phi)}-x_k)/\sqrt\varepsilon$.
The latter is bounded by comparison with $A^\dagger b_\varepsilon$.
Choose a fixed rescaled ball that contains this minimizer with a margin and also contains the localized global minimizers of Step 4.
The $C^2$ error preserves strict convexity on this ball.
Comparison with the boundary places the unique minimizer in the interior. At this minimizer,
$(A^\top A/a_\varepsilon+T_\Phi^{-1})(\xi-\xi_k^*)=O(\sqrt\varepsilon)$, so $\abs{\xi-\xi_k^*}\le C\sqrt\varepsilon$.
This gives a location error $C\varepsilon$, and the value is within $C\alpha\sqrt\varepsilon$ of the score. By Step 4 these branches contain every global minimizer, and comparing their values gives
(i) and (ii). Replacing $w_k$ by its Laplace value changes $\alpha\log w_k$ by $O(\alpha\sqrt\varepsilon)$.

\emph{Step 6: hard constraint.} For $y\in\ran A$, the same localization uses feasible test points without a data term.
In rescaled coordinates the constraint is $A\xi=b_\varepsilon$.
The quadratic Hessian restricted to $\ker A$ is $T_\Phi^{-1}$.
Constrained minimization gives the pseudoinverse formulas, with location error $C\varepsilon$ and energy error $C\sqrt\varepsilon$.
These are the bounds of \Cref{thm:mixture-selection} in setting (b), with $\theta=\kappa_b\alpha\sqrt\varepsilon$ and $\ell=\kappa_b\varepsilon$ for a
constant $\kappa_b$, and with balls of radius of order $\sqrt\varepsilon$. \qed

\section{Proofs for \texorpdfstring{\Cref{sec:bridge}}{Section 4.3}}\label{app:proofs-bridge}

We now establish the exact representation used above for Gaussian mixtures, the properties of the proximal hull, and the
proximal-hull theorem that explains bridge decisions.

\begin{proof}[Proof of \Cref{thm:simplex}]
We combine the variational representation with entropy duality, exchange the extrema, and solve the resulting quadratic problem.

Write $q_j(z)=\tfrac12\norm{z-m_j}^2_{M_j^{-1}}$, so that $e_j=q_j+c_j^R$ in \Cref{lem:R-sup}. We use the entropy duality
$-\log\sum_je^{-b_j}=\min_{p\in\Delta_L}\sum_jp_j(b_j+\log p_j)$.
Its gap is $\mathrm{KL}(p\|\rho)$ with $\rho_j\propto e^{-b_j}$.
This turns \eqref{eq:R-sup} into $R_t(u)=\sup_z\min_p\Phi(z,p)$ with
$\Phi(z,p)=\sum_jp_jq_j(z)+H_R(p)-\norm{z-u}^2/(2t)$.

We compute the supremum for fixed proportions. For fixed $p$, $\sum_jp_jq_j(z)=\tfrac12\norm{z-\bar m_p}^2_{\bar M_p^{-1}}+\delta_p$, and
since $\bar M_p^{-1}=\sum_jp_jM_j^{-1}\preceq(c_-+t)^{-1}I$ with $c_-=\min_j\lambda_{\min}(C_j)$, we get $\bar C_p\succeq c_-I$ and $\Phi(\cdot,p)$ is
strictly concave. Completing the square gives
\[
 \sup_z\Phi(z,p)=\delta_p+\tfrac12\norm{u-\bar m_p}^2_{\bar C_p^{-1}}+H_R(p),
\]
with attainment at $\zeta_p=u+t\bar C_p^{-1}(u-\bar m_p)$.

We can now exchange the extrema.
The function $\Phi$ is concave in $z$ and convex in $p$ on the compact convex simplex, so Sion's
minimax theorem gives $\sup_z\min_p=\min_p\sup_z$, which is \eqref{eq:simplex}. At the saddle point, $z=D_t^{-1}(u)$ by \Cref{lem:R-sup}.
The minimizing $p$ is the minimizer of the entropy duality at $b_j=e_j(z)$, i.e.\ $p_j\propto w_j e^{-q_j(z)}\det(I+C_j/t)^{-1/2}$, the
posterior component probabilities. It is unique and interior.

For the fiber, we minimize the quadratic
$\tfrac12\norm{u-\bar m_p}^2_{\bar C_p^{-1}}$ over $S_y$ by Gaussian conditioning and interchange the two minima.
\end{proof}

\begin{proof}[Proof of \Cref{lem:hull}]
Put $F=f+\norm\cdot^2/(2s)$. Since $f$ is real-valued and bounded below, $F^{**}$ is a finite convex function with $F^{**}\le F$.
(i) Completing the square gives $e_sf(z)=\norm z^2/(2s)-F^*(z/s)$. Hence
$\sup_z\{e_sf(z)-\norm{z-u}^2/(2s)\}=\sup_z\{\inn zu/s-F^*(z/s)\}-\norm u^2/(2s)=F^{**}(u)-\norm u^2/(2s)=h_sf(u)$.
Taking $z=u$ gives $e_sf\le h_sf$, and $F^{**}\le F$ gives $h_sf\le f$. A function $\phi\le f$ with $\phi+\norm\cdot^2/(2s)$ convex
satisfies $\phi+\norm\cdot^2/(2s)\le F^{**}$, since $F^{**}$ is the largest convex minorant of $F$.
(ii) If $F^{**}(u)=F(u)$, pick $p\in\partial F^{**}(u)$, which is nonempty because $F^{**}$ is finite and convex.
Then $F(x)\ge F(u)+\inn p{x-u}$ for all $x$, which says that $u$ minimizes $f(x)+\norm{x-sp}^2/(2s)$. Conversely such a minimizer gives an
affine minorant of $F$ touching at $u$, hence contact. If $f$ is differentiable at $u$, necessarily $p=\nabla f(u)+u/s$, which is the
stated inequality. If $f\ge0$ is lower semicontinuous and $h_sf(u)=0$, then $e_sf(u)=0$, so there are $x_k$ with
$f(x_k)+\norm{x_k-u}^2/(2s)\to0$. Thus $x_k\to u$ and $f(u)\le\liminf f(x_k)=0$. The converse follows from $0\le e_sf\le h_sf\le f$.
(iii) Put $\ell_h=\inf_Sh_sf$, $\ell_f=\inf_Sf$, $S_h=\argmin_Sh_sf$ and $S_f=\argmin_Sf$. Then $\ell_h\le\ell_f$ follows from $h_sf\le f$. If $\ell_h=\ell_f$ and $v\in S_f$, then $\ell_h\le h_sf(v)\le f(v)=\ell_f$. If $\ell_h<\ell_f$, a point of
$S_h$ with $h_sf=f$ would have value at least $\ell_f$.
\end{proof}

The following two rules are used to pass between the scales of \eqref{eq:prior-family} and \eqref{eq:low-temp} and to compare
nearby energies.

\begin{lemma}[Scaling and stability]\label{lem:hull-calc}
Let $f:\mathbb R^n\to\mathbb R$ be bounded below and $s>0$.
\begin{enumerate}
\item[(i)] For $c>0$ and $\lambda>0$, $h_s(cf)=c\,h_{cs}f$ and $h_s\bigl[f(\cdot/\lambda)\bigr]=(h_{s/\lambda^2}f)(\cdot/\lambda)$.
 Hence $h_s\bigl[\lambda^2f(\cdot/\lambda)\bigr]=\lambda^2(h_sf)(\cdot/\lambda)$. The same relations hold for $e_s$.
\item[(ii)] $h_s$ and $e_s$ are monotone, that is, $f_1\le f_2$ implies $h_sf_1\le h_sf_2$ and $e_sf_1\le e_sf_2$, and
 they commute with the addition of constants. Consequently, if $\abs{f_1-f_2}\le\eta$ on $\mathbb R^n$, then
 $\abs{h_sf_1-h_sf_2}\le\eta$ and $\abs{e_sf_1-e_sf_2}\le\eta$.
\end{enumerate}
\end{lemma}
\begin{proof}
(i) For $c>0$, $(c\phi)^{**}=c\,\phi^{**}$, and $cf+\norm\cdot^2/(2s)=c\bigl(f+\norm\cdot^2/(2cs)\bigr)$, which gives $h_s(cf)=c\,h_{cs}f$.
For an invertible linear map $L$, $(\phi\circ L)^{**}=\phi^{**}\circ L$, and $f(x/\lambda)+\norm x^2/(2s)=\bigl(f+\norm\cdot^2/(2s/\lambda^2)\bigr)(x/\lambda)$,
which gives the second relation. Combining the two, $h_s[\lambda^2f(\cdot/\lambda)]=\lambda^2h_{\lambda^2s}[f(\cdot/\lambda)]=\lambda^2(h_sf)(\cdot/\lambda)$.
For $e_s$ the same relations follow by the substitutions $x\mapsto\lambda x$ in the infimum.
(ii) With $F_i=f_i+\norm\cdot^2/(2s)$, conjugation reverses order twice, so $F_1\le F_2$ implies $F_1^{**}\le F_2^{**}$, and $(F+c)^{**}=F^{**}+c$; for $e_s$ both
properties are immediate. If $\abs{f_1-f_2}\le\eta$, then $f_1\le f_2+\eta$ gives $h_sf_1\le h_sf_2+\eta$, and symmetrically.
\end{proof}

\begin{proof}[Proof of \Cref{cor:hull}]
\emph{Strategy.} We pass to the rescaled family \eqref{eq:low-temp}, apply a uniform Laplace principle to its log-partition and
localize the maximizers so that one conjugation gives the proximal hull. A final step returns to the MAP energy in the original units.
For mixtures, the exact representation of \Cref{thm:simplex} gives the non-asymptotic bounds.

\emph{Step 0: substitution.} Let $V=X/\lambda$, so $V\sim\pi_\varepsilon$ with $\varepsilon=\lambda^{-2}$, and let $\Psi^\varepsilon$, $D^\varepsilon$, $R^\varepsilon$, $E^\varepsilon$
be the log-partition, denoiser, induced and smoothed energies of $\pi_\varepsilon$ at variance $\varepsilon t$. For $z=\lambda\zeta$,
$(\inn z{\lambda V}-\lambda^2\norm V^2/2)/t=(\inn\zeta V-\norm V^2/2)/(\varepsilon t)$, so \eqref{eq:entropic-logpartition} gives
$\Psi_t(\lambda\zeta)=\lambda^2\Psi^\varepsilon(\zeta)$. Conjugation gives $\Psi_t^*(\lambda v)=\lambda^2(\Psi^\varepsilon)^*(v)$, and hence
$R_t(\lambda v)=\lambda^2t^{-1}\bigl((\Psi^\varepsilon)^*(v)-\norm v^2/2\bigr)=R^\varepsilon(v)$. Equivalently, in \eqref{eq:entropic-primal} the relative
entropy is invariant under the bijection $x=\lambda v$, and $\norm{x-u}^2/(2t)=\norm{v-w}^2/(2\varepsilon t)$ for $u=\lambda w$.
The smoothed density transforms as a density, $p_t(\lambda\zeta)=\lambda^{-n}p^\varepsilon(\zeta)$, so $E_t(\lambda\zeta)=E^\varepsilon(\zeta)+n\log\lambda$.

Throughout Steps 1--3, $0<\varepsilon\le1$ and $F_t(x)=tg(x)+\norm x^2/2$. Since $\rho_-\le\rho_\varepsilon\le\rho_+$, replacing $\rho_\varepsilon$ by $1$
changes the unnormalized log-partition $\widehat\Psi^\varepsilon$ below by at most $\varepsilon tm_\rho$ and $\varepsilon\log Z_\varepsilon$ by at most $\varepsilon m_\rho$,
with $m_\rho=\max\{\abs{\log\rho_-},\abs{\log\rho_+}\}$, uniformly in all variables.
Conjugation preserves uniform additive bounds, so we may take $\rho_\varepsilon\equiv1$.

\emph{Step 1: Laplace principle.} Put $\widehat\Psi^\varepsilon=\Psi^\varepsilon+\varepsilon t\log Z_\varepsilon$. By substitution,
\[
 \widehat\Psi^\varepsilon(z)=\varepsilon t\log\int\exp\Bigl(\frac{\inn zx-F_t(x)}{\varepsilon t}\Bigr)dx .
\]
For $z$ in a compact set, $F_t(x)-\inn zx\ge(tc_0+\tfrac12)\norm x^2-\norm z\norm x-tc_1$ is coercive uniformly in $z$, so the
infimum $-F_t^*(z)$ is attained in a fixed ball. On that ball $F_t$ is uniformly continuous, so for every $\eta>0$ the exponent is
within $\eta$ of its maximum on a ball of radius $r(\eta)>0$ independent of $z$.
Outside a large ball the integral is bounded by a
Gaussian tail, uniformly in $z$. Taking $\varepsilon t\log$ and letting $\varepsilon\to0$ and then $\eta\to0$ gives
$\widehat\Psi^\varepsilon\to F_t^*$ locally uniformly. The same argument gives $\varepsilon\log Z_\varepsilon\to-\min g=0$.
Since $\varepsilon E^\varepsilon(u)-\varepsilon\log Z_\varepsilon-\tfrac{n\varepsilon}2\log(2\pi\varepsilon t)=-\varepsilon\log\int\exp[-(g(x)+\norm{x-u}^2/(2t))/\varepsilon]\,dx$,
it also gives $\varepsilon E^\varepsilon\to e_tg$ locally uniformly.

\emph{Step 2: one conjugation.} We localize the maximizers uniformly so that the limit passes through conjugation.
Fix $M$ and $\norm u\le M$.
For $z\neq0$, consider the ball of radius $\tfrac12$ about $(M+2)z/\norm z$.
On this ball, $\inn zx\ge(M+\tfrac32)\norm z$ and $F_t$ is bounded by a constant depending only on $M$.
Integration over the ball gives $\widehat\Psi^\varepsilon(z)\ge(M+\tfrac32)\norm z-C_M$ for all $z$ and $\varepsilon\le1$. Hence $\inn uz-\widehat\Psi^\varepsilon(z)\le-\tfrac12\norm z+C_M$,
and since $\widehat\Psi^\varepsilon(0)$ is bounded, the maximizers in $(\widehat\Psi^\varepsilon)^*(u)$ lie in a fixed ball for $\norm u\le M$, as do
those of $F_t^{**}(u)$. Locally uniform convergence passes through the suprema: $(\widehat\Psi^\varepsilon)^*\to F_t^{**}$ locally
uniformly. Since $(\Psi^\varepsilon)^*=(\widehat\Psi^\varepsilon)^*+\varepsilon t\log Z_\varepsilon$ and $\varepsilon t\log Z_\varepsilon\to0$ by Step 1,
\[
 \varepsilon R^\varepsilon=t^{-1}\bigl((\Psi^\varepsilon)^*-\tfrac12\norm\cdot^2\bigr)\longrightarrow t^{-1}\bigl(F_t^{**}-\tfrac12\norm\cdot^2\bigr)
 =\Bigl(g+\frac{\norm\cdot^2}{2t}\Bigr)^{**}-\frac{\norm\cdot^2}{2t}=h_tg
\]
locally uniformly. The prior has a positive density, so $D^\varepsilon$ is a diffeomorphism of $\mathbb R^n$ onto itself
and \Cref{thm:bayes-structure} applies on all of $\mathbb R^n$.

\emph{Step 2$'$: the MAP energy and the original units.} By Step 0, $\lambda^{-2}R_t(\lambda v)=\varepsilon R^\varepsilon(v)\to h_tg(v)$ and
$\lambda^{-2}E_t(\lambda v)=\varepsilon E^\varepsilon(v)+\varepsilon n\log\lambda\to e_tg(v)$, locally uniformly. Put $f_\lambda(v)=\lambda^{-2}G_\lambda(\lambda v)$, so that
$G_\lambda=\lambda^2f_\lambda(\cdot/\lambda)$ and, by \Cref{lem:hull-calc}(i), $h_t(G_\lambda)(\lambda v)=\lambda^2h_tf_\lambda(v)$ and $e_t(G_\lambda)(\lambda v)=\lambda^2e_tf_\lambda(v)$. Since
$Z_\lambda=\lambda^nZ_\varepsilon$,
\[
 f_\lambda(v)-g(v)=-\lambda^{-2}\log\rho_\lambda(v)+\varepsilon\log Z_\varepsilon+\varepsilon n\log\lambda\longrightarrow0
\]
uniformly on $\mathbb R^n$, by the bounds on $\rho_\lambda$ and Step 1. By \Cref{lem:hull-calc}(ii), $h_tf_\lambda\to h_tg$ and $e_tf_\lambda\to e_tg$
uniformly. Together these give $\lambda^{-2}\bigl(R_t(\lambda v)-h_t(G_\lambda)(\lambda v)\bigr)\to0$ and the same for $E_t$, locally uniformly in $v$,
which is (i).

\emph{Step 3: fibers.} A uniform coercivity bound keeps the fiber minimizers in a bounded set.
With $a=tc_0+\tfrac12$, Gaussian integration gives
$\widehat\Psi^\varepsilon(z)\le\norm z^2/(4a)+tc_1+\tfrac{n\varepsilon t}2\log(\pi\varepsilon t/a)$.
Conjugating gives
$\varepsilon R^\varepsilon(u)\ge c_0\norm u^2-c_1+\varepsilon\log Z_\varepsilon-\tfrac{n\varepsilon}2\log(\pi\varepsilon t/a)$, and the last two terms are bounded and
tend to zero. Thus the $\varepsilon R^\varepsilon$ are equicoercive, and $h_tg\ge c_0\norm\cdot^2-c_1$ since $(c_0+\tfrac1{2t})\norm\cdot^2-c_1$ is a
convex minorant of $g+\norm\cdot^2/(2t)$. By Step 0, $u$ minimizes $R_t$ on $S_y$, $y=\lambda\nu$, if and only if $u/\lambda$ minimizes $R^\varepsilon$ on
$\{Av=\nu\}$. Minimizers of the continuous coercive function $R^\varepsilon$ on this fiber exist.
Comparison with a fixed feasible point bounds them.
Locally uniform convergence and continuity of $h_tg$ show that every accumulation point minimizes $h_tg$ on $\{Av=\nu\}$.
This proves (ii); the core and bridge alternatives are \Cref{lem:hull}(iii).

\emph{Part (iii): Gaussian mixtures.} We work in the original coordinates; no separation is used.
Write $\gamma_j(u)=\tfrac12\norm{u-m_j}^2_{C_j^{-1}}$, $\Gamma=\min_j\gamma_j$ and $q_j(\zeta)=\tfrac12\norm{\zeta-m_j}^2_{M_j^{-1}}$, $M_j=C_j+tI$. The infimal convolution of
$\gamma_j$ with $\norm\cdot^2/(2t)$ is $q_j$, that is, $q_j=e_t\gamma_j$, and since the infimum commutes with $\min_j$,
$\min_jq_j=e_t\Gamma$.

\emph{Induced energy.} By \Cref{thm:simplex}, $R_t(u)=\min_p\{\mathcal A(p,u)+H_R(p)\}$ with
$\mathcal A(p,u)=\delta_p+\tfrac12\norm{u-\bar m_p}^2_{\bar C_p^{-1}}$. As in the proof of \Cref{thm:simplex}, completing the square gives
$\mathcal A(p,u)=\sup_\zeta\{\sum_jp_jq_j(\zeta)-\norm{\zeta-u}^2/(2t)\}$. The function maximized is concave in $\zeta$, since its Hessian is
$\bar M_p^{-1}-t^{-1}I\prec0$, and linear in $p$ on the compact simplex. Sion's minimax theorem gives
\[
 \min_p\mathcal A(p,u)=\sup_\zeta\Bigl\{\min_jq_j(\zeta)-\frac{\norm{\zeta-u}^2}{2t}\Bigr\}=\sup_\zeta\Bigl\{e_t\Gamma(\zeta)-\frac{\norm{\zeta-u}^2}{2t}\Bigr\}=h_t\Gamma(u),
\]
the last equality by \Cref{lem:hull}(i). Since $\sum_jp_jc_j^R(t)\in[\min_jc_j^R(t),\max_jc_j^R(t)]$ and $\sum_jp_j\log p_j\in[-\log L,0]$,
$H_R(p)\in[\min_jc_j^R(t)-\log L,\ \max_jc_j^R(t)]$, which gives
\begin{equation}\label{eq:hull-mixture-gamma}
 \min_jc_j^R(t)-\log L\le R_t-h_t\Gamma\le\max_jc_j^R(t).
\end{equation}

\emph{Smoothed and prior energies.} $p_t=\sum_jw_j\mathcal N(m_j,M_j)$ gives $E_t=-\log\sum_je^{-(q_j+c_j^E(t))}$, and
$\pi=\sum_jw_j\mathcal N(m_j,C_j)$ gives $-\log\pi=-\log\sum_je^{-(\gamma_j+c_j^G)}$. The elementary bounds
$\min_ja_j-\log L\le-\log\sum_je^{-a_j}\le\min_ja_j$, together with $\min_jq_j+\min_jc_j\le\min_j(q_j+c_j)\le\min_jq_j+\max_jc_j$ and
the same for $\gamma_j$, give
\[
 \min_jc_j^E(t)-\log L\le E_t-e_t\Gamma\le\max_jc_j^E(t),\qquad \min_jc_j^G-\log L\le-\log\pi-\Gamma\le\max_jc_j^G .
\]

\emph{Comparison with the MAP energy.} Put $a=\min_jc_j^G-\log L$ and $b=\max_jc_j^G$, so $\Gamma+a\le-\log\pi\le\Gamma+b$. By the
monotonicity and the commutation with constants of \Cref{lem:hull-calc}(ii), $h_t\Gamma+a\le h_t(-\log\pi)\le h_t\Gamma+b$. With the bounds
for $R_t$,
\[
 \min_ic_i^R(t)-\max_jc_j^G-\log L\le R_t-h_t(-\log\pi)\le\max_ic_i^R(t)-\min_jc_j^G+\log L,
\]
which gives \eqref{eq:hull-mixture-bound}. The same argument with $e_t$ gives the bound for $E_t$.

\end{proof}

\begin{lemma}[Contact for Gaussian mixtures]\label{lem:mixture-contact}
In the setting of \Cref{cor:hull}(iii), let $\Gamma=\min_j\gamma_j$ with $\gamma_j(u)=\tfrac12\norm{u-m_j}^2_{C_j^{-1}}$, let
$q_j(\zeta)=\tfrac12\norm{\zeta-m_j}^2_{(C_j+tI)^{-1}}$ and $D_j=\{\zeta:q_j(\zeta)\le q_i(\zeta)\text{ for all }i\}$. Then $h_t\Gamma(u)=\Gamma(u)$
if and only if $z_j(u):=u+tC_j^{-1}(u-m_j)\in D_j$ for some $j$.
\end{lemma}
\begin{proof}
By the proof of \Cref{cor:hull}(iii), $h_t\Gamma(u)=\sup_\zeta\{\min_jq_j(\zeta)-\norm{\zeta-u}^2/(2t)\}$. The maximizer $\zeta(u)$ is
unique because every $q_j-\norm{\cdot-u}^2/(2t)$ is uniformly strongly concave. For a single quadratic the maximizer is
$z_j(u)=u+tC_j^{-1}(u-m_j)$, and the maximum is $h_t\gamma_j(u)=\gamma_j(u)$, since $\gamma_j$ is convex. If $z_j(u)\in D_j$, evaluating at $z_j(u)$
gives $h_t\Gamma(u)\ge q_j(z_j(u))-\norm{z_j(u)-u}^2/(2t)=\gamma_j(u)\ge\Gamma(u)$, and $h_t\Gamma\le\Gamma$ gives contact. Conversely, let
$h_t\Gamma(u)=\Gamma(u)$ and let $j$ attain $\Gamma(u)=\gamma_j(u)$. Then
\[
\begin{aligned}
 \Gamma(u)&=\sup_\zeta\Bigl\{q_j(\zeta)-\tfrac{\norm{\zeta-u}^2}{2t}\Bigr\}\ge q_j(\zeta(u))-\tfrac{\norm{\zeta(u)-u}^2}{2t}\\
 &\ge\min_iq_i(\zeta(u))-\tfrac{\norm{\zeta(u)-u}^2}{2t}=h_t\Gamma(u)=\Gamma(u).
\end{aligned}
\]
Both inequalities are therefore equalities.
Thus $q_j(\zeta(u))=\min_iq_i(\zeta(u))$, and $\zeta(u)$ maximizes the strictly concave function $q_j-\norm{\cdot-u}^2/(2t)$.
Its maximizer is $z_j(u)$. Hence $z_j(u)=\zeta(u)\in D_j$.
\end{proof}

\section{Proofs for \texorpdfstring{\Cref{sec:recovery}}{Section 5}}\label{app:proofs-recovery}

We turn from the selected points to their distance from the truth, proving the bias bounds, the local error expansion and the oracle scale.

\begin{proof}[Proof of \Cref{thm:bias}]
We use monotonicity to bound displacement along the fiber.

(i) We begin with existence and uniqueness.
Under the constraint qualification, $W_t+N_{\mathcal S}$ is maximally monotone by the sum theorem for maximally monotone operators \cite[Ch.~25]{BauschkeCombettes2017}.
It is $c$-strongly monotone since $N_{\mathcal S}$ is monotone.
A strongly monotone maximally monotone operator is surjective \cite[Ch.~23]{BauschkeCombettes2017}.
Thus $0\in W_t(x^\dagger_t)+N_{\mathcal S}(x^\dagger_t)$ has a solution, which is unique by strong monotonicity.

For the bound, let $v^\dagger\in W_t(x^\dagger_t)\cap N$ and
$d:=x^*-x^\dagger_t\in K$. Strong monotonicity at the pair $(x^*,x^\dagger_t)$ gives, for any $w^*\in W_t(x^*)$,
$c\norm{d}^2\le\inn{w^*-v^\dagger}{d}=\inn{w^*}{d}=\inn{P_Kw^*}d\le\norm{P_Kw^*}\norm{d}$, since $v^\dagger\perp K\ni d$. Taking the infimum
over $w^*\in W_t(x^*)$ gives the bound.
If $x^*\notin\dom W_t$, the same bound at any $x\in\mathcal S\cap\dom W_t$ and the triangle inequality give
$\rho_t(x^*)\le\norm{x-x^*}+c^{-1}\dist\bigl(0,P_KW_t(x)\bigr)$.

The domain condition determines which form of the bound applies. For the exact family $\dom W_t=\ran D_t$, so $x^*\in\dom W_t$ fails at boundary points of the convex support of a compactly supported prior, for
instance at an atom of a finite prior that is an extreme point of the convex hull of its atoms. There this general form must be used.

(ii) On a convex part of the fiber, the same estimate uses only monotonicity along the fiber.
Let $x^\dagger_t\in U$ be selected, so $P_KW_t(x^\dagger_t)=0$, and $d:=x^*-x^\dagger_t\in K$. By \eqref{eq:local-fibre-monotone},
$c_K\norm d^2\le\inn{P_KW_t(x^*)-P_KW_t(x^\dagger_t)}{d}=\inn{P_KW_t(x^*)}d\le\norm{P_KW_t(x^*)}\norm d$. If $x^\ddagger\in U$ is another
selected point, \eqref{eq:local-fibre-monotone} at $(x^\dagger_t,x^\ddagger)$ gives $c_K\norm{x^\dagger_t-x^\ddagger}^2\le0$.

For a Bayes denoiser, the fiber Hessian supplies the monotonicity constant.
With $U\subseteq\ran D_t$, $W_t=\nabla R_t$ is smooth on $U$.
For $x,x'\in U$, the segment between them lies in $U\subseteq\mathcal S$.
Putting $h:=x-x'\in K$, we have
$\inn{P_K(W_t(x)-W_t(x'))}{h}=\int_0^1h^\top\nabla^2R_t(x'+sh)h\,ds$. By \eqref{eq:fibre-hessian}, for $h\in K$ and $u\in U$,
$h^\top\nabla^2R_t(u)h=h^\top\bigl(\Sigma_{K\mid N}(z)^{-1}-t^{-1}I\bigr)h\ge\bigl(\lambda_{\max}(\Sigma_{K\mid N}(z))^{-1}-t^{-1}\bigr)\norm h^2$ with
$z=D_t^{-1}(u)\in D_t^{-1}(U)$, which gives \eqref{eq:local-fibre-monotone} with the stated $c_K$.

\end{proof}

\begin{proof}[Proof of \Cref{thm:sharp-local-reconstruction}]
\emph{Strategy.} We rescale the equation to obtain an invertible linearization, use a contraction to construct the solution, and expand the two components to find the error.
Dividing the kernel component by $\alpha$ gives a block-triangular linearization at the truth whose invertibility needs only $H_K\succ0$.

By \Cref{prop:small-sigma} and its proof, on a fixed closed ball $U$ about $x^*$ contained in $\ran D_t$ for small $t$,
$D_t^{-1}=I+t\nabla g+(t^2/2)\nabla\Delta g+O_{C^1}(t^3)$, so that, uniformly in $C^1(U)$,
\begin{equation}\label{eq:sharp-local-W}
 \widehat W_t(x)=\nabla g(x)+\tfrac t2\nabla\Delta g(x)+\psi_t(x)+O(t^2),\qquad \norm{\psi_t}_{C^1(U)}\le\varepsilon_W(t).
\end{equation}
To keep the linearization invertible as the parameters vanish, we use the equivalent system
\[
 \mathcal H(x)=
 \begin{pmatrix}
 A^*(Ax-Ax^*-v)+\alpha P_N\widehat W_t(x)\\
 P_K\widehat W_t(x)
 \end{pmatrix}=0\quad\hbox{in }N\oplus K .
\]
In coordinates $x=x^*+n+k$, at $(\alpha,t,\psi_t,v)=(0,0,0,0)$ its derivative at $x^*$ is
\[
 M=\begin{pmatrix}B&0\\P_KH|_N&H_K\end{pmatrix},
\]
which is invertible.

We now construct the solution by contraction.
On a sufficiently small fixed closed ball, \eqref{eq:sharp-local-W} shows that $\nabla\mathcal H$ differs from $M$ by
$O(\alpha+t+\varepsilon_W(t)+\norm{x-x^*})$, and $\norm{\mathcal H(x^*)}=O(q)$, using $P_K\nabla g(x^*)=0$. For small $q$ the map
$x\mapsto x-M^{-1}\mathcal H(x)$ is a contraction with constant at most $1/2$ that maps the ball into itself.
We take $U$ to be this contraction ball. It depends only on $g$, $A$ and $x^*$. The contraction gives a unique zero in $U$ and
$d=\hat x-x^*=O(q)$.

We obtain the error by expanding the two components at $x^*$.
Using $\psi_t(x^*+d)=\psi_t(x^*)+O(\varepsilon_W(t)\norm d)$ gives
\[
 Bn=A^*v-\alpha P_N\nabla g(x^*)+O(q^2),\qquad
 H_Kk=-P_KHn-\tfrac t2P_K\nabla\Delta g(x^*)-P_K\psi_t(x^*)+O(q^2).
\]
Solving gives \eqref{eq:sharp-local-expansion}.

For the exact force, we also verify local minimality. For $\psi_t=0$, the objective Hessian has $N$ block $B+O(\alpha)$ and kernel Schur
complement $\alpha(H_K+O(t+\norm d+\alpha))\succ0$, so the zero is a strict local minimum.

The expansion also gives the scale requirement in (ii).
For $\alpha\asymp\delta$, the terms due to data noise and prior weight are $O(\delta)$, and $q^2\le C(\delta^2+t^2)$.
Absorbing $Ct^2$ into $\norm b\,t/2$ gives $\tfrac12\norm b\,t-C\delta\le\norm{\hat x-x^*}\le\tfrac32\norm b\,t+C\delta$ for small
parameters, which is (ii).
For $b=0$ the term $Ct^2$ remains.

For (iii), the rescaled residual is $(P_N\zeta,P_K\zeta/\alpha)$, and the contraction argument gives a uniform inverse bound for the rescaled
system.
\end{proof}

\begin{lemma}[Force error of a perturbed denoiser]\label{lem:perturbed-force}
In the setting of \Cref{thm:sharp-local-reconstruction}, let $\widehat D_t=D_t+\eta\phi$ with $\eta\ge0$ and $\phi\in C_b^2$, used with its
local inverse branch, and put $e=H_K^{-1}P_K\phi(x^*)$. Its force $\widehat W_t=W_t+\psi_t$ satisfies $\varepsilon_W(t)\le C\eta/t$ and
$-H_K^{-1}P_K\psi_t(x^*)=\frac\eta te+O(\eta+\eta^2/t)$. Hence \eqref{eq:sharp-local-expansion} holds with the term $\frac\eta te$ and
$q=\delta+\alpha+t+\eta/t$.
\end{lemma}
\begin{proof}
We transfer the denoiser perturbation to its inverse. Let $u\in U$. The local inverse $z=\widehat D_t^{-1}(u)$ solves $z=D_t^{-1}(u-\eta\phi(z))$, so
$z=D_t^{-1}(u)+O(\eta)$ and, since $\nabla D_t^{-1}=I+O(t)$ and $D_t^{-1}(u)-u=O(t)$,
$\widehat D_t^{-1}(u)=D_t^{-1}(u)-\eta\phi(u)+O_{C^1}(\eta t+\eta^2)$.
The $C^1$ bound uses $\phi\in C_b^2$. Dividing by $t$,
$\widehat W_t=W_t+\psi_t$ with $\psi_t=-(\eta/t)\phi+O_{C^1}(\eta+\eta^2/t)$, so $\varepsilon_W(t)\le C\eta/t$ and
$-H_K^{-1}P_K\psi_t(x^*)=\frac\eta te+O(\eta+\eta^2/t)$. Both error terms are $O(q^2)$ with $q=\delta+\alpha+t+\eta/t$.
\end{proof}

\begin{proof}[Proof of \Cref{cor:sharp-scale}]
(i) By \Cref{lem:perturbed-force}, \eqref{eq:sharp-local-expansion} holds with $q=\delta+\alpha+t+\eta/t=O(\delta+\sqrt\eta)$ for $t\asymp\sqrt\eta$,
and its terms are $O(\delta)$ (data noise, prior weight), $O(t)=O(\sqrt\eta)$ (scale bias) and $O(\eta/t)=O(\sqrt\eta)$ (model error).
\Cref{thm:sharp-local-reconstruction}(iii) adds $C(\norm{P_N\zeta}+\norm{P_K\zeta}/\alpha)\le C(\alpha\varepsilon+\varepsilon)$.

(ii) We rescale the noise level to separate the leading minimization from the remainder.
Write $t=t_\eta s$, $\lambda=(\norm e/\norm b)^{1/2}$ and $m=(\norm b\norm e)^{1/2}$.
Then $\norm{tb-(\eta/t)e}=\sqrt\eta\,\Phi(s)$ with
\[
 \Phi(s)^2=m^2(s^2+s^{-2})-2\inn be=\Phi(1)^2+m^2(s-s^{-1})^2,\qquad
 \Phi(1)^2=2(\norm b\norm e-\inn be),
\]
so $s=1$ is the unique minimizer of $\Phi$, and $q=\delta+\alpha+t+\eta/t$ satisfies
$t+\eta/t=\sqrt\eta(\lambda s+\lambda^{-1}s^{-1})$.

We next show that the remainder does not change the limiting minimizer.
By \eqref{eq:sharp-local-expansion} and \Cref{lem:perturbed-force},
$\norm{\hat x-x^*}/\sqrt\eta=\Phi(s)+\mathcal R(s)$ with
$\abs{\mathcal R(s)}\le C(\delta+q^2)/\sqrt\eta$. On $[M^{-1},M]$, $q\asymp\sqrt\eta$ and
$\delta=o(\sqrt\eta)$, so $\mathcal R\to0$ uniformly.

To exclude minimizers outside this interval, we bound the remainder by the growth of the leading term.
Elsewhere $q^2\le\varepsilon_0q$ gives
$\abs{\mathcal R(s)}\le C\varepsilon_0\max\{\lambda,\lambda^{-1}\}(s+s^{-1})+o(1)$, while for
$M^2\ge3$, $\Phi(s)\ge m\abs{s-s^{-1}}\ge\tfrac m2(s+s^{-1})$. If $\varepsilon_0$ is so small that
$C\varepsilon_0\max\{\lambda,\lambda^{-1}\}<m/4$, then $\Phi+\mathcal R\ge\tfrac m4M-o(1)$ outside
$[M^{-1},M]$, which exceeds $\Phi(1)+o(1)\le2m+o(1)$ once $M>8+4/m$. Hence the minimizers lie in
$[M^{-1},M]$ for small $\eta$.
There $\Phi+\mathcal R\to\Phi$ uniformly, so $s^*\to1$ and
$\min(\Phi+\mathcal R)\to\Phi(1)$ follow.
\end{proof}